%% file: main.tex
\pdfoutput=1

\documentclass[11pt]{article}

\usepackage[final]{acl}

\usepackage{times}
\usepackage{latexsym}

\usepackage[T1]{fontenc}

\usepackage[utf8]{inputenc}

\usepackage{microtype}

\usepackage{inconsolata}

\usepackage{graphicx}

\usepackage{hyperref}
\usepackage{array}
\usepackage{booktabs}
\usepackage{multirow}
\usepackage{multicol}
\usepackage[round-mode=places, table-format=2.3, round-precision=2]{siunitx}
\usepackage{amsmath,amssymb}
\usepackage{bbm}
\usepackage{subcaption}
\usepackage[export]{adjustbox}
\usepackage{supertabular}
\usepackage{afterpage}

\usepackage{comment}
\usepackage{cleveref}
\usepackage{placeins}
\usepackage[table]{xcolor}

\usepackage{textcomp}  
\usepackage{scalerel}  
\usepackage{relsize}
\def\affiluhel{\scalerel*{\includegraphics{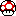}}{\textrm{\footnotesize\textbigcircle}}}
\def\affileqcontrib{\scalerel*{\includegraphics{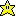}}{\textrm{\footnotesize\textbigcircle}}}
\def\affilsilo{\scalerel*{\includegraphics{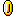}}{\textrm{\footnotesize\textbigcircle}}}
\def\affilaman{\scalerel*{\includegraphics{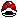}}{\textrm{\footnotesize\textbigcircle}}}
\def\affilubs{\scalerel*{\includegraphics{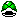}}{\textrm{\footnotesize\textbigcircle}}}
\def\affilcimat{\scalerel*{\includegraphics{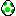}}{\textrm{\footnotesize\textbigcircle}}}
\def\affilmilano{\scalerel*{\includegraphics{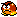}}{\textrm{\footnotesize\textbigcircle}}}
\def\affilcharles{\scalerel*{\includegraphics{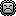}}{\textrm{\footnotesize\textbigcircle}}}
\def\affildarmstadt{\scalerel*{\includegraphics{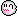}}{\textrm{\footnotesize\textbigcircle}}}
\def\affilaveni{\scalerel*{\includegraphics{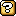}}{\textrm{\footnotesize\textbigcircle}}}
\def\affilhitz{\scalerel*{\includegraphics{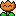}}{\textrm{\footnotesize\textbigcircle}}}
\def\affilupenn{\scalerel*{\includegraphics{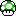}}{\textrm{\footnotesize\textbigcircle}}}

\newcommand\orgname[1]{\noindent\textbf{#1}}

\title{SemEval-2025 Task 3: Mu-SHROOM, the Multilingual Shared Task on Hallucinations and Related Observable Overgeneration Mistakes}

\author{
 \textbf{Raúl Vázquez\textsuperscript{\affiluhel\affileqcontrib}}\qquad
 \textbf{Timothee Mickus\textsuperscript{\affiluhel\affileqcontrib}} 
\\
 \textbf{Elaine Zosa\textsuperscript{\affilsilo}}\qquad
 \textbf{Teemu Vahtola\textsuperscript{\affiluhel}}\qquad
 \textbf{J\"org Tiedemann\textsuperscript{\affiluhel}}\qquad
 \textbf{Aman Sinha\textsuperscript{\affilaman}} 
\\
 \textbf{Vincent Segonne\textsuperscript{\affilubs}}\qquad
 \textbf{Fernando S\'anchez-Vega\textsuperscript{\affilcimat}}\qquad
 \textbf{Alessandro Raganato\textsuperscript{\affilmilano}}\qquad
 \textbf{Jindřich Libovický\textsuperscript{\affilcharles}} 
\\
 \textbf{Jussi Karlgren\textsuperscript{\affilsilo}}\qquad
 \textbf{Shaoxiong Ji\textsuperscript{\affildarmstadt\affiluhel}}\qquad
 \textbf{Jindřich Helcl\textsuperscript{\affilcharles}}\qquad
 \textbf{Liane Guillou\textsuperscript{\affilaveni}} 
\\
 \textbf{Ona de Gibert\textsuperscript{\affiluhel}}\qquad
 \textbf{Jaione Bengoetxea\textsuperscript{\affilhitz}}\qquad
 \textbf{Joseph Attieh\textsuperscript{\affiluhel}}\qquad
 \textbf{Marianna Apidianaki\textsuperscript{\affilupenn}}
\\
\\
 \textsuperscript{\affileqcontrib}{\small Equal contribution. Other authors listed in reverse alphabetical order.}
\\[.5em]
 \textsuperscript{\affiluhel}University of Helsinki\quad
 \textsuperscript{\affilsilo}SiLO\quad
 \textsuperscript{\affilaman}Université de Lorraine \& ICANS Strasbourg\\
 \textsuperscript{\affilubs}Université Bretagne Sud\quad
 \textsuperscript{\affilcimat}CIMAT A. C.\quad
 \textsuperscript{\affilmilano}University of Milano-Bicocca\quad
 \textsuperscript{\affildarmstadt}TU Darmstadt\\
 \textsuperscript{\affilaveni}Aveni\quad
 \textsuperscript{\affilcharles}Charles University\quad
 \textsuperscript{\affilhitz}HiTZ Basque Center for Language Technology - Ixa \\
 \textsuperscript{\affilupenn}University of Pennsylvania
\\
 \small{
   \textbf{Correspondence:} {\tt \{raul.vazquez,timothee.mickus\}@helsinki.fi}
 }
}

\begin{document}
\maketitle
\begin{abstract}

We present the Mu-SHROOM shared task which is focused on detecting hallucinations and other overgeneration mistakes in the output of instruction-tuned large language models (LLMs). 
Mu-SHROOM addresses general-purpose LLMs in 14 languages, and frames the hallucination detection problem as a span-labeling task. We received 2,618 submissions from 43 participating teams employing diverse methodologies. The large number of submissions underscores the interest of the community in hallucination detection. We present the results of the participating systems and conduct an empirical analysis  to identify key factors contributing to strong performance in this task.  
We also emphasize relevant current challenges, notably the varying degree of hallucinations 
across languages and the high annotator disagreement when  labeling hallucination spans.

\begin{center}
  \begin{minipage}{\linewidth}
    \centering
    \raisebox{-0.2\height}{\includegraphics[width=1em]{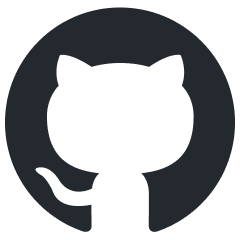}}%
    \hspace{0.5em}%
    {\small\texttt{\href{https://github.com/Helsinki-NLP/mu-shroom}{\tt Helsinki-NLP/mu-shroom}}}
  \end{minipage}
  \begin{minipage}{\linewidth}
    \centering
    \raisebox{-0.2\height}{\includegraphics[width=1em]{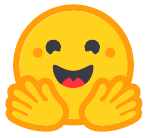}}%
    \hspace{0.5em}%
    {\small\texttt{\href{https://huggingface.co/datasets/Helsinki-NLP/mu-shroom}{\tt Helsinki-NLP/mu-shroom}}}
  \end{minipage}
\end{center}
\end{abstract}

\section{Lets a-go! Introduction}

As generative AI systems become increasingly integrated into real-world applications we expect them to produce fluent and coherent text  \citep[e.g.,][]{rohrbach-etal-2018-object, Lee2018HallucinationsIN}. 
\begin{figure}[ht!]
    \centering
    \includegraphics[width=0.7\columnwidth]{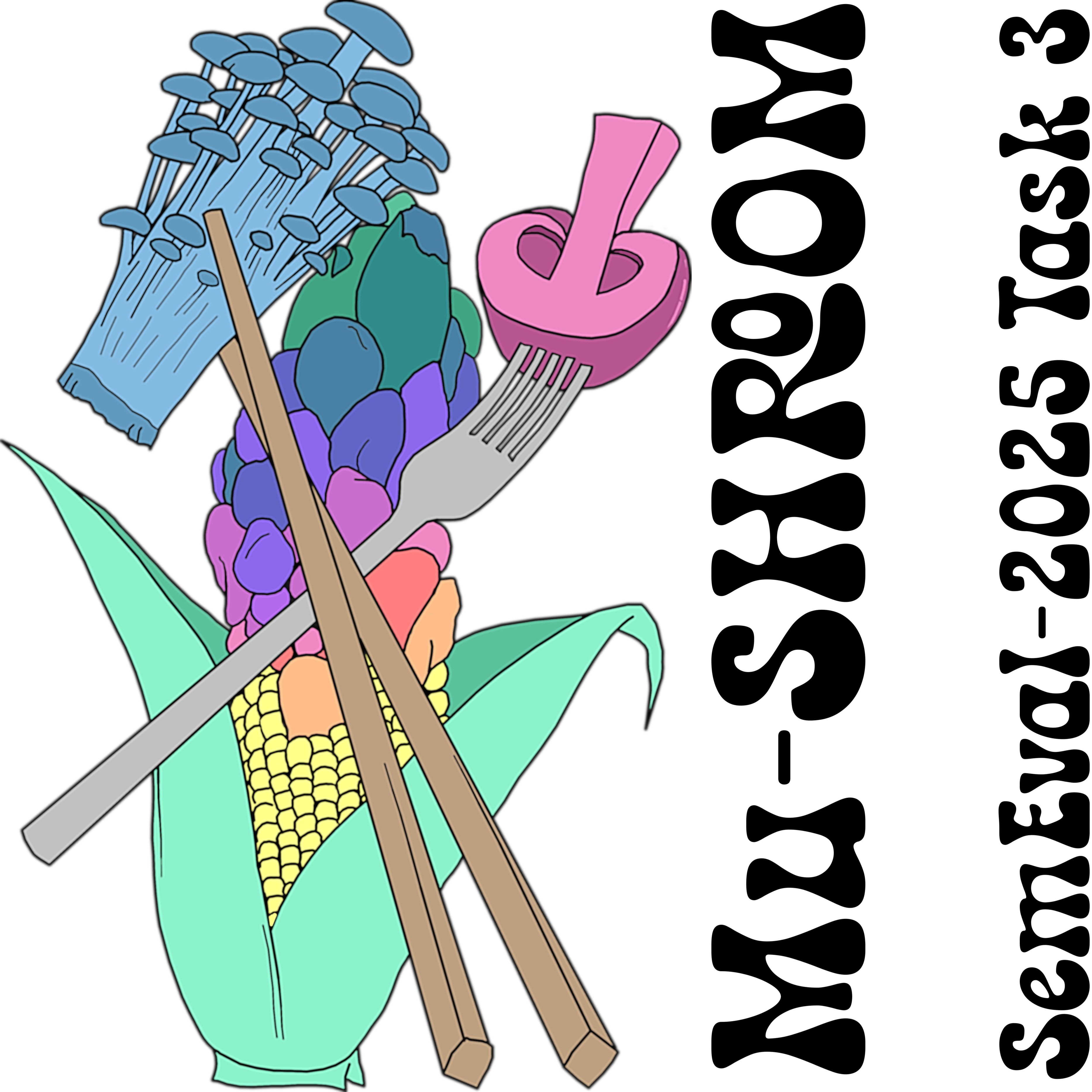}
    \caption{The Mu-SHROOM logo.}
    \label{fig:logo}
\end{figure}
However, a critical issue undermines their reliability: these models frequently generate outputs that are highly fluent but factually incorrect, a phenomenon known as hallucination. 
Hallucinations, as presently observed, are characterized by a disregard of the truth value of statements in favor of persuasive or plausible-sounding language, carrying consequences such as the spread of misinformation, and erosion of user trust \citep{Hicks2024}. 
Compounding this issue is the tendency of hallucinations to "snowball": when models are prompted to provide evidence or explanations for a false claim, they often generate coherent but false statements, further entrenching misinformation \citep{Zhang2023SirensSI, Hicks2024}. Addressing hallucinations is crucial for building systems that the public can trust.  Despite its significance, detecting hallucinations at scale remains a major challenge, with no clear universally effective solution currently available.

The Mu-SHROOM task\footnote{\url{https://helsinki-nlp.github.io/shroom/2025}} aims to contribute to advancing research in this direction. Mu-SHROOM builds on the SHROOM shared task \citep{mickus-etal-2024-semeval}, expanding its scope and addressing key limitations. Unlike SHROOM which focused solely on English, Mu-SHROOM incorporates multilingual data across 14 languages to account for potential variations in hallucination rates \citep{guerreiro-etal-2023-looking}. It also addresses  general-purpose LLMs, reflecting the dominance of such models in current research, and introduces token-level annotations for more precise hallucination detection. By providing a 
richly annotated multilingual dataset and evaluation metrics, Mu-SHROOM aims to advance research on hallucination patterns, improve detection methodologies, and foster community collaboration in NLG and factual consistency assessment.

The Mu-SHROOM dataset consists of a collection of prompts, model outputs, logits, and identifiers for openly available LLMs. 
The dataset encompasses 10 languages with validation and test data (Modern Standard Arabic, German, English, Spanish, Finnish, French, Hindi, Italian, Swedish and Mandarin Chinese), 4 test-only (``surprise'') languages (Catalan, Czech, Basque and Farsi), as well as  unlabeled training data for English, Spanish, French, and Chinese. 
Supplementary metadata, including raw annotations before post-processing and the Wikipedia URLs used as references, as well as 
the scripts used to generate model outputs for all 14 languages and code for the annotation and submission interfaces are all publicly available.\footnote{See \url{https://github.com/Helsinki-NLP/mu-shroom} and \url{https://huggingface.co/datasets/Helsinki-NLP/mu-shroom}}


The shared task attracted a total of 43 teams, resulting in over 2,600 submissions during the three-week evaluation phase. The strong participation and diverse methodologies signal the task's success. Notably, many teams relied on a few key models, often using synthetic data for fine-tuning or zero-shot prompting. While 64–71\% of the teams outperform our baseline, top-scoring systems perform at random for the most challenging items. We present the results and provide a thorough analysis of the strengths and limitations of current hallucination detection systems. 


\section{Down the warp pipe: Related works}
Hallucination in NLG has been widely studied since the shift to neural methods \citep{vinyals2015neural,raunak-etal-2021-curious,maynez-etal-2020-faithfulness, Augenstein2024}. 
Despite significant progress, there remains minimal consensus on the optimal framework for detecting and mitigating hallucinations, partly due to the diversity of tasks that NLG encompasses \citep{ji-etal-2023-survey, huang-etal-2024-survey}. Recent advances further highlight the urgency for addressing this issue, as hallucinations can lead to the propagation of incorrect or misleading information, particularly in high-stakes domains such as healthcare, legal systems, and education \citep{zhang2023snowball,Zhang2023SirensSI}.
This has led to a  recent but flourishing body of work interested in detecting and mitigating hallucinations \cite{Farquhar2024,gu2024anahv,mishra2024finegrained}, as well as studies on how to best define and articulate this phenomenon \citep{guerreiro-etal-2023-looking,rawte-etal-2023-troubling,huang-etal-2024-survey,liu2024hallucinationscode}.


More immediately relevant to our shared task are pre-existing benchmarks and datasets. 
\citet{li-etal-2023-halueval} introduced HaluEval which is focused on dialogue systems but relies on closed, non-transparent models, limiting reproducibility. Other benchmarks, such as those by \citet{liu-etal-2022-token} and \citet{zhou-etal-2021-detecting}, use synthetic data for token-level hallucination detection. The SHROOM dataset \citep{mickus-etal-2024-semeval} provides 4k multi-annotated datapoints for task-specific NLG systems. Recently, \citet{niu-etal-2024-ragtruth} introduced RAGTruth, a large-scale corpus with 18,000 annotated responses for analyzing word-level hallucinations in RAG frameworks. \citet{ijcai2024p0687} proposed FactCHD to specifically study hallucinations due to fact conflation. Additionally, \citet{rawte-etal-2023-troubling} introduced a comprehensive dataset and a vulnerability index to quantify LLMs' susceptibility to hallucinations.
Most of these datasets focus on English (or Chinese, \citealp{cheng2023evaluatinghallucinationschineselarge}).


\section{Collecting the coins: Data}
\label{sec:data}

We begin with a description of the general process, and then note specific \emph{ad-hoc} departures from this process for each language. 
The dataset covers 38 LLMs over 14 languages, out of which 4 (CA, CS, EU, FA) are test-only with about 100 datapoints. The other 10 languages (AR, DE, EN, ES, FI, FR, HI, IT, SV, ZH) include both a validation split of 50 datapoints  and a test split of 150 datapoints.\footnote{
    Due to technical and replicability issues, we manually removed 1 datapoint from EU test, 1 from SV val and 3 from SV test. A handful of languages contained extra test items.
}

The construction of the dataset started with an automatic extraction of 400 Wikipedia pages, with a focus on pages  available in multiple languages of interest.  We eventually increased this extraction to 762 links to guarantee a large number of Wikipedia pages for all languages.  From this point, the process we follow for creating the Mu-SHROOM dataset is divided into two phases: datapoint creation and data annotation. 

\paragraph{Data creation.}
The datapoint creation for each language was spearheaded by one of our organizers proficient in the language. 
Appendix \ref{appx:datacreation_guidelies} (esp. \Cref{fig:data_guidelines}) describes the process in detail. 
In short, we manually selected and read 200 Wikipedia pages (100 for test-only languages), and wrote for each page one question that could be answered with the information it contained. 
Due to variations across Wiki projects, the set of selected pages  and the constructed questions  vary across languages.\footnote{
E.g., $\approx75\%$ of HI datapoints have no equivalent  in other languages. 
}
Questions had to be \emph{factual} (i.e., not a matter of opinion) and \emph{closed} (i.e., answerable with a closed set of answers, such as numbers, places, names, etc). 

For each question, we then generated multiple LLM answers: We identified existing open-weight instruction-tuned LLMs capable of handling the languages of interest (cf. Table \ref{tab:llms} in Appendix for a list), and produce multiple outputs for each question by varying generation hyperparameters (top $p$, top $k$, temperature).
We then manually selected one output to annotate for each question which satisfied a set of  criteria: It was  fluent and in the language of interest; it was relevant to the input question; it appeared to contain hallucinations or data worth annotating. A subset of the remaining  outputs was set aside to serve as an unlabeled training  set. 

\paragraph{Data annotation.} 
We frame the data annotation task as a span-labeling task where human annotators are asked to highlight text spans in the model output that contain an overgeneration or hallucination. Within this task, we define hallucination as ``\emph{content that contains or describes facts that are not supported by a provided reference}''. 

Annotation were collected using a custom platform displaying the input question, the answer output by the model, and the source the Wikipedia page from which the question was derived. The annotators' task was to highlight all spans of text in the answer that were not supported by  information present in the Wikipedia page, which  corresponds to an overgeneration or hallucination.

In order to accommodate the complete set of languages in the Mu-SHROOM task using a common set of annotation guidelines, and to cover all eventualities, the annotators were instructed to highlight the minimum number of \textit{characters} 
that would need to be edited or deleted in order to provide a correct answer. The annotators were encouraged to be conservative when highlighting spans, and to focus on content words rather than function words.

With the aim of constraining the scope of the task and ensuring the reliability of the source information used, the annotators were restricted to consulting Wikipedia in order to identify hallucinated content. Whilst the reference Wikipedia page provided should ideally be sufficient for the task, annotators were permitted to browse other Wikipedia articles in order to verify information the reference might not contain, as long as they provided details of any such pages. 
The complete set of annotation guidelines given to the annotators is provided in Appendix \ref{appx:annotation_guidelies}.
All selected outputs from the datapoint creation phase were annotated by at least three annotators, usually with the same three individuals handling all 200 datapoints; exceptions are listed in Appendix \ref{appx:data nonstandard}.

\begin{table}[!t]
    \centering
    \resizebox{0.95\linewidth}{!}{
    \begin{tabular}{l *{7}{S}}
\toprule
 & {{\textbf{AR}}} & {{\textbf{CA}}} & {{\textbf{CS}}} & {{\textbf{DE}}} & {{\textbf{EN}}} & {{\textbf{ES}}} & {{\textbf{EU}}} \\
\midrule
\textbf{Val.} & 0.772233 &  {{---}} &  {{---}} & 0.746592 & 0.448125 & 0.584173 &  {{---}} \\
\textbf{Test} & 0.757736 & 0.802891 & 0.712317 & 0.718128 & 0.485954 & 0.514418 & 0.742699 \\[0.4em]
\midrule
 & {{\textbf{FA}}} & {{\textbf{FI}}} & {{\textbf{FR}}} & {{\textbf{HI}}} & {{\textbf{IT}}} & {{\textbf{SV}}} & {{\textbf{ZH}}} \\
\midrule
\textbf{Val.} & {{---}} & 0.743486 & 0.732893 & 0.796118 & 0.853803 & 0.741530 & 0.566410 \\
\textbf{Test} & 0.751385 & 0.785420 & 0.813802 & 0.801658 & 0.873526 & 0.782353 & 0.584843 \\
\bottomrule
\end{tabular}}
\caption{Annotator agreement measured as Intersection over Union (IoU, cf.~\cref{eq:agg}).}
    \label{tab:agreement}
\end{table}

\paragraph{Annotator agreement.} An overview of the agreement rates obtained by our annotators is shown in \Cref{tab:agreement}, computed as the intersection over union (IoU) of the characters marked as hallucinations by the annotators.
To measure this, assuming $C_n$ is the set of character indices marked as hallucination by our $n$\textsuperscript{th} annotator, we compute
\begin{equation}
    \mathrm{agg} = \frac{1}{n \cdot \left| C_\mathrm{all} \right|} \sum \limits_n \sum \limits_{c_i \in C_\mathrm{all}} \mathbbm{1}\left\{c_i \in C_n\right\} \label{eq:agg}
\end{equation}
where $C_\mathrm{all} = \bigcup C_n $. This is equivalent to a multiset-based IoU, where we keep one copy of a character index for each annotator that marked it as a hallucination.

\begin{figure}
    \centering
    \includegraphics[width=0.85\linewidth]{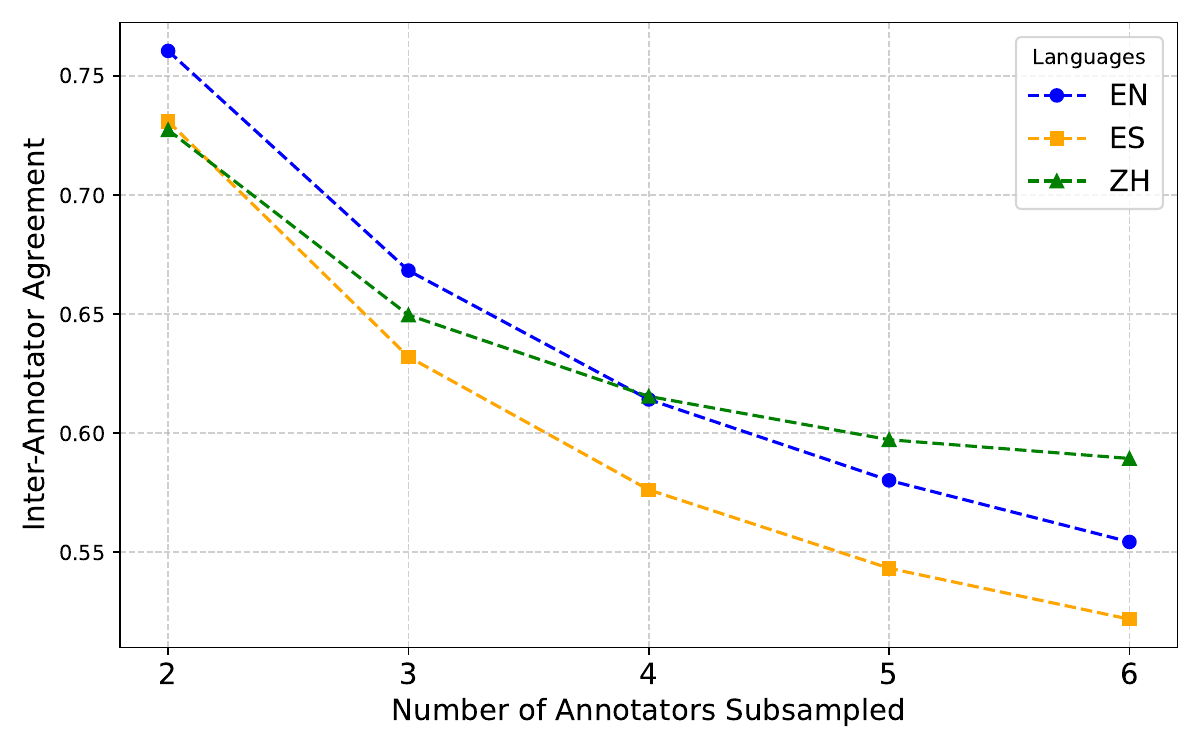}
    \caption{Effects of annotator pool size on inter-annotator agreement (100 random samples, $\sigma\leq\num{0.011725468243379729}$)}
    \label{fig:subsamp-annot}
\end{figure}

Empirically, we observe that ES, EN and ZH yield lower agreement rates, which we can partly link to the higher  number of annotators: 
Remark that a character index marked by a single annotator penalizes the agreement rate by $\frac{n - 1}{n \cdot | C_\mathrm{all}|}$, which tends to $\frac{1}{| C_\mathrm{all}|}$ as $n$ grows.
In fact, if we subsample a lower number of annotations per item for EN, ES and ZH, we obtain the curve in Figure \ref{fig:subsamp-annot} which empirically demonstrates this effect.
It is still worth highlighting that not all of the disagreement we observe can be reduced to this effect, suggesting that the different annotation conditions (see Appendix \ref{appx:data nonstandard}) may also play a significant role, or that there is something fundamentally distinct regarding hallucinations in higher--resource languages.

\begin{table}[!h]
    \centering
    \resizebox{0.95\columnwidth}{!}{
    \begin{tabular}{>{\bf}l *{9}{S[table-format=2.0, round-precision=0]}}
    \toprule
    
    \multirow{2}{*}{\textbf{Error type}} & \multicolumn{8}{c}{\textbf{Language}} \\
    & {{\textbf{AR}}} & {{\textbf{CA}}} & {{\textbf{CS}}}& {{\textbf{ES}}} & {{\textbf{EU}}} & {{\textbf{FI}}} & {{\textbf{FR}}} & {{\textbf{IT}}} & {{\textbf{ZH}}}\\
    \midrule
    Fluency    & 7 & 18 & 24 &  1 & 68 & 16  & 1 &  3 & 11 \\
    Factuality & 97 & 79 & 82 & 66 & 46 & 87  & 57 & 70 & 96\\
    \bottomrule
    \end{tabular}
    }
    \caption{Number of factuality and fluency mistakes in random samples of LLM productions ($n=100$).}
    \label{tab:error rate per type}
\end{table}

\paragraph{Fluency vs. factuality.} One assumption we have adopted thus far, but which needs further verification, is the extent to which hallucinations are indeed a major problem for LLMs. 
To assess this, we manually re-annotated 100 independently sampled LLM outputs from different languages, distinguishing between fluency and factuality errors. Results in Table \ref{tab:error rate per type} show that factuality issues are more pervasive than fluency mistakes, except in Basque. This explains the shift in NLG evaluation priorities, with factual accuracy now outweighing grammaticality as a primary challenge. Additionally, the results reveal a coverage gap across languages: while Spanish, French, Italian and perhaps also Arabic outputs are nearly perfectly fluent, Czech, Catalan, Basque and  Finnish offer a more challenging picture, perhaps due to the fewer available resources; with Basque standing out as an exception, with 68 fluency errors compared to 46 factuality errors. Notably, at least half of the outputs across all languages in this small-scale study contain errors, underscoring  the unreliability of instruction-tuned LLMs and the need for cautiousness when deploying them in real-world applications.

\section{It's a me, Wario: Metrics and baselines}

\paragraph{Metrics.}
We compare the participants' submissions 
using  two metrics: an intersection-over-union metric ($\mathrm{IoU}$) and a correlation metric ($\rho$).
In order to apply the $\mathrm{IoU}$ metric, we first binarize annotations by considering whether a majority of annotators ($>50\%$) marked a character as hallucinated, and then compare the set of indices marked by the system being rated to this binarized set of annotations.
Formally, for one datapoint:
\begin{align}
    C_\mathrm{bin} &= \left\{~c_i ~\middle|~ 0.5 < \sum\limits_n \frac{1}{n}\mathbbm{1}\left\{c_i \in C_n\right\}\right\} \nonumber \\
    \mathrm{IoU} &= {\left|\hat{C}_\mathrm{bin} \cap C_\mathrm{bin}\right|} ~/~ {\left|\hat{C}_\mathrm{bin} \cup C_\mathrm{bin}\right|}
\end{align}
\noindent where $C_\mathrm{bin}$ is the set of binarized character-level annotations derived from the $n$ different sets of annotations $C_n$, and $\hat{C}_\mathrm{bin}$ is the set of characters that the system predicts as hallucinated. 

On the other hand, the $\rho$ metric tries to factor in the lack of thorough consensus we observed in Section 
\ref{sec:data}. 
A drawback of the binarized annotation scheme is that it assumes a single ground truth, which may prove inaccurate  or overly simplistic \citep{Aroyo2015,plank-2022-problem}.
To sidestep this issue, 
we  consider whether the empirical probability of a character being marked by our annotators aligns with the probability derived from the participants' models.
For a given datapoint of length $k$, we formally measure:
\begin{align}
    \Pr\null_{c_i} &= \sum\limits_n \frac{1}{n}\mathbbm{1}\left\{c_i \in C_n\right\}  \nonumber\\
    \mathbf{c} &= \left(~\Pr\null_{c_1}, ~ \dots~,~ \Pr\null_{c_k}\right) \nonumber \\
    \hat{\mathbf{c}} &= \left(~p(c_1 ~| ~ \theta), ~ \dots~,~p(c_k ~| ~ \theta)\right) \nonumber \\
    \rho &= \mathrm{Spearman}\left(\mathbf{c},~\hat{\mathbf{c}} \right)
\end{align}
\noindent where $p(c_i ~| ~ \theta)$ stands for the probability that character $c_i$ is in a hallucinated span, as assigned by a given participating  system, and $\Pr\null_{c_i}$ is our empirical probability. The $\rho$ metric assesses how well the model captures the relative likelihood of hallucination rather than just the binary decision.
In effect, we are measuring the human calibration of the participants' systems \citep{baan-etal-2022-stop}.\footnote{
A handful of datapoints do not contain hallucinations. In such cases, we assign an IoU of 1 if the system's predicted set is also empty, 0 otherwise, and a $\rho$ of 1 if the model assigns the same probability to all tokens, 0 otherwise.
}

The two metrics make different assumptions regarding our data. With the IoU metric, we assume that annotators can reach a consensus as to what counts as hallucination, whereas with the $\rho$ metric, we expect that models should be able to match human variation closely. 
In the interest of lowering the barrier to entry for the shared task, we rank participating systems according to their highest IoU scores and break eventual ties depending on the $\rho$ scores.\footnote{We also provide alternative rankings based on $\rho$ scores in Appendix \ref{appx:alt rankings}.}
In the same vein, we also allowed participants to submit  binary predictions ($\hat{C}_\mathrm{bin}$), continuous predictions ($\hat{\mathbf{c}}$), or both.
If a submission was missing either binary or continuous predictions, we applied default heuristics to derive the missing prediction from the other. We converted continuous predictions $\hat{\mathbf{c}}$ into binary predictions by applying a cutoff of 0.5, and binary predictions $\hat{C}_\mathrm{bin}$ into continuous predictions by assigning a probability of 1 or 0 based on membership. Formally:
\begin{align*}
  \hat{C}_\mathrm{bin}  &= \left\{ ~c_i ~\middle|~ p(c_i ~| ~ \theta) > 0.5 ~\right\} \\
  \mathbf{\hat{c}} &= \left(\mathbbm{1}\left\{c_1 \in \hat{C}_\mathrm{bin}\right\}, ~\dots ~, \mathbbm{1}\left\{c_k \in \hat{C}_\mathrm{bin}\right\} \right)
\end{align*}

\paragraph{Baselines.} To lower the barrier to entry to the shared task, we provided participants with an XLM-R-based baseline system \emph{neural} fine-tuned on the entire test set for token-level classification.\footnote{We used \url{FacebookAI/xlm-roberta-base}. We finetuned for 5 epochs with a learning rate of 2e-5.} This classifier directly maps tokens in an LLM's answer to binary probabilities, without any intermediate fact verification step. 
In addition to this neural baseline, we consider two heuristics: 
\emph{mark-all} where all characters are marked as hallucinated with probability 1, and  \emph{mark-none}  where no  hallucination is found, i.e., all characters get a probability of 0.

The neural baseline is meant first and foremost as a tool for participants to build upon and demonstrate how to map characters to tokens. Without any means of verification of the facts underpinning an LLM output, we have low expectations that this baseline will perform well, especially in zero-shot settings. 
The two heuristics assign probabilities of  0 or 1 uniformly to all characters, which entails that every LLM output is mapped to a constant series of probability. This corresponds to a correlation score of $\rho=0$ in most cases.
As for IoU scores, given our data selection protocol (cf. Section \ref{sec:data}), we expect our dataset to be biased towards samples that contain hallucinations. Therefore, the mark-none baseline should yield lower IoU scores than the mark-all baseline. 

\section{It's a me, Mario: Participants' systems}

\begin{table*}[!ht]
    \centering
    \resizebox{0.99\linewidth}{!}{
        \input{tables/participant_info/team_alphaorder}
    }
    \caption{Summary of 43 participating teams (listed in alphabetical order). First column contains the team handle, second column contains languages the team participated in, and the last column briefly describes their respective approaches.}
    \label{tab:teams_systems}
\end{table*}

43 teams submitted their systems during the evaluation phase, and 35 teams wrote a paper describing their system. In total, we received 2,618 submissions across all languages. In average, 27.2 teams participated in each language. 41 teams submitted systems for English (EN), followed by 32  for Spanish (ES) and 30 for French (FR). 
The languages with the least number of participants 
were our surprise languages: Catalan (CA) with 21 teams; Czech (CS), Basque (EU) and Farsi (FA), with 23 teams each.
Overall, we remark a wide variety of approaches, ranging from QA-- or NER--based finetuning, to time series--based analyses of logits \citep{aryal-akomoize-2025-howard} and to zero-shot RAG-based approaches.
We present an overview of the participating systems in \Cref{tab:teams_systems} and spotlight a few approaches below, noteworthy in that they portray clearly different methodologies that nonetheless performed reasonably well within the shared task.

The \textbf{UCSC} system \citep{huang-etal-2025-ucsc} is designed as a three-stage pipeline: (i) context retrieval, wherein they retrieve relevant pieces of information to assess the factuality of the LLM outputs; (ii) hallucinated fact detection, wherein they identify the incorrect facts based on the retrieved contexts; and (iii) span mapping, wherein the incorrect facts are mapped onto specific segments of the output. The approach furthermore employs prompt optimization to maximize performances.
Multistage frameworks were also deployed by other teams, for instance, \textbf{iai\_msu} \cite{pukemo-etal-2025-iai_msu} developed a three-step approach, with a retrieval-based first step, a self-refine second step, and an ensembling third step.

Another noteworthy entry is that of \textbf{CCNU} \citep{liu-chen-2025-ccnu} --- whose report also incorporates information about unsuccessful attempts and some discussion of the working definition of `hallucination' we used within this shared task.
The CCNU system attempts to emulate a crowd-sourcing approach by utilizing multiple LLM-based agents with different expertise and different knowledge sources. Such crowd-emulation approaches turned out fairly popular within the shared task and were also deployed by a.o.  UCSC \citep{huang-etal-2025-ucsc} or \textbf{Swushroomsia} \citep{mitrovic-etal-2025-swushroomsia}.

Lastly, the \textbf{SmurfCat} system  \citep{rykov-etal-2025-smurfcat} offers an interesting perspective on external knowledge incorporation: \citeauthor{rykov-etal-2025-smurfcat} constructed a synthetic dataset derived from Wikipedia, viz. PsiloQA, so as to finetune LLMs for hallucination span detection. They further refine their models' raw predictions using white-box techniques derived from uncertainty quantification, a perspective also explored, e.g., by \textbf{MALTO} \citep{savelli-etal-2025-malto}. 

The variety of approaches deployed by the participating teams is a clear indicator of the potential for future improvements.

\section{Rainbow Road Completed: Results}


\begin{figure}
    \begin{center}
        
    \begin{subfigure}[b]{0.475\linewidth}
    \includegraphics[max width=\linewidth, trim={0.25cm 0.25cm 0.25cm 0.25cm}, clip]{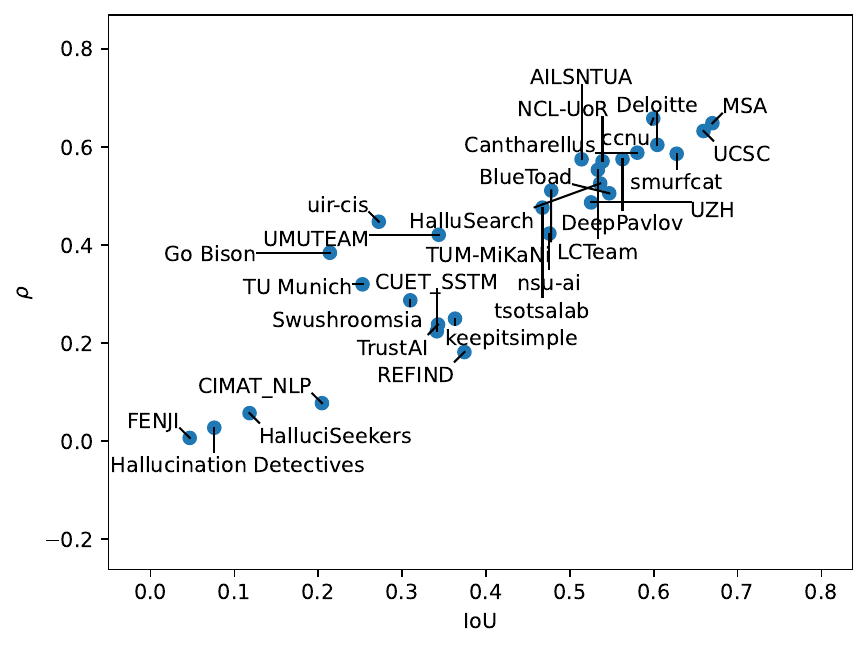}
    \caption{AR}
    \end{subfigure}
    \begin{subfigure}[b]{0.475\linewidth}
    \includegraphics[max width=\linewidth, trim={0.25cm 0.25cm 0.25cm 0.25cm}, clip]{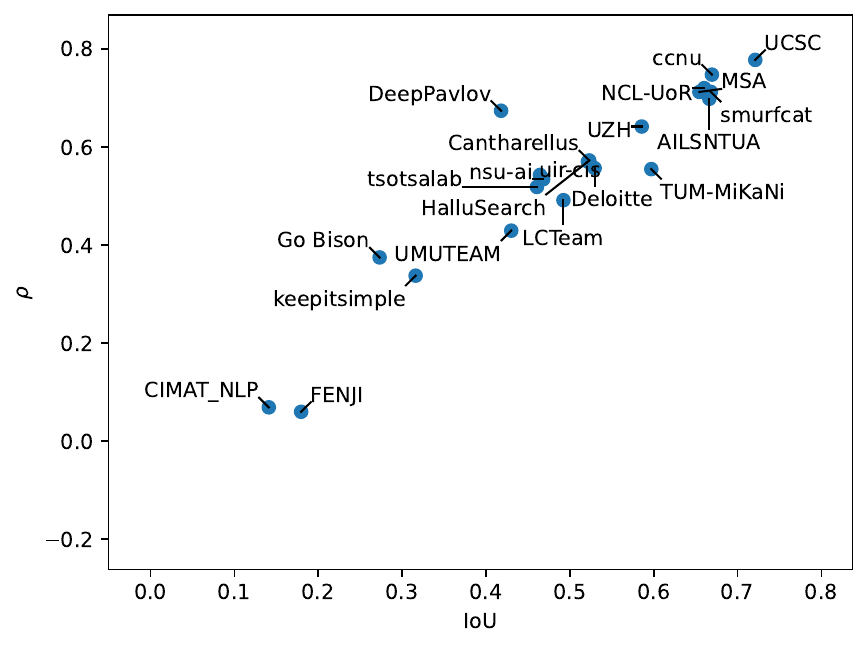}
    \caption{CA}
    \end{subfigure}
    
    \begin{subfigure}[b]{0.475\linewidth}
    \includegraphics[max width=\linewidth, trim={0.25cm 0.25cm 0.25cm 0.25cm}, clip]{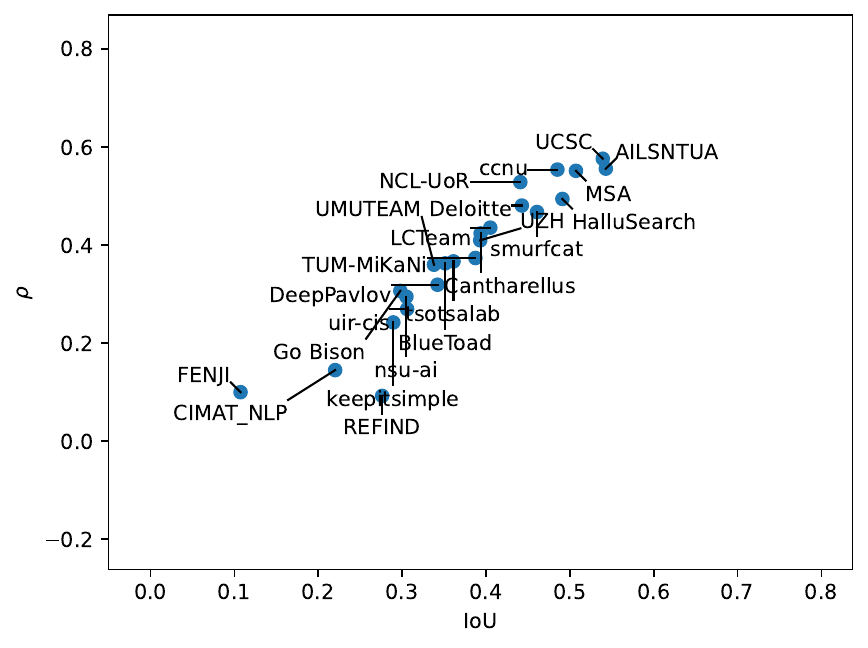}
    \caption{CS}
    \end{subfigure}
    \begin{subfigure}[b]{0.475\linewidth}
    \includegraphics[max width=\linewidth, trim={0.25cm 0.25cm 0.25cm 0.25cm}, clip]{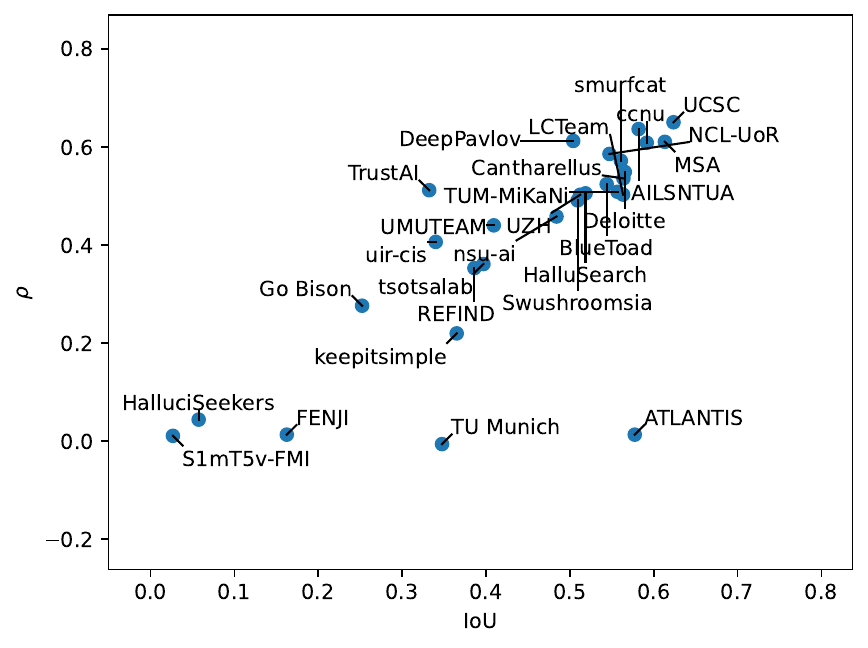}
    \caption{DE}
    \end{subfigure}

    \begin{subfigure}[b]{0.475\linewidth}
    \includegraphics[max width=\linewidth, trim={0.25cm 0.25cm 0.25cm 0.25cm}, clip]{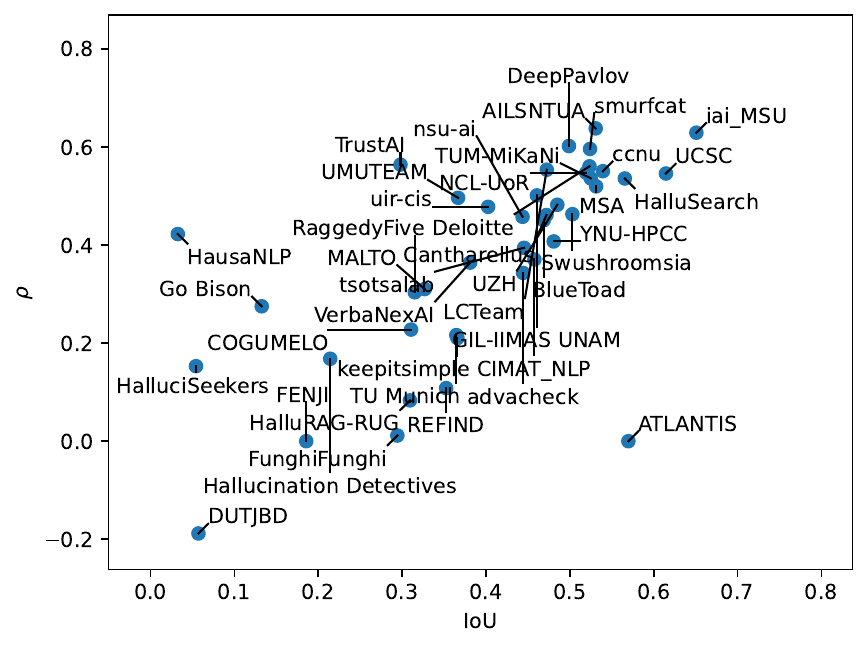}
    \caption{EN}
    \end{subfigure}
    \begin{subfigure}[b]{0.475\linewidth}
    \includegraphics[max width=\linewidth, trim={0.25cm 0.25cm 0.25cm 0.25cm}, clip]{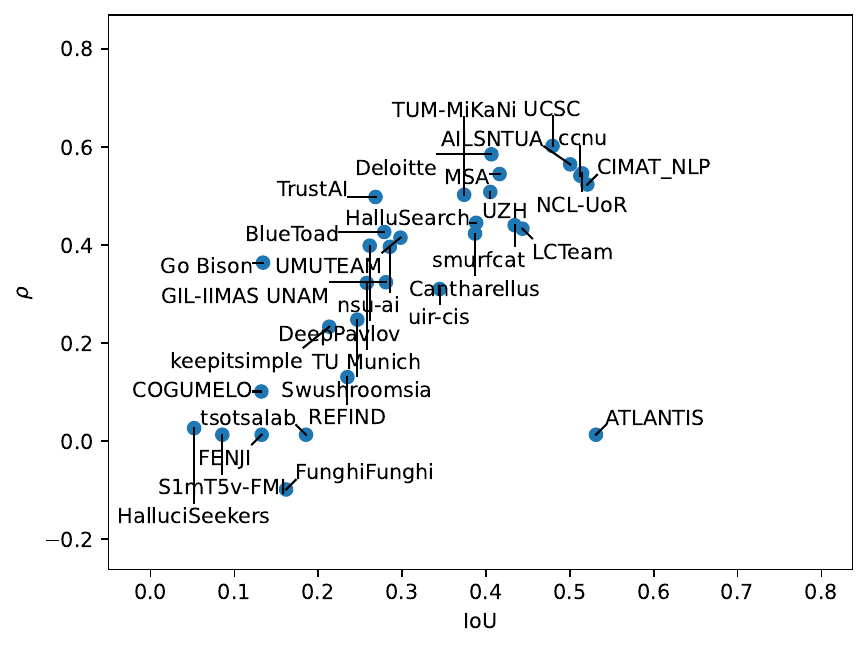}
    \caption{ES}
    \end{subfigure}
    
    \begin{subfigure}[b]{0.475\linewidth}
    \includegraphics[max width=\linewidth, trim={0.25cm 0.25cm 0.25cm 0.25cm}, clip]{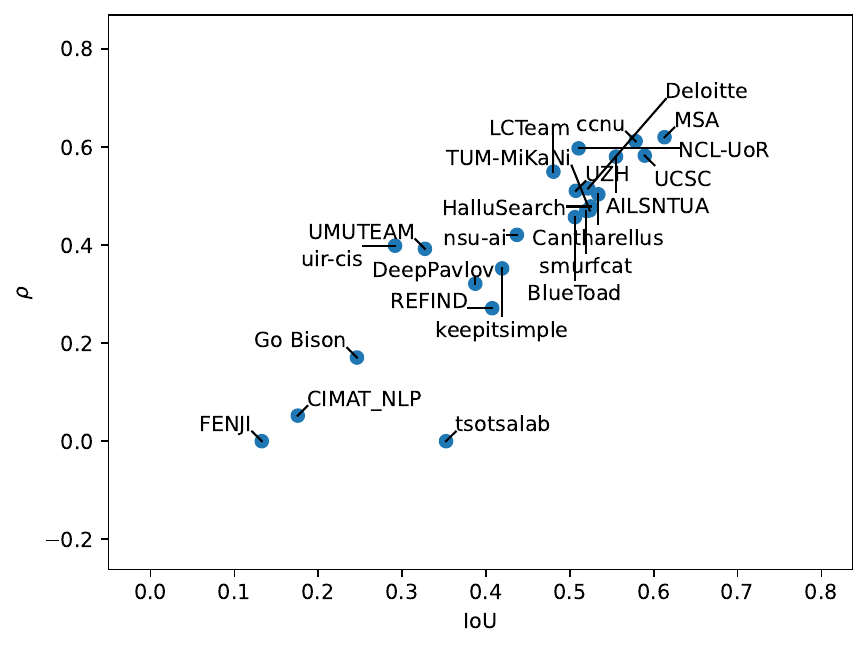}
    \caption{EU}
    \end{subfigure}
    \begin{subfigure}[b]{0.475\linewidth}
    \includegraphics[max width=\linewidth, trim={0.25cm 0.25cm 0.25cm 0.25cm}, clip]{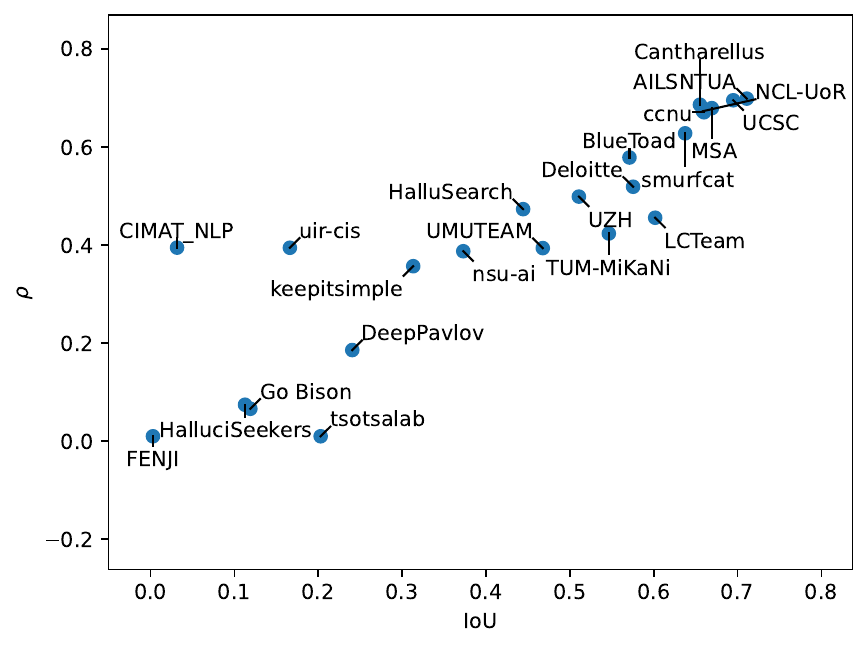}
    \caption{FA}
    \end{subfigure}

    \begin{subfigure}[b]{0.475\linewidth}
    \includegraphics[max width=\linewidth, trim={0.25cm 0.25cm 0.25cm 0.25cm}, clip]{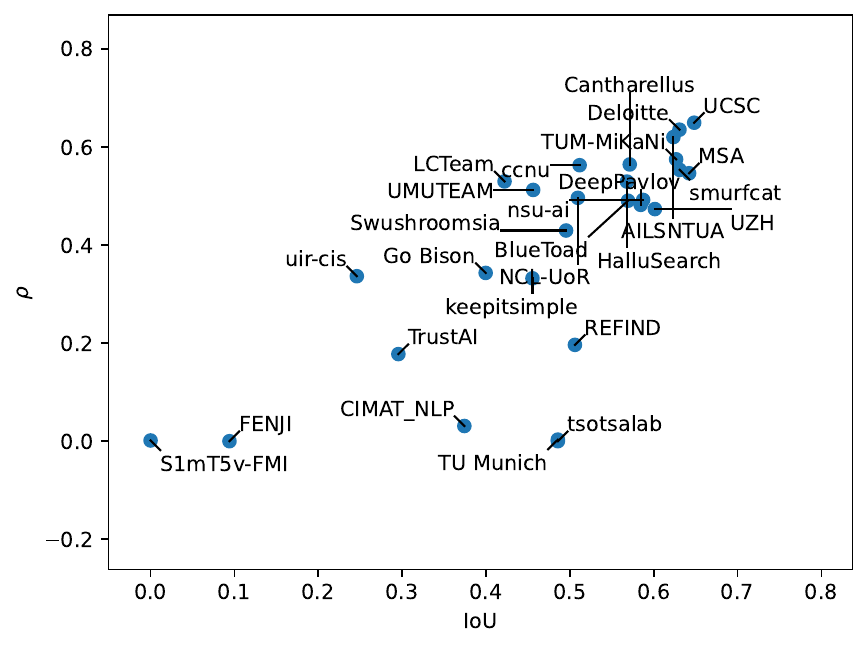}
    \caption{FI}
    \end{subfigure}
    \begin{subfigure}[b]{0.475\linewidth}
    \includegraphics[max width=\linewidth, trim={0.25cm 0.25cm 0.25cm 0.25cm}, clip]{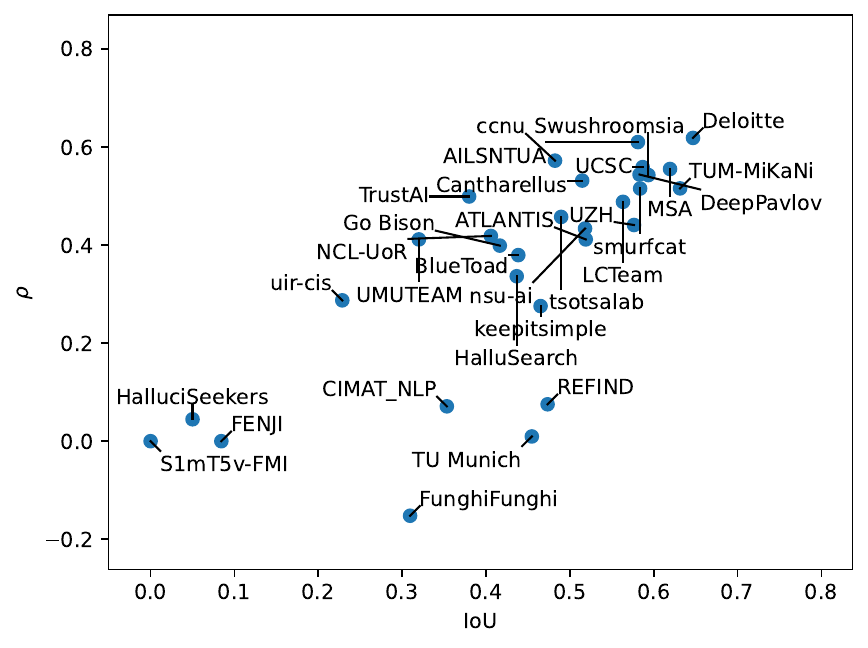}
    \caption{FR}
    \end{subfigure}
    
    \begin{subfigure}[b]{0.475\linewidth}
    \includegraphics[max width=\linewidth, trim={0.25cm 0.25cm 0.25cm 0.25cm}, clip]{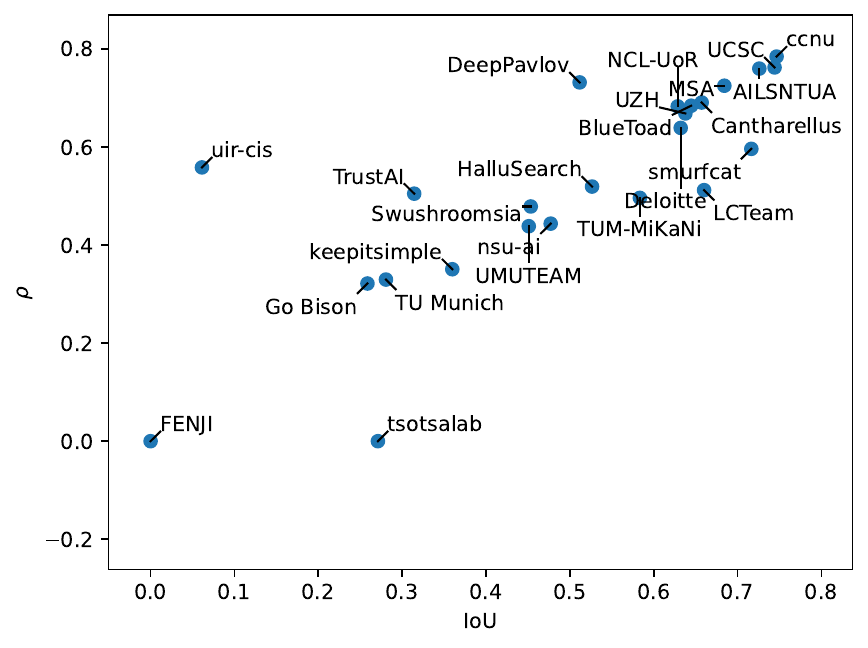}
    \caption{HI}
    \end{subfigure}
    \begin{subfigure}[b]{0.475\linewidth}
    \includegraphics[max width=\linewidth, trim={0.25cm 0.25cm 0.25cm 0.25cm}, clip]{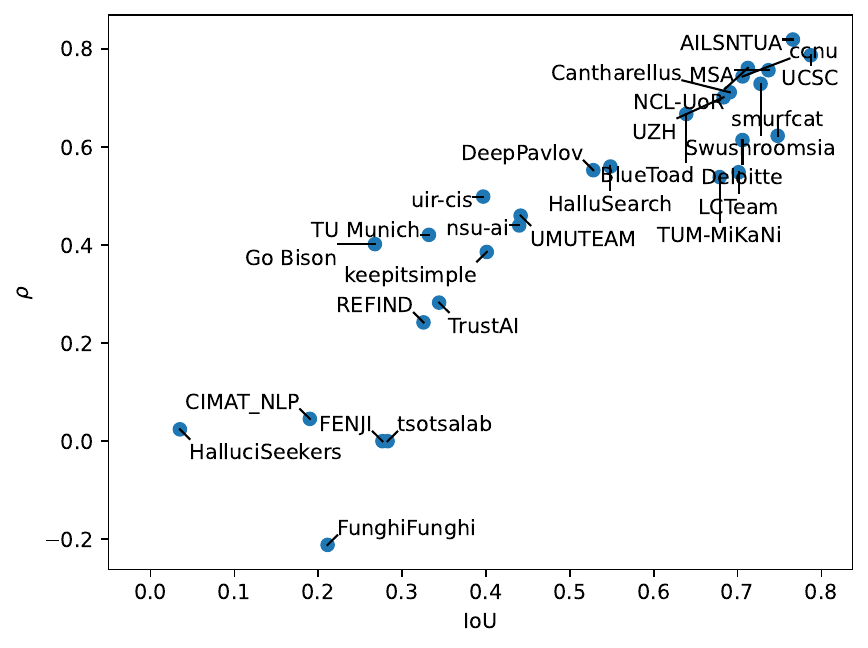}
    \caption{IT}
    \end{subfigure}

    \begin{subfigure}[b]{0.475\linewidth}
    \includegraphics[max width=\linewidth, trim={0.25cm 0.25cm 0.25cm 0.25cm}, clip]{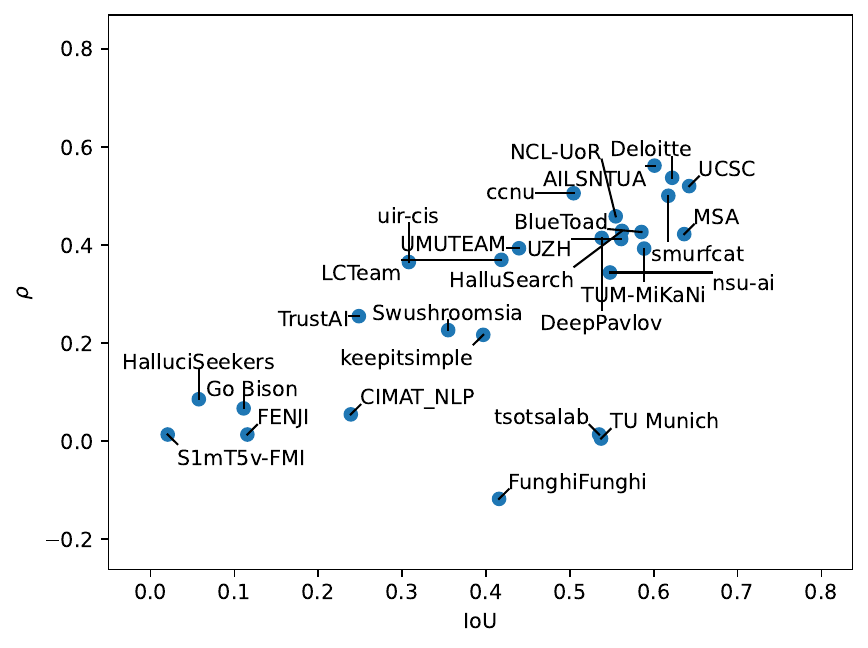}
    \caption{SV}
    \end{subfigure}
    \begin{subfigure}[b]{0.475\linewidth}
    \includegraphics[max width=\linewidth, trim={0.25cm 0.25cm 0.25cm 0.25cm}, clip]{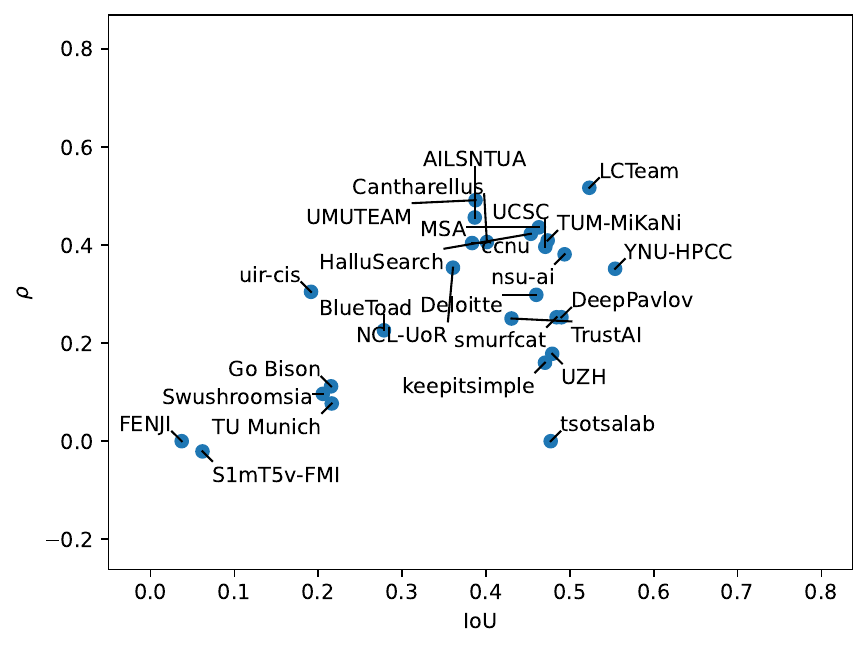}
    \caption{ZH}
    \end{subfigure}
    \end{center}
    
    \caption{Overview of the performance by the best systems from each team in each language. }
    \label{fig:rankings}
\end{figure}

We include an overview of the highest scoring systems  from each team per  language in \Cref{fig:rankings}.
In the interest of space, we defer tables of ranking to Appendix \ref{appx:rankings}.
Most teams outperformed the baselines. The mark-none and neural baselines rank extremely low, both in terms of IoU and $\rho$. 
The mark-all baseline performs better in terms IoU, but remains far below the top teams, highlighting the need for more sophisticated strategies. 

The most consistent top performers coincidentally made submissions to all 14 languages.\footnote{
    Note that we generally do not find evidence that rank differences are statistically significant, cf. \Cref{appx:rankings}.
}
    \textbf{UCSC} \citep{huang-etal-2025-ucsc} appears in the top 3 teams for 11 languages, securing 5 wins (CA, DE, FI, IT, SV) and 5 second-place finishes (CS, EN, EU, FA, HI). Their systems demonstrate a stable IoU-to-$\rho$ ratio mean(IoU$/\rho$)$ = 1.01$.
    \textbf{MSA} \citep{hikal-etal-2025-msa} ranks in the top 3 for 8 languages, winning in 2 (AR, EU) and securing second place in 3 others (DE, FI, SV) and mean(IoU$/\rho$)$=1.03$.
    \textbf{AILS-NTUA} \citep{karkani-etal-2025-ails} performs well across multiple languages, winning in  2 (CS, FA), but showing a less balanced performance between the two metrics: mean(IoU$/\rho$)$=0.93$.
    \textbf{CCNU} \citep{liu-chen-2025-ccnu} ranks first in HI and appears in the top-5 in 9 languages with mean(IoU$/\rho$)$=0.93$. 
    \textbf{Deloitte} \citep{chandler-etal-2025-deloitte} ranks first in FR and places the top-5 in 3 languages.
    \textbf{SmurfCat} \citep{rykov-etal-2025-smurfcat} consistently ranks in the top-5 across 7 languages and never falls out of the top-10.
    \textbf{ATLANTIS} \citep{kobus-etal-2025-atlantis} participated in 4 languages and won the 1st place in ES. However, their $\rho$ scores are near zero in three of their languages, including ES and EN, where they placed 1st and 3rd, respectively. Due to this imbalance, we report the inverse ratio: mean($\rho$/IoU)$ = 0.2$. 
    \textbf{iai\_MSU} \citep{pukemo-etal-2025-iai_msu} competed only in EN, where it secured the 1st place with a system that performs well in both metrics: mean(IoU$/\rho$)$=1.03$.
    \textbf{YNU-HPCC} \citep{chen-etal-2025-ynu} participated in ZH and EN, placing 1st and 15th, respectively. While strong in IoU, its systems struggle in $\rho$, particularly for ZH, resulting in a mean(IoU$/\rho)=1.38$. 

Table~\ref{tab:language_rankings} presents the average performance of systems across languages, highlighting the difficulty differences across languages. We compute the mean IoU and $\rho$ across all teams (excluding baselines) for each language and rank them accordingly based on the average IoU. IT and HI emerge as the highest-ranked languages, with both high IoU and $\rho$, suggesting that systems perform well in both precision and ranking reliability. 
\begin{table}[ht]
\centering
\resizebox{0.97\columnwidth}{!}{
\begin{tabular}{clSSlSS}
    \toprule 
    \rule{0pt}{3ex}
    \multirow{2}{*}{\textbf{Rank}} & \multirow{2}{*}{\textbf{Lang}} & {{\multirow{2}{*}{\textbf{$\overline{\mbox{IoU}}$}}}} & {{\multirow{2}{*}{\textbf{$\bar\rho$}}}} & \multicolumn{3}{c}{\textbf{Top team}} \\
    & & & & \textbf{Name} & {{\textbf{IoU}}} & {{$\rho$}}\\
    \midrule
    1 & IT & 0.51 & 0.46  & UCSC & 0.78 & 0.78 \\
    2 & HI & 0.50 & 0.52  & ccnu & 0.74 & 0.78 \\
    3 & CA & 0.49 & 0.53  & UCSC & 0.72 & 0.77 \\
    4 & FI & 0.48 & 0.39  & UCSC & 0.64 & 0.64 \\
    5 & DE & 0.44 & 0.41  & UCSC & 0.62 & 0.65 \\
    6 & FR & 0.44 & 0.36  & Deloitte & 0.64 & 0.61 \\
    7 & EU & 0.44 & 0.40  & MSA & 0.61 & 0.62 \\
    8 & SV & 0.43 & 0.29  & UCSC & 0.64 & 0.52 \\
    9 & FA & 0.43 & 0.43  & AILSNTUA & 0.71 & 0.69 \\
    10 & AR & 0.42 & 0.40 & MSA & 0.66 & 0.64 \\
    11 & EN & 0.40 & 0.37 & iai\_MSU & 0.65 & 0.62 \\
    12 & ZH & 0.37 & 0.27 & YNU-HPCC & 0.55 & 0.35 \\
    13 & CS & 0.37 & 0.37 & AILSNTUA & 0.54 & 0.55 \\
    14 & ES & 0.31 & 0.33 & ATLANTIS & 0.53 & 0.01 \\
    \bottomrule
    \end{tabular} 
}
\caption{Ranking of the languages based on the mean IoU ($\overline{\mbox{IoU}}$), presenting also the mean $\rho$ ($\bar\rho$) and the top performing team with their scores.}\label{tab:language_rankings}
\end{table}
Conversely, ES, ZH, and CS rank lowest, with ES standing out due to its top-performing system achieving an almost zero $\rho$. This suggests that certain languages pose greater challenges for models, potentially due to dataset properties, linguistic complexity, or limitations in training data.  However, as shown in \Cref{tab:rho_stats}, while the most challenging languages tend to have lower $\bar\rho$ values, the overall rankings indicate that these datasets are not unreliable. 

\begin{figure}[t!]
\centering
\includegraphics[width=\columnwidth, clip,trim=0.2 0.5 0.4cm 1.2cm ]{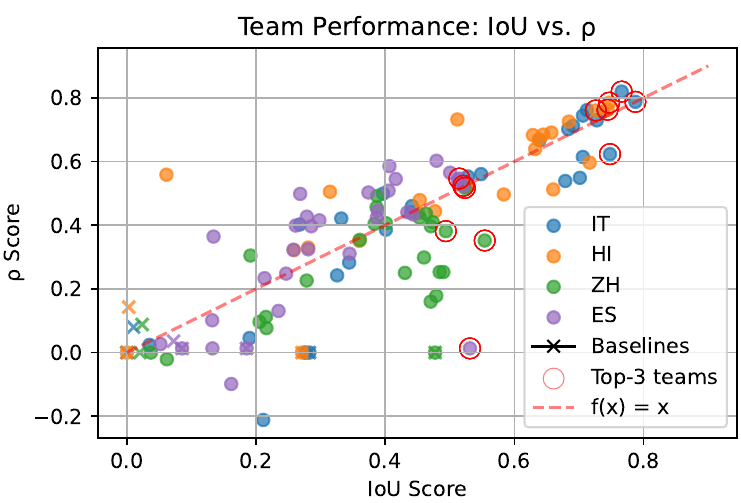}
\caption{Scatter plot of IoU versus $\rho$ scores for all participating teams in the top two and bottom two performing languages, ranked by average IoU scores. 
}\label{fig:scatter_iou-rho}
\end{figure}

\Cref{fig:scatter_iou-rho} further illustrates team performance by scatter-plotting IoU against $\rho$ for teams competing in the top two and bottom two languages from \Cref{tab:language_rankings}. Most high-performing teams (circled in red) cluster in the top-right corner, exhibiting strong results for both metrics, while lower-ranked teams are spread towards the bottom-left. While only a subset of languages is displayed for clarity, we observe similar trends across the full dataset. Notably, IoU scores tend to be higher than $\rho$, as indicated by the majority of points falling below the red dotted line. This highlights  the importance of considering $\rho$ for evaluating ranking consistency.

Some teams show high $\rho$ but low IoU, suggesting they are good at ranking hallucinations but struggle with binary classification.  
\begin{table}[ht!]
\centering 
\resizebox{0.75\columnwidth}{!}{
    \begin{tabular}{>{\bf}lS[table-format=2.2]S[table-format=2.2]S[table-format=2.2]S[table-format=2.2]}
    \toprule
    \textbf{Lang} & \textbf{$\bar\rho$} & \textbf{$\sigma_\rho$} & \textbf{Min ($\rho$)} & \textbf{Max ($\rho$)}  \\
    \midrule
    IT & 0.4681851357142857 & 0.2816421105507776 & -0.211589 & 0.819506 \\
    HI & 0.5284267500000001 & 0.21450288361926106 & 0.0 & 0.78466 \\
    \midrule
    ES & 0.33797787812499996 & 0.2059993142327933 & -0.098563 & 0.602325 \\
    CS & 0.37876343913043475 & 0.14347249597716294 & 0.0924161 & 0.57625 \\
    ZH & 0.277568626923077 & 0.1601157416784085 & -0.020898 & 0.517056 \\
    \bottomrule
    \end{tabular}
    }
    \caption{The mean ($\bar\rho$), standard deviation ($\sigma_\rho$), maximum and  minimum $\rho$ values for the top-2 (IT, HI) and the worse 3 languages: ES, CS, ZH.}\label{tab:rho_stats}
\end{table}
An example from the table is HausaNLP (\citealp{bala-etal-2025-hausanlp}; EN: $\rho=0.42$, IoU$=0.03$), with highly correlated predictions but almost no correct identifications.
When the gap between IoU and $\rho$ is small --- for teams like UCSC and AILSNTUA --- shows the reliability of both metrics not just in raw intersection but in their robustness in ranking, implying that high IoU does not always correlate with high $\rho$.
A big gap in these two metrics when $\rho \ll$ IoU, as we observe for ATLANTIS, indicates that the models are good at making binary decisions but poor at ranking how hallucinated a character is compared to others. Conversely, we observed for teams with IoU $\ll\rho$  that their models can rank characters well in terms of hallucination 
but fail in making the correct binary selections. For instance, HausaNLP shows an extremely low IoU despite a decent $\rho$, meaning its predictions are correlated but far from accurate. Other trends we observe from the general ranking are: TrustAI and Swushroomsia \citep{mitrovic-etal-2025-swushroomsia} present consistent gaps between IoU and $\rho$ in the same ballpark as $\rho=0.54$, IoU$=0.28$; DeepPavlov performs well in CA and EN ($\rho=0.67$, IoU$=0.41$; $\rho=0.61$, IoU$=0.44$) but has significantly lower IoU in ES ($\rho=0.42$, IoU$=0.21$), indicating poor precision; and NLP\_CIMAT \citep{stack-sanchez-etal-2025-nlp_cimat} show highly inconsistent performance across languages (ES: $\rho=0.54$, IoU$=0.47$; FI: $\rho=0.04$, IoU$=0.37$; AR: $\rho=0.09$, IoU$ = 0.14$).

\section{1-UP! Discussion}
\label{sec:discussion}

In order to deepen our understanding of the factors relevant to the success of participating teams within our shared task, we now turn to an analysis of 
metadata collected during the shared task. Participants were asked to fill in a form to describe how their systems worked and what type of resources they used.\footnote{We manually excluded partially filled responses. Metadata was collected for each \emph{submission} rather than for each \emph{system}, i.e., a system may correspond to multiple submissions (e.g., when the system has multilingual capabilities, or when participants tested multiple hyperparametrizations).}
The trends we discuss below are therefore based on self-reporting.
We $z$-normalize performance per language before analysis so as to factor out the varying intrinsic difficulty of the different language datasets.

A first obvious trend in our results is that $\num{36.98030634573304}\%$ of the submissions are reported as prompt-based. The IoU scores of prompt-based submissions are not statistically distinct from the IoU scores of other submissions, but we do find a statistical difference for $\rho$ scores, which are usually lower than in other submissions (Mann-Whitney U test: $p$-value $ < 0.002$, common language effect size: $f=\num{45.96433267587114}\%$). This echoes findings in the previous iteration of the shared task \citep{mickus-etal-2024-semeval}, which pointed out that fine-tuning based approaches were usually more successful on hallucination detection.

Even more prominent is the use of RAG: $\num{52.60393873085339}\%$ of the submissions report using RAG, and are assigned statistically higher IoU scores (Mann-Whitney U, $p$-value $ < 10^{-59}$, $f=\num{69.80313666204218}\%$) and $\rho$ scores ($p$-value $ < 10^{-39}$, $f=\num{66.15924060084531}\%$).
On a related note, if we focus on the data used by participants, we find that  $\num{34.879649890590805}\%$ of submissions which primarily used the data we provided tend to have lower IoU (Mann-Whitney U, $p$-value $ < 10^{-21}$, $f=\num{37.59553635272055}\%$) and $\rho$ scores ($p$-value $ <10^{-22}$, $f=\num{37.28122765478069}\%$).
The preponderance of retrieval-based approaches and their noteworthy success along with the limited performance of submissions relying mainly on the provided data, both showcase that one of the key challenges of the task is finding appropriate references for  assessing LLM outputs.

\begin{table}[!t]
    \centering
\resizebox{0.95\columnwidth}{!}{
    \begin{tabular}{>{\bf}l S[table-format=2.2] l S[table-format=2.2] l S[table-format=2.2]}
\toprule
{\textbf{Model}} & {{\multirow{2}{*}{\textbf{\% subs}}}} & \multicolumn{2}{c}{\textbf{IoU}}  & \multicolumn{2}{c}{$\mathbf{\rho}$} \\
\textbf{family} && {{$p$-val.}} & {{$f~(\%)$}} & {{$p$-val.}} & {{$f~(\%)$}}\\
\midrule
BERT & 12.12253829321663 & {{$< 10^{-4}$}} & 42.46857335998964 & \num{0.08753746555870451} & {{---}} \\
Claude & 3.019693654266958 & {{$< 10^{-5}$}} & 65.87139643174802 & {{$< 10^{-14}$}} & 78.23143933448439 \\
DeepSeek & 2.3194748358862145 & {{$< 10^{-11}$}} & 77.86865151822548 & {{$< 10^{-13}$}} & 80.15359775478461 \\
Flan-T5 & 5.776805251641138 & {{$< 10^{-18}$}} & 26.614378808990978 & {{$< 10^{-13}$}} & 30.279103154161213 \\
GPT & 16.01750547045952 & {{$< 10^{-7}$}} & 59.114634500551 & {{$< 10^{-2}$}} & 55.07165902094955 \\
Llama & 12.341356673960613 & {{$< 10^{-15}$}} & 35.167992691813346 & {{$< 10^{-29}$}} & 29.08137793310035 \\
Qwen & 18.03063457330416 & {{$< 10^{-16}$}} & 63.06338411457658 & {{$< 10^{-21}$}} & 65.27454527547832 \\
XLM-R & 10.328227571115974 & \num{0.011050827498649152} & 44.956303612345 & {{$< 10^{-2}$}} & 44.354315044130665 \\
\bottomrule
\end{tabular}
}
    \caption{Overview of main PLM families used by participating teams, proportion of relevant submissions, and their effects on scores (Mann-Whitney U tests comparing the scores of submissions using PLMs of the given model family vs. other submissions, along with common-language effect size $f$ where significant). }
    \label{tab:u-test per llm}
\end{table}

Another factor of interest is whether specific PLMs stand out as more or less appropriate for the task of detecting hallucinated spans.
In \Cref{tab:u-test per llm}, we provide an overview of the PLMs most frequently used by participants to tackle the shared task, along with the results of U tests comparing scores assigned to submissions using this PLM vs. submissions not relying on it.
This allows us to get insights regarding which PLMs tended to yield comparatively higher scores.
Given the large number of models, we group them by family, i.e., the GPT family contains GPT-3, GPT-3.5, GPT-4 and other variants, while some other families include multilingual variants (e.g., Flan-T5 includes MT5).
The BERT family is used as a catch-all for large group of language-specific models (e.g., CamemBERT), and smaller encoder-based PLMs (e.g., ALBERT or DeBERTa).
Several submissions mentioned multiple PLMs and a handful mentioned using none. For the sake of clarity, we do not include  PLMs that were only used in a small minority (<1\%) of submissions.
Overall, we find that Llama-based, Flan-T5 and BERT-based systems tended to perform less well than other systems. 
The DeepSeek family appears to be highly competitive, as there is a $78\%$ chance that any DeepSeek-based submission will outrank a randomly selected non-DeepSeek-based submission.
Here as well, $\rho$ and IoU performances appear roughly in line with one another.

\begin{table}[!t]
    \centering
\resizebox{0.8\columnwidth}{!}{
    \begin{tabular}{>{\bf}l *{2}{l S}}
        \toprule
        \multirow{2}{*}{\textbf{Lang.}} & \multicolumn{2}{c}{\textbf{IoU}} &  \multicolumn{2}{c}{\textbf{$\rho$}} \\
        & {{\textbf{$p$-val.}}} & {{\textbf{correl.}}} & {{\textbf{$p$-val.}}} & {{\textbf{correl.}}} \\
        \midrule
        AR & {{$< 10^{-323}$}} & 0.24042701037402298 & {{$< 10^{-323}$}} & 0.24384791508821863 \\
        CA & {{$< 10^{-207}$}} & 0.2593638512317918 & {{$< 10^{-60}$}} & 0.1407653202676284 \\
        CS & {{$< 10^{-156}$}} & 0.22218884886000528 & {{$< 10^{-56}$}} & 0.13416973082568331 \\
        DE & {{$< 10^{-323}$}} & 0.24963963325681002 & {{$< 10^{-131}$}} & 0.14523184054449714 \\
        EN & {{$< 10^{-323}$}} & 0.26236381911804063 & {{$< 10^{-138}$}} & 0.09516883180582052 \\
        ES & {{$< 10^{-323}$}} & 0.34754592131748724 & {{$< 10^{-323}$}} & 0.2676589896818773 \\
        EU & {{$< 10^{-281}$}} & 0.29506955375906446 & {{$< 10^{-92}$}} & 0.17133481693680858 \\
        FA & {{$< 10^{-115}$}} & 0.19749195456608384 & {{$< 10^{-99}$}} & 0.1835447255880333 \\
        FI & {{$< 10^{-323}$}} & 0.25786737858721287 & {{$< 10^{-141}$}} & 0.15670020619067604 \\
        FR & {{$< 10^{-296}$}} & 0.2062110051161137 & {{$< 10^{-16}$}} & 0.04820779047653345 \\
        HI & {{$< 10^{-246}$}} & 0.2255948823787385 & {{$< 10^{-77}$}} & 0.12716693442496066 \\
        IT & {{$< 10^{-168}$}} & 0.1633631431535463 & {{$< 10^{-167}$}} & 0.16317634143559817 \\
        SV & {{$< 10^{-179}$}} & 0.18692837942204646 & {{$< 10^{-8}$}} & 0.039891456486058754 \\
        ZH & {{$< 10^{-323}$}} & 0.3656591796898794 & {{$< 10^{-10}$}} & 0.04401693585004858 \\
        \bottomrule
    \end{tabular}
}
    \caption{Spearman correlation of inter-annotator agreement (\Cref{eq:agg}) vs. datapoint-level scores.}
    \label{tab:effect of agreement on scores}
\end{table}

The last factor we explore is related to our earlier observations regarding inter-annotator agreement (reported in \Cref{sec:data}, \Cref{tab:agreement}). 
We would expect different levels of inter-annotator agreement across languages 
to impact performance. 
This line of thought should also apply at the datapoint level: Items where annotations are less consensual, as per \Cref{eq:agg}, might lead to lower scores. 
We explicitly evaluate this by computing the Spearman correlation between the inter-annotator agreement metric and the scores assigned to a given datapoint. The results are summarized in \Cref{tab:effect of agreement on scores}.
We observe low to moderate correlations across all setups.  In other words, while annotator agreement rates do impact the success of a model, other factors of variation still play an important role. 

\section{The Princess is in another article: Conclusions}

The Mu-SHROOM multilingual shared-task was an overall success.
We received 2,618 submissions from 43 teams, including a handful of participants from the first iteration of the SHROOM shared task. Whilst the level of participation varied by language, over 20 teams competed in each of the 14 languages.
Participating teams deployed a vast array of methodologies, ranging from QA-- or NER--based pretraining to synthetic data generation and RAG approaches, which will serve as starting points for future research. 
We also observed a high number of student-lead teams. One of the goals of the shared task is to lower the barrier to entry to current challenges in NLP, hence we take the interest of students as a further indicator of success.

Beyond these participation numbers, the data collected for Mu-SHROOM also allowed us to highlight a number of often-overlooked points in the literature.
The prevalence and severity of hallucinated outputs varies across languages (see \Cref{tab:error rate per type}); for some languages, we in fact observe fluency to be a more pressing challenge for LLMs than factuality.
The metadata collected from participants' submissions (see Section \ref{sec:discussion}) also allowed us to highlight some of the challenges underpinning hallucination detection. The ability to retrieve accurate references matters, but so do the base pretrained LM used by participants and (to a lesser extent) the agreement rates of annotators.
Regarding this latter point, it is worth stressing that we find genuine disagreement among our annotators as to where a hallucination begins and ends.

If Mu-SHROOM has allowed us to establish the importance of multilingual data for hallucination detection, much remains to be done in order to fully assess 
LLM technologies' tendency to produce non-factual information. 
One other aspect we have left outside the scope of this shared task is that of mitigating hallucinations, 
a step that is however necessary and complementary to our endeavors.
We have constructed the present shared task as a means to draw the attention of the community towards some challenges tied to hallucination detection --- and attention is indeed needed, given that even top-scoring teams do not detect 20\% or more of the hallucination spans.

\section*{The Boo's we avoid: Limitations and Ethical considerations}
We strive to uphold the principles outlined in the \href{https://www.aclweb.org/portal/content/acl-code-ethics}{ACL Code of Ethics}.

\paragraph{Terminology.} One important limitation of our work is the terminology surrounding hallucinations in AI-generated text. \citet{Hicks2024} argue that this metaphor can be misleading, implying that AI models perceive information incorrectly rather than simply generating outputs based on probabilistic patterns without any underlying understanding or intent. This framing may contribute to misconceptions among policymakers, investors, and the general public, shaping unrealistic expectations about AI systems’ capabilities and failures. While we use the term hallucination in this work due to its established presence in the literature, we acknowledge its limitations and the broader implications of language in shaping discussions around AI reliability.
\paragraph{Broader Impact.}
Hallucinated outputs from large language models pose a significant risk, as they can be exploited to propagate disinformation and reinforce misleading narratives. Detecting such outputs is a critical step toward understanding the underlying causes of this phenomenon and contributing to ongoing efforts to mitigate hallucinations. By addressing this challenge, we aim to support the development of more reliable and trustworthy generative language models.

\paragraph{Data and Annotators.}
The dataset we release may contain false or misleading statements, reflecting the nature of the task. While annotated portions of the data are explicitly labeled as such, unannotated portions may include unverified or inaccurate content.
To ensure a respectful and safe annotation process, we manually pre-filtered the data provided to annotators, removing profanities and other objectionable material. However, the unannotated portion of the dataset has not undergone the same level of scrutiny and may include offensive, obscene, or otherwise inappropriate content.

\section*{Acknowledgments}
The construction of the Mu-SHROOM dataset was made possible thanks to a grant from the Otto Malm foundation.
This work was also supported by the ICT 2023 project ``Uncertainty-aware neural language models'' funded by the Academy of Finland (grant agreement  \textnumero{}~345999). Liane Guillou was funded by UK Research and Innovation (UKRI) under the UK government’s Horizon Europe funding guarantee [grant number 10039436 (Utter)] whilst working at the University of Edinburgh.
Alessandro Raganato thanks Morteza Ghorbaniparvariji for their contribution to the Farsi annotations.
Raúl Vázquez and Timothee Mickus thank Malvina Nissim, along with the students in her ``Shared Task'' class at the University of Groningen --- her commitment to this course is one of the reasons that Mu-SHROOM was able to reach a broad audience. 
Timothee Mickus also thanks the volunteers for testing the annotation interface, especially Jean-Michel Jézéquel.

\bibliography{anthology,custom,systems}

\appendix

\section{The super Mu-SHROOM party jamboree: Organizers' roles}
Our long line of mushroom friendly people behind this edition of the SHROOM Shared task are as follows:

\orgname{Raúl Vázquez}:  Grant application writing \& accounting, Spanish data creation \& selection, datapoint creation guidelines, annotation guidelines, annotator recruitment \& briefing sessions, annotator training, advertisement, overall leadership, paper writing, reviewing process.

\orgname{Timothee Mickus}: Websites development, French validation data creation \& selection, English data creation \& selection, German data creation, datapoint creation guidelines, annotation guidelines, annotator recruitment \& briefing sessions, data analysis, advertisement, overall leadership, paper writing, reviewing process.

\orgname{Elaine Zosa}: Baseline system development.

\orgname{Teemu Vahtola}: Finnish data creation \& selection.

\orgname{Jörg Tiedemann}: German data selection, advertisement.

\orgname{Aman Sinha}: Hindi data creation \& selection, advertisement, annotator recruitment, reviewing process.

\orgname{Vincent Segonne}: French test data creation \& selection.

\orgname{Fernando Sánchez-Vega}: Spanish data annotator recruitment, advertisement.

\orgname{Alessandro Raganato}: Italian \& Farsi data creation \& selection, annotation guidelines, annotator recruitment, advertisement.

\orgname{Jind\v{r}ich Libovický}: Czech data creation \& selection, data analysis.

\orgname{Jussi Karlgren}: Swedish data creation \& selection.

\orgname{Shaoxiong Ji}: Chinese data creation \& selection, advertisement, reviewing process.

\orgname{Jind\v{r}ich Helcl}: Czech data creation \& selection, data analysis.

\orgname{Liane Guillou}: English data selection, lead role for annotation guidelines development, paper writing.

\orgname{Ona de Gibert}: Catalan data creation \& selection, advertisement, reviewing process.

\orgname{Jaione Bengoetxea}: Basque data creation \& selection, advertisement.

\orgname{Joseph Attieh}: Arabic data creation \& selection, advertisement.

\orgname{Marianna Apidianaki}: Annotator recruitment, paper writing.

\section{The map of the Mu-SHROOM kingdom: Supplementary information on dataset creation}
\subsection{Dataset details}

\begin{table*}[!th]
    \centering
\resizebox{!}{0.425\textheight}{
    \begin{tabular}{l >{\tt\arraybackslash}l l S[round-precision=0, table-format=2.0] S[round-precision=0, table-format=3.0]}
\toprule
         \textbf{Lang.} & \textrm{\textbf{HF identifier}} & \textbf{Publication}  & {{\textbf{N. val.}}} & {{\textbf{N. test}}}\\
\midrule\multirow{3}{*}{AR}
& SeaLLMs/SeaLLM-7B-v2.5 & \citet{damonlpsg2023seallm} & 17 & 86 \\
& arcee-ai/Arcee-Spark &  {{---}}  & 12 & 13\\
& openchat/openchat-3.5-0106-gemma & \citet{wang2023openchat} & 21 & 51 \\
\midrule\multirow{3}{*}{CA}
& meta-llama/Meta-Llama-3-8B-Instruct & \citet{grattafiori2024llama3herdmodels} & {{---}} &27 \\
& mistralai/Mistral-7B-Instruct-v0.3 & --- & {{---}} & 34 \\
& occiglot/occiglot-7b-es-en-instruct & --- & {{---}} & 39 \\
\midrule\multirow{2}{*}{CS}
& meta-llama/Meta-Llama-3-8B-Instruct & \citet{grattafiori2024llama3herdmodels} & {{---}} & 56 \\
& mistralai/Mistral-7B-Instruct-v0.3 & --- & {{---}} & 44 \\
\midrule\multirow{3}{*}{DE}
& TheBloke/SauerkrautLM-7B-v1-GGUF & --- & 7 & 28 \\
& malteos/bloom-6b4-clp-german-oasst-v0.1 & \citet{ostendorff2023efficientlanguagemodeltraining} & 27 & 75 \\
& occiglot/occiglot-7b-de-en-instruct & {{---}} & 16 & 47 \\
\midrule\multirow{3}{*}{EN}
& TheBloke/Mistral-7B-Instruct-v0.2-GGUF & --- & 19 & 53 \\
& tiiuae/falcon-7b-instruct & \citet{falcon40b} & 15 & 47 \\
& togethercomputer/Pythia-Chat-Base-7B & --- & 16 & 54 \\
\midrule\multirow{3}{*}{ES}
& Iker/Llama-3-Instruct-Neurona-8b-v2 &  --- & 12 & 45 \\
& Qwen/Qwen2-7B-Instruct & \citet{yang2024qwen2technicalreport} & 18 & 62\\
& meta-llama/Meta-Llama-3-8B-Instruct  & \citet{grattafiori2024llama3herdmodels} & 20 & 45 \\
\midrule\multirow{2}{*}{EU}
& google/gemma-7b-it & --- & {{---}} & 23 \\
& meta-llama/Meta-Llama-3-8B-Instruct  & \citet{grattafiori2024llama3herdmodels} & {{---}} & 76 \\
\midrule\multirow{6}{*}{FA}
& CohereForAI/aya-23-35B & \citet{aryabumi2024aya23openweight} & {{---}} & 10 \\
& CohereForAI/aya-23-8B & \citet{aryabumi2024aya23openweight} & {{---}} & 7 \\
& Qwen/Qwen2.5-7B-Instruct & \citet{yang2024qwen2technicalreport} & {{---}} & 1 \\
& meta-llama/Llama-3.2-3B-Instruct & --- & {{---}} & 20 \\
& meta-llama/Meta-Llama-3.1-8B-Instruct  & \citet{grattafiori2024llama3herdmodels} & {{---}} & 24 \\
& universitytehran/PersianMind-v1.0 & \citet{rostami2024persianmindcrosslingualpersianenglishlarge} & {{---}} & 38 \\
\midrule\multirow{2}{*}{FI}
& Finnish-NLP/llama-7b-finnish-instruct-v0.2 & --- & 25 & 84 \\
& LumiOpen/Poro-34B-chat & \citet{luukkonen2024poro} & 25 & 66\\
\midrule\multirow{5}{*}{FR}
& bofenghuang/vigogne-2-13b-chat & --- & 15 & 35\\
& croissantllm/CroissantLLMChat-v0.1 & \citet{faysse2024croissantllm} & 8 & 49 \\
& meta-llama/Meta-Llama-3.1-8B-Instruct & \citet{grattafiori2024llama3herdmodels} & 8 & 10 \\
& mistralai/Mistral-Nemo-Instruct-2407 & --- & 10 & 26\\
& occiglot/occiglot-7b-eu5-instruct & --- & 9 & 30 \\
\midrule\multirow{3}{*}{HI}
& meta-llama/Meta-Llama-3-8B-Instruct  & \citet{grattafiori2024llama3herdmodels} & 4 & 7 \\
& nickmalhotra/ProjectIndus & \cite{malhotra2024projectindus} & 44 & 128 \\
& sarvamai/OpenHathi-7B-Hi-v0.1-Base & --- & 2 & 15\\
\midrule\multirow{4}{*}{IT}
& Qwen/Qwen2-7B-Instruct &  \citet{yang2024qwen2technicalreport} & 14 & 35 \\
& meta-llama/Meta-Llama-3.1-8B-Instruct & \citet{grattafiori2024llama3herdmodels} & 6 & 11\\
& rstless-research/DanteLLM-7B-Instruct-Italian-v0.1 & --- & 2 & 14\\
& sapienzanlp/modello-italia-9b & --- & 28 & 90\\
\midrule\multirow{3}{*}{SV}
& AI-Sweden-Models/gpt-sw3-6.7b-v2-instruct-gguf & --- & 29 & 112\\
& LumiOpen/Poro-34B-chat & \citet{luukkonen2024poro} & 16 & 28\\
& LumiOpen/Viking-33B & --- & 4 & 7\\
\midrule\multirow{5}{*}{ZH}
& 01-ai/Yi-1.5-9B-Chat & \citet{ai2024yiopenfoundationmodels} & 8 & 24\\
& Qwen/Qwen1.5-14B-Chat & \citet{qwen} & 10 & 27\\
& THUDM/chatglm3-6b & \citet{glm2024chatglm} & 0 & 1 \\
& baichuan-inc/Baichuan2-13B-Chat & --- & 25 & 68\\
& internlm/internlm2-chat-7b & \citet{cai2024internlm2} & 7 & 30\\
\bottomrule
    \end{tabular}}
    \caption{LLMs considered for each language. N. val.: corresponding number of datapoints in val; N. test: corresponding number of datapoints in test.}
    \label{tab:llms}
\end{table*}

In \Cref{tab:llms}, we provide an overview of the models used for every language in the shared task.
There are a total of 38 different LLMs, all available through the HuggingFace platform.\footnote{\href{https://huggingface.co/}{\tt huggingface.co}}
In practice, a number of these models correspond to variants of the same base model or family, including language-specific fine-tuned versions, incremental releases, or models with different parameter counts from the same model family.
It is worth stressing that the models themselves are not balanced: for instance, over $85\%$ of the Hindi test set correspond to a single model (viz. \texttt{nickmalhotra/ProjectIndus}).


\subsection{Annotation guidelines}\label{appx:annotation_guidelies}
\begin{figure*}[!th]
{\smaller
    \textbf{Example}

    \vspace{-0.5em}

    \begin{center}
        \fbox{%
            \parbox{\textwidth}{%
                \textbf{Question:} During which centuries did William II of Angoulême live?

                \textbf{Answer:} William II, also known as Guillaume II or "William the Good," was a French nobleman who lived from around 1099 to 1137. He was Count of Angoulême and Poitou from 1104 until his death in 1137. Therefore, William II lived during the 11th and 12th centuries.
            }%
        }
    \end{center}

    \begin{center}
        \includegraphics[]{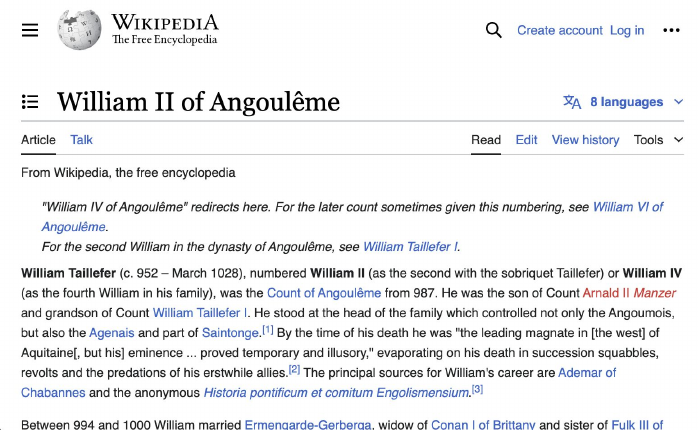}
    \end{center}

    \vspace{1em}

    \textbf{Annotated Example}\\
    In the \textbf{William II of Angoulême} example, we find that the answer text contains information that is not present in the Wikipedia article.

    \begin{center}
        \fbox{%
            \parbox{\textwidth}{%
                \textbf{Question:} During which centuries did William II of Angoulême live?

                \textbf{Answer:} William II, also known as Guillaume II or "William the Good," was a French nobleman who lived from around 1099 to 1137. He was Count of Angoulême and Poitou from 1104 until his death in 1137. Therefore, William II lived during the 11th and 12th centuries.
            }%
        }
    \end{center}

    \vspace{0.5em}
    
    We therefore annotate the example as follows:\\

    William II , also known as \textbf{\textcolor{red}{Guillaume II}} or `` \textbf{\textcolor{red}{William the Good}} , '' was a French nobleman who lived from around \textbf{\textcolor{red}{1099 to 1137}} . He was Count of Angoulême and \textbf{\textcolor{red}{Poitou from 1104 until his death in 1137}} . Therefore , William II lived during the \textbf{\textcolor{red}{11th and 12th}} centuries . \\

    (Explanation: in the text above, spans highlighted in bolded red text are overgenerations / hallucinations as the information that they contain is not supported by the Wikipedia article)

}   
    \caption{Annotation guidelines: Illustrative example.}
    \label{fig:guidelines_example}
\end{figure*}

In Figures~\ref{fig:guidelines_instructions} and \ref{fig:guidelines_example}, we provide an exact copy of the annotation guidelines and the illustrative example given to the annotators.
These guidelines are based on five of the organizers' experience of annotating the trial set, and were provided to annotators recruited for the validation and test splits.
For all languages except EN and ZH, we also organized a briefing session for annotators so as to ensure the guidelines were properly understood and that participants were aware of existing communication channels through which they could ask for clarifications.

\begin{figure*}[!th]
{\smaller
    \begin{center}
    \textbf{Mu-SHROOM Annotation Guidelines}
    \end{center}

    \textbf{Introduction}\\
    In this annotation project you will be shown a series of question-answer pairs plus a relevant Wikipedia article. The answer will be a passage of text produced by a Large Language Model (LLM) in response to the question. You will be asked to identify, with respect to the Wikipedia article: which tokens in the answer constitute the overgeneration or ``hallucination''.\\

    \textbf{Annotation Guidelines}
    \vspace{-0.5em}
    \begin{enumerate}
        \setlength\itemsep{-0.3em}
        \item Carefully read the answer text.
        \item Highlight each span of text in the answer text that is not supported by the information present in the Wikipedia article (i.e. contains an overgeneration or hallucination). Your annotations should include only the \textbf{minimum} number of characters* in the text that should be edited/deleted in order to provide a correct answer (*in the case of Chinese, these will be ``character components''). As a general ``rule of thumb'' you are encouraged to annotate \textbf{\textit{conservatively}} and to focus on \textbf{\textit{content words}} rather than \textbf{\textit{function words}}. Please note that this is not a strict guideline, and you should rely on your best judgements when annotating examples.
        \vspace{-0.3em}
        \begin{itemize}
            \item In the annotation platform: To highlight a span of one or more characters in the text, click on the first character and drag the mouse to the last character - it will change to red text. To remove highlighting, click anywhere on the highlighted red span - it will revert to black text.
        \end{itemize}
        \vspace{-0.3em}
        \item If the answer text does not contain a hallucination, write ``NO HALLUCINATION'' in the comment box.
        \item If you are unsure about how to annotate an example, write ``UNSURE'' in the comment box. Please only use this option as a last resort.
        \item Ensure that you double-check your annotations prior to moving to the next example. From the ``See previous annotations'' link you can \textbf{\textit{edit}} or \textbf{\textit{delete}} previous annotations.
    \end{enumerate}
    
    Note:
    \vspace{-0.5em}
    \begin{itemize}
        \setlength\itemsep{-0.3em}
        \item You should not consult any other sources of information, e.g. web searches, other web pages, or your own knowledge. Use only Wikipedia as your source.
        \item Ideally, the wikipedia entry provided should suffice for the annotation, however, you are allowed/encouraged to browse other Wikipedia articles to verify information that is not contained in the provided article.
        \item If you do consult other Wikipedia articles, please add a comment to this effect and include links to the articles that include information that informed your annotation.
        \item You are encouraged to leave comments where relevant e.g. if the annotation of an example is not straightforward, or if there is anything else you wish to bring to our attention
        \item You are encouraged to \textbf{\textit{review}} your annotations prior to finishing the task
    \end{itemize}
        
    \textbf{Hallucination Definition}\\
    Hallucination: content that contains or describes facts that are not supported by the provided reference. In other words: hallucinations are cases where the answer text is more specific than it should be, given the information available in the Wikipedia page.\\

    \textbf{Content/Function Word Definition}\\
    Content words contribute to the meaning of the sentence in which they occur. Nouns (Barack Obama, cake, cat etc.), main verbs (eat, run, think etc.), adjectives (small, red, angry etc.) and adverbs (quickly, loudly etc.) are usually content words.\\

    Function words are structural and typically have very little substantive meaning. Auxiliary verbs (could, must, need, will etc.), articles (a, an, the etc.), prepositions (in, out, under etc.), and conjunctions (and, but, till, as etc.) are usually function words.
}

    \caption{Annotation guidelines: Instructions.}
    \label{fig:guidelines_instructions}
\end{figure*}

\begin{figure*}[!bh]
     \centering
   \includegraphics[angle=90, max height=0.85\textheight, max width=0.85\textwidth, trim= 0cm 5cm 0cm 5cm]{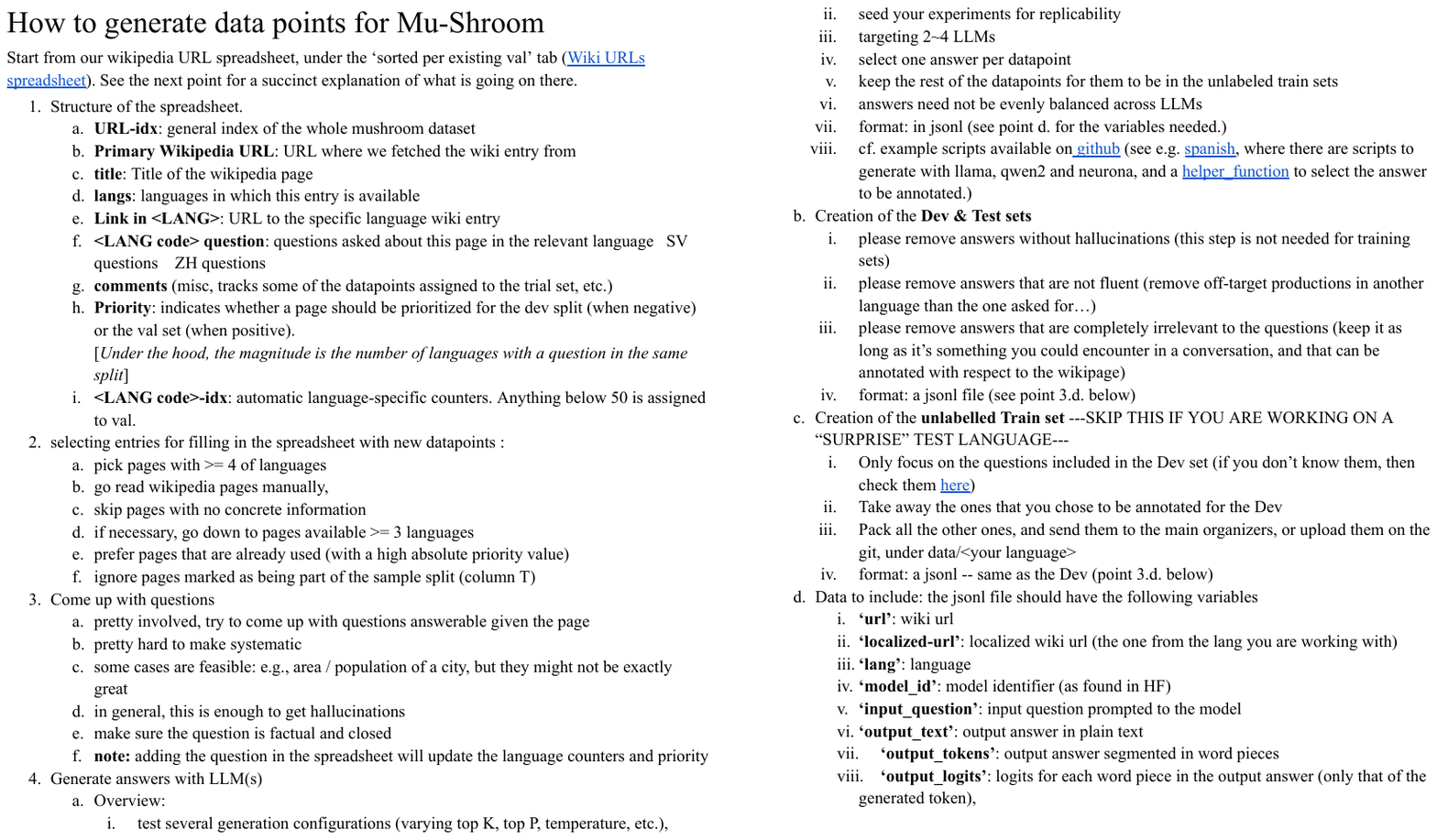}   \caption{Datapoint creation guidelines.}  

    \label{fig:data_guidelines}
\end{figure*}


\subsection{Datapoint creation guidelines}\label{appx:datacreation_guidelies}
In \Cref{fig:data_guidelines}, we provide an exact copy of the annotation guidelines given to the organizers in charge of each language.

 \subsection{Departures from the general guidelines}\label{appx:data nonstandard}
In practice, some ad-hoc modifications to the data creation process were adopted, depending on the challenges intrinsic to individual languages.
We list the exceptions to these rules for each language below, and the available means for annotation: 


\begin{itemize}
    
\item\textbf{CS:} The Czech split was built from Wikipedia pages with no equivalent in other languages. 


\item\textbf{EN:} The dataset was annotated with a large pool of annotators that individually annotated about 20 datapoints. In total, some datapoints were annotated by up to 12 annotators. 

\item\textbf{ES:} The test split was annotated by 6 annotators; the first release of the validation split contained only 3 annotations, which was increased to 6 in the final data released.







\item\textbf{SV:} Due to replicability concerns, a handful of datapoints were removed. One of the SV models is not instruction-tuned. 

\item\textbf{ZH:} The dataset was annotated with a large pool of annotators that individually annotated about 20 datapoints. In total, some datapoints were annotated by up to 6 annotators. A subset of items correspond to the same questions, with answers from different LLMs (or different settings). 

\end{itemize}

\FloatBarrier
\clearpage

\section{The Lost Levels: detailed rankings}
\label{appx:rankings}
\subsection{Official $\mathrm{IoU}$-based rankings}

In Table~\ref{tab:rankings}, we provide detailed rankings across all languages.
We also include the probability $\Pr(\mathrm{rank})$ of any given submission outranking the submission one rank below, which we compute through random permutation: 
We re-sample with replacement the datapoints in both submissions $100~000$ times, and then compute the proportion of samples where the higher-ranking submission still outperforms the lower-ranking submission, based on IoU scores.
For instance, team MSA (ranked 1\textsuperscript{st} on Arabic) outranks team UCSC (ranked 2\textsuperscript{nd} on Arabic) in $65.24\%$ of the random samples we perform, suggesting that the advantage of team MSA's approach is in part contingent on the test data.
More broadly, this bootstrapping approach reveals that the rankings are not stable --- in most case, we find the probability of a lower-ranking submission outranking the next best submission under resampling to be greater than $1 - \Pr > 0.05$, i.e., we find limited statistical evidence that performances are significantly better within higher ranked submissions.

\input{tables/iou-results/results-all}

\subsection{Alternative $\rho$-based rankings}
\label{appx:alt rankings}

In Table~\ref{tab:alt rankings}, we provide  alternative rankings of participating teams based on their best $\rho$ submission.
We also include the probability $\Pr(\mathrm{rank})$ of a $\rho$-based ranking being stable, which as previously we compute through bootstrapping. Here again, we find that stable rankings (where $\Pr(\mathrm{rank})>0.95$) are the exception and not the norm.

One key observation to be stressed is that the rankings are significantly impacted by the metric we use.

\input{tables/cor-results/results-all}

\end{document}

%% file: tables/participant_info/team_alphaorder.tex
\rowcolors{2}{gray!25}{white}
\begin{tabular}{p{.35\linewidth} p{.25\linewidth} p{0.9\linewidth}}
    \toprule
    \bf Team \& \bf Paper & \bf Languages & \bf Overview \\
    \midrule
Advacheck \citep{voznyuk-etal-2025-advacheck} & EN & NER-based keyword extraction, Wikipedia-based RAG, LLM edition-based prompting. \\
AILSNTUA \citep{karkani-etal-2025-ails} & All & Translate-test (to EN and ZH) prompt-based approaches using synthetic few-shot examples. \\
ATLANTIS \citep{kobus-etal-2025-atlantis} & DE, EN, ES, FR & RAG + LLM prompting; RAG-based approaches; token-level classifiers. \\
BlueToad \citep{pronk-etal-2025-bluetoad} & AR, CS, DE, EN, ES, EU, FA, FI, FR, HI, IT, SV, ZH & QA-finetuned base PLMs; fine-tuning on synthetic data \\
CCNU \citep{liu-chen-2025-ccnu} & All & Prompting \& RAG \\
COGUMELO \citep{creo-etal-2025-cogumelo} & EN, ES & NER-finetuning; perplexity-based assessments \\
CUET\_SSTM & AR & NER-finetuning. \\
Deloitte \citep{chandler-etal-2025-deloitte} & All & Binary token-level classifiers, trained using web-search results, task instruction and datapoint as inputs. \\
DeepPavlov & All & White-box approaches \\
DUTJBD \citep{yin-etal-2025-dutjbd} & EN & --- \\
FENJI \citep{alberts-etal-2025-fenji} & All & Dense passage retrieval for Flan-T5 prompting. \\
FiRC-NLP \citep{tufa-etal-2025-firc} & All & Prompt-based approaches, incorporating external references. \\
FunghiFunghi \citep{ballout-etal-2025-funghifunghi} & EN, ES, FR, IT, SV & Translate-train (to EN) and synthetic datasets. \\
GIL-IIMAS UNAM \citep{lopez-ponce-etal-2025-gil} & EN, ES & Wikipedia-based RAG. \\
HalluRAG-RUG \citep{abdi-etal-2025-hallurag} & EN & Wikipedia-based RAG, followed by a summarization step and a zero-shot prompting to annotate the items. \\
HalluSearch \citep{abdallah-el-beltagy-2025-hallusearch} & All & Factual statement decomposition and verification through real-world context retrieval. \\
HalluciSeekers & AR, DE, EN, ES, FA, FR, IT, SV & --- \\
Hallucination Detectives \citep{elchafei-abu-elkheir-2025-hallucination} & AR, EN & Semantic role labeling, dependency parsing, and token-logit confidence scores to construct spans \\
HausaNLP \citep{bala-etal-2025-hausanlp} & EN & Finetuning approaches. \\
Howard University - AI4PC \citep{aryal-akomoize-2025-howard} & All & Time-series anomaly detection across the sequence of logits. \\
iai\_MSU \citep{pukemo-etal-2025-iai_msu} & EN & RAG \\
keepitsimple \citep{vemula-krishnamurthy-2025-keepitsimple} & All & Multiple LLM generated responses are compared with model output text by modeling information entropy for detecting uncertainty. \\
LCTeam \citep{maldonado-rodríguez-etal-2025-lcteam} & All & Label transfer via translate-train (to CA, CS, ES, FR, IT, ZH, \& between phylogenetically related languages); Wikipedia-based RAG and summarization approaches. \\
MALTO \citep{savelli-etal-2025-malto} & EN & Logits of a larger model are used to assess the truthfulness of the sentence predicted by the single smaller model. \\
MSA \citep{hikal-etal-2025-msa} & All & Weak supervised fine-tuning approaches \\
NCL-UoR \citep{hong-etal-2025-ncl} & All & Keyword extraction and Wikipedia-based retrieval, detection using closed-source APIs, post-processing with non-linear probability optimization or stochastic prompt-based labeling. \\
NLP\_CIMAT \citep{stack-sanchez-etal-2025-nlp_cimat} & AR, CA, CS, EN, ES, EU, FA, FI, FR, IT, SV & MLP-based classifiers probing the hidden layers of a Llama 3.1 model; few shot inference with chatGPT3.5-turbo using Wikipedia contexts. \\
nsu-ai & All & prompt based approaches \\
RaggedyFive \citep{heerema-etal-2025-raggedyfive} & EN & RAG + NLI across trigrams in LLM answers. \\
REFIND \citep{lee-yu-2025-refind} & AR, CS, DE, EN, ES, EU, FI, FR, IT & Context sensitivity-based token-level identification matched against externally retrieved documents; FAVA-based pipeline. \\
S1mT5v-FMI & DE, ES, FI, FR, SV, ZH & --- \\
SmurfCat \citep{rykov-etal-2025-smurfcat} & All & Qwen-based approach, deriving continuous annotation through repeated sampling. \\
Swushroomsia \citep{mitrovic-etal-2025-swushroomsia} & AR, DE, EN, ES, FI, FR, HI, IT, SV, ZH & Prompting-based approach \\
Team Cantharellus \citep{mo-etal-2025-team} & AR, CA, CS, DE, EN, ES, EU, FA, FI, FR, HI, IT, ZH & Prompting-based approach (GPT-4o-mini) to find hallucinated words/parts of text in each datapoint; fine-tuning on synthetic data. \\
TrustAI & AR, DE, EN, ES, FI, FR, HI, IT, SV, ZH & Variations on the neural baseline \\
tsotsalab & All & GPT-4 finetuning; counterfactual comparisons with external references. \\
TU Munich & AR, DE, EN, ES, FI, FR, HI, IT, SV, ZH & Synthetic data generation (MKQA-based). \\
TUM-MiKaNi \citep{anschutz-etal-2025-tum} & All & Wikipedia-based retrieval used as input for prompting-based approaches; BERT-based regression. \\
UCSC \citep{huang-etal-2025-ucsc} & All & Elaborate prompting approaches (CoT, few-shot reasoning); pre-translation (to EN) before RAG-based prompting; token masking-based approaches. \\
uir-cis \citep{huang-etal-2025-uir} & All & Comparison of extracted triples to external references. \\
UMUTeam \citep{pan-etal-2025-umuteam} & All & Classifier-based, compare outputs to be annotated with those from larger LLMs. \\
UZH \citep{wastl-etal-2025-uzh} & All & Prompting to generate a set of answers, using either an external model (GPT-4o-mini) or the model that produced the datapoint, followed by a embedding similarity--based detection step to mark counterfactual spans. \\
VerbaNexAI \citep{morillo-etal-2025-verbanexai} & EN & Retrieval-based approaches \\
YNU-HPCC \citep{chen-etal-2025-ynu} & EN, ZH & Prompting, RAG; MRC. \\
\bottomrule
\end{tabular}

%% file: tables/iou-results/results-all.tex
\begin{center}
\tablefirsthead{\toprule 
\textbf{Lang} & \textbf{Team} & \textbf{IoU} & $\rho$ & $\Pr(\mathrm{rank})$ \\ \midrule}
\tablehead{%
\multicolumn{4}{c}%
{{ \smaller (\textit{Continued from previous column})}} \\
\toprule
\textbf{Lang} & \textbf{Team} & \textbf{IoU} & $\rho$ & $\Pr(\mathrm{rank})$ \\ \midrule }
\tabletail{%
\midrule \multicolumn{4}{c}{{\smaller (\textit{Continued on next column})}} \\}
\tablelasttail{%
\bottomrule
}
\bottomcaption{\label{tab:rankings} Official rankings, all languages, all teams. Column $\Pr(\mathrm{rank})$ tracks a bootstrapped probability of a given team outranking the team one rank below.}
\scriptsize\begin{supertabular}{llrrr}
AR & MSA & 0.6700 & 0.6488 & 0.6524 \\
AR & UCSC & 0.6594 & 0.6328 & 0.8339 \\
AR & SmurfCat & 0.6274 & 0.5864 & 0.7528 \\
AR & Deloitte & 0.6043 & 0.6046 & 0.5605 \\
AR & CCNU & 0.5995 & 0.6583 & 0.7493 \\
AR & Team Cantharellus & 0.5804 & 0.5886 & 0.6839 \\
AR & DeepPavlov & 0.5628 & 0.5754 & 0.6908 \\
AR & BlueToad & 0.5470 & 0.5058 & 0.5848 \\
AR & NCL-UoR & 0.5390 & 0.5710 & 0.5292 \\
AR & HalluSearch & 0.5362 & 0.5258 & 0.5341 \\
AR & LCTeam & 0.5335 & 0.5537 & 0.6058 \\
AR & UZH & 0.5253 & 0.4871 & 0.6804 \\
AR & AILS-NTUA & 0.5140 & 0.5751 & 0.9473 \\
AR & TUM-MiKaNi & 0.4778 & 0.5114 & 0.5395 \\
AR & nsu-ai & 0.4756 & 0.4236 & 0.6333 \\
AR & tsotsalab & 0.4673 & 0.4765 & 1.0000 \\
AR & REFIND & 0.3743 & 0.1818 & 0.7772 \\
AR & keepitsimple & 0.3631 & 0.2499 & 0.5420 \\
AR & \emph{Baseline (mark all)} & 0.3614 & 0.0067 & 0.7736 \\
AR & UMUTeam & 0.3436 & 0.4211 & 0.5191 \\
AR & TrustAI & 0.3428 & 0.2380 & 0.5724 \\
AR & CUET\_SSTM & 0.3413 & 0.2242 & 0.8613 \\
AR & Swushroomsia & 0.3097 & 0.2874 & 0.8740 \\
AR & uir-cis & 0.2722 & 0.4477 & 0.7168 \\
AR & TU Munich & 0.2527 & 0.3200 & 0.9389 \\
AR & Howard University - AI4PC & 0.2138 & 0.3844 & 0.6589 \\
AR & NLP\_CIMAT & 0.2044 & 0.0775 & 1.0000 \\
AR & HalluciSeekers & 0.1180 & 0.0572 & 0.9504 \\
AR & Hallucination Detectives & 0.0760 & 0.0275 & 0.9604 \\
AR & FENJI & 0.0467 & 0.0067 & 0.0000 \\
AR & \emph{Baseline (mark none)} & 0.0467 & 0.0067 & 0.6335 \\
AR & \emph{Baseline (neural)} & 0.0418 & 0.1190 &  \\
\midrule
CA & UCSC & 0.7211 & 0.7779 & 0.9763 \\
CA & CCNU & 0.6694 & 0.7479 & 0.5158 \\
CA & SmurfCat & 0.6681 & 0.7127 & 0.5246 \\
CA & AILS-NTUA & 0.6664 & 0.6986 & 0.5662 \\
CA & NCL-UoR & 0.6602 & 0.7203 & 0.5531 \\
CA & MSA & 0.6545 & 0.7126 & 0.9598 \\
CA & TUM-MiKaNi & 0.5971 & 0.5551 & 0.6188 \\
CA & UZH & 0.5857 & 0.6420 & 0.9131 \\
CA & Deloitte & 0.5295 & 0.5571 & 0.5684 \\
CA & Team Cantharellus & 0.5231 & 0.5727 & 0.5149 \\
CA & HalluSearch & 0.5215 & 0.5704 & 0.7249 \\
CA & LCTeam & 0.4924 & 0.4917 & 0.6992 \\
CA & nsu-ai & 0.4682 & 0.5346 & 0.5327 \\
CA & uir-cis & 0.4644 & 0.5432 & 0.5359 \\
CA & tsotsalab & 0.4607 & 0.5187 & 0.7431 \\
CA & UMUTeam & 0.4301 & 0.4295 & 0.6018 \\
CA & DeepPavlov & 0.4179 & 0.6742 & 1.0000 \\
CA & keepitsimple & 0.3161 & 0.3377 & 0.9524 \\
CA & Howard University - AI4PC & 0.2731 & 0.3749 & 0.9220 \\
CA & \emph{Baseline (mark all)} & 0.2423 & 0.0600 & 0.9385 \\
CA & FENJI & 0.1796 & 0.0600 & 0.8567 \\
CA & NLP\_CIMAT & 0.1410 & 0.0690 & 0.9614 \\
CA & \emph{Baseline (mark none)} & 0.0800 & 0.0600 & 0.9523 \\
CA & \emph{Baseline (neural)} & 0.0524 & 0.0645 &  \\
\midrule
CS & AILS-NTUA & 0.5429 & 0.5560 & 0.5468 \\
CS & UCSC & 0.5393 & 0.5763 & 0.9177 \\
CS & MSA & 0.5073 & 0.5516 & 0.6934 \\
CS & HalluSearch & 0.4911 & 0.4942 & 0.5633 \\
CS & CCNU & 0.4852 & 0.5541 & 0.7415 \\
CS & SmurfCat & 0.4608 & 0.4676 & 0.7554 \\
CS & Deloitte & 0.4428 & 0.4808 & 0.5248 \\
CS & NCL-UoR & 0.4409 & 0.5285 & 0.8016 \\
CS & LCTeam & 0.4051 & 0.4357 & 0.6666 \\
CS & Team Cantharellus & 0.3936 & 0.4239 & 0.5111 \\
CS & UZH & 0.3931 & 0.4098 & 0.5595 \\
CS & TUM-MiKaNi & 0.3874 & 0.3738 & 0.7537 \\
CS & tsotsalab & 0.3613 & 0.3668 & 0.6218 \\
CS & BlueToad & 0.3514 & 0.3628 & 0.6707 \\
CS & DeepPavlov & 0.3422 & 0.3192 & 0.5628 \\
CS & UMUTeam & 0.3380 & 0.3600 & 0.7693 \\
CS & uir-cis & 0.3060 & 0.2695 & 0.5014 \\
CS & nsu-ai & 0.3051 & 0.2948 & 0.6184 \\
CS & Howard University - AI4PC & 0.2978 & 0.3066 & 0.6098 \\
CS & keepitsimple & 0.2895 & 0.2423 & 0.9132 \\
CS & REFIND & 0.2761 & 0.0924 & 0.9998 \\
CS & \emph{Baseline (mark all)} & 0.2632 & 0.1000 & 0.9056 \\
CS & NLP\_CIMAT & 0.2201 & 0.1450 & 0.9962 \\
CS & \emph{Baseline (mark none)} & 0.1300 & 0.1000 & 0.7318 \\
CS & FENJI & 0.1073 & 0.1000 & 0.6631 \\
CS & \emph{Baseline (neural)} & 0.0957 & 0.0533 &  \\
\midrule
DE & UCSC & 0.6236 & 0.6507 & 0.6539 \\
DE & MSA & 0.6133 & 0.6107 & 0.7561 \\
DE & CCNU & 0.5917 & 0.6089 & 0.6607 \\
DE & AILS-NTUA & 0.5820 & 0.6367 & 0.5643 \\
DE & ATLANTIS & 0.5774 & 0.0133 & 0.6602 \\
DE & Deloitte & 0.5655 & 0.5493 & 0.5232 \\
DE & Team Cantharellus & 0.5639 & 0.5361 & 0.5091 \\
DE & LCTeam & 0.5634 & 0.5031 & 0.5355 \\
DE & SmurfCat & 0.5608 & 0.5721 & 0.5489 \\
DE & TUM-MiKaNi & 0.5569 & 0.5088 & 0.6174 \\
DE & NCL-UoR & 0.5473 & 0.5860 & 0.5351 \\
DE & BlueToad & 0.5439 & 0.5243 & 0.7899 \\
DE & HalluSearch & 0.5187 & 0.5056 & 0.5959 \\
DE & UZH & 0.5123 & 0.5028 & 0.5426 \\
DE & Swushroomsia & 0.5093 & 0.4914 & 0.5644 \\
DE & DeepPavlov & 0.5040 & 0.6126 & 0.8116 \\
DE & nsu-ai & 0.4841 & 0.4584 & 0.9939 \\
DE & UMUTeam & 0.4093 & 0.4403 & 0.6649 \\
DE & tsotsalab & 0.3969 & 0.3614 & 0.6207 \\
DE & REFIND & 0.3862 & 0.3530 & 0.7106 \\
DE & keepitsimple & 0.3651 & 0.2199 & 0.8853 \\
DE & TU Munich & 0.3476 & -0.0059 & 0.9854 \\
DE & \emph{Baseline (mark all)} & 0.3451 & 0.0133 & 0.5550 \\
DE & uir-cis & 0.3400 & 0.4066 & 0.5767 \\
DE & TrustAI & 0.3323 & 0.5121 & 0.9964 \\
DE & Howard University - AI4PC & 0.2522 & 0.2764 & 0.9986 \\
DE & FENJI & 0.1624 & 0.0133 & 1.0000 \\
DE & HalluciSeekers & 0.0573 & 0.0440 & 0.9901 \\
DE & \emph{Baseline (neural)} & 0.0318 & 0.1073 & 1.0000 \\
DE & \emph{Baseline (mark none)} & 0.0267 & 0.0133 & 0.0000 \\
DE & S1mT5v-FMI & 0.0267 & 0.0109 &  \\
\midrule
EN & iai\_MSU & 0.6509 & 0.6294 & 0.9665 \\
EN & UCSC & 0.6146 & 0.5461 & 0.9625 \\
EN & ATLANTIS & 0.5698 & 0.0000 & 0.5621 \\
EN & HalluSearch & 0.5656 & 0.5360 & 0.8407 \\
EN & CCNU & 0.5394 & 0.5509 & 0.6083 \\
EN & MSA & 0.5314 & 0.5200 & 0.5070 \\
EN & AILS-NTUA & 0.5308 & 0.6381 & 0.5924 \\
EN & TUM-MiKaNi & 0.5249 & 0.5363 & 0.5124 \\
EN & SmurfCat & 0.5241 & 0.5963 & 0.5104 \\
EN & Deloitte & 0.5234 & 0.5608 & 0.5569 \\
EN & NCL-UoR & 0.5195 & 0.5477 & 0.7152 \\
EN & Swushroomsia & 0.5030 & 0.4632 & 0.5547 \\
EN & DeepPavlov & 0.4989 & 0.6021 & 0.6875 \\
EN & UZH & 0.4850 & 0.4824 & 0.5656 \\
EN & YNU-HPCC & 0.4807 & 0.4075 & 0.6000 \\
EN & LCTeam & 0.4725 & 0.5538 & 0.5108 \\
EN & Team Cantharellus & 0.4721 & 0.4613 & 0.5451 \\
EN & BlueToad & 0.4688 & 0.4509 & 0.6230 \\
EN & GIL-IIMAS UNAM & 0.4607 & 0.5015 & 0.5468 \\
EN & NLP\_CIMAT & 0.4577 & 0.3707 & 0.6814 \\
EN & tsotsalab & 0.4454 & 0.3946 & 0.5206 \\
EN & advacheck & 0.4443 & 0.3432 & 0.5063 \\
EN & nsu-ai & 0.4436 & 0.4578 & 0.8966 \\
EN & uir-cis & 0.4025 & 0.4781 & 0.7260 \\
EN & VerbaNexAI & 0.3810 & 0.3643 & 0.6902 \\
EN & UMUTeam & 0.3667 & 0.4966 & 0.5090 \\
EN & keepitsimple & 0.3660 & 0.2104 & 0.5712 \\
EN & TU Munich & 0.3646 & 0.2164 & 0.9208 \\
EN & REFIND & 0.3525 & 0.1082 & 0.9991 \\
EN & \emph{Baseline (mark all)} & 0.3489 & 0.0000 & 0.7490 \\
EN & MALTO & 0.3269 & 0.3104 & 0.6742 \\
EN & RaggedyFive & 0.3151 & 0.3038 & 0.5591 \\
EN & COGUMELO & 0.3107 & 0.2277 & 0.5233 \\
EN & HalluRAG-RUG & 0.3093 & 0.0833 & 0.6466 \\
EN & TrustAI & 0.2980 & 0.5642 & 0.5582 \\
EN & FunghiFunghi & 0.2943 & 0.0116 & 0.9975 \\
EN & Hallucination Detectives & 0.2142 & 0.1682 & 0.8576 \\
EN & FENJI & 0.1856 & 0.0000 & 0.9790 \\
EN & Howard University - AI4PC & 0.1325 & 0.2752 & 1.0000 \\
EN & DUTJBD & 0.0571 & -0.1883 & 0.5740 \\
EN & HalluciSeekers & 0.0542 & 0.1530 & 1.0000 \\
EN & HausaNLP & 0.0325 & 0.4226 & 0.0000 \\
EN & \emph{Baseline (mark none)} & 0.0325 & 0.0000 & 0.5153 \\
EN & \emph{Baseline (neural)} & 0.0310 & 0.1190 &  \\
\midrule
ES & ATLANTIS & 0.5311 & 0.0132 & 0.6503 \\
ES & NLP\_CIMAT & 0.5209 & 0.5237 & 0.5948 \\
ES & NCL-UoR & 0.5146 & 0.5464 & 0.5271 \\
ES & CCNU & 0.5125 & 0.5415 & 0.6663 \\
ES & AILS-NTUA & 0.5004 & 0.5648 & 0.7948 \\
ES & UCSC & 0.4794 & 0.6023 & 0.8980 \\
ES & LCTeam & 0.4434 & 0.4335 & 0.6173 \\
ES & SmurfCat & 0.4342 & 0.4406 & 0.7016 \\
ES & MSA & 0.4162 & 0.5450 & 0.6848 \\
ES & Deloitte & 0.4065 & 0.5853 & 0.5258 \\
ES & UZH & 0.4051 & 0.5085 & 0.7683 \\
ES & HalluSearch & 0.3883 & 0.4456 & 0.5202 \\
ES & Team Cantharellus & 0.3869 & 0.4236 & 0.6723 \\
ES & TUM-MiKaNi & 0.3739 & 0.5027 & 0.8242 \\
ES & uir-cis & 0.3447 & 0.3104 & 0.9255 \\
ES & UMUTeam & 0.2980 & 0.4152 & 0.6798 \\
ES & nsu-ai & 0.2854 & 0.3966 & 0.6198 \\
ES & GIL-IIMAS UNAM & 0.2807 & 0.3243 & 0.5467 \\
ES & BlueToad & 0.2787 & 0.4267 & 0.6647 \\
ES & TrustAI & 0.2683 & 0.4983 & 0.6320 \\
ES & DeepPavlov & 0.2614 & 0.3989 & 0.5866 \\
ES & TU Munich & 0.2578 & 0.3229 & 0.6731 \\
ES & Swushroomsia & 0.2466 & 0.2480 & 0.6459 \\
ES & REFIND & 0.2348 & 0.1308 & 0.7627 \\
ES & keepitsimple & 0.2131 & 0.2335 & 1.0000 \\
ES & \emph{Baseline (mark all)} & 0.1853 & 0.0132 & 0.0000 \\
ES & tsotsalab & 0.1853 & 0.0132 & 0.9626 \\
ES & FunghiFunghi & 0.1616 & -0.0986 & 0.9017 \\
ES & Howard University - AI4PC & 0.1341 & 0.3643 & 0.5256 \\
ES & FENJI & 0.1325 & 0.0132 & 0.5085 \\
ES & COGUMELO & 0.1321 & 0.1013 & 0.9591 \\
ES & \emph{Baseline (mark none)} & 0.0855 & 0.0132 & 0.0000 \\
ES & S1mT5v-FMI & 0.0855 & 0.0132 & 0.8743 \\
ES & \emph{Baseline (neural)} & 0.0724 & 0.0359 & 0.8347 \\
ES & HalluciSeekers & 0.0519 & 0.0266 &  \\
\midrule
EU & MSA & 0.6129 & 0.6202 & 0.8451 \\
EU & UCSC & 0.5894 & 0.5826 & 0.6768 \\
EU & CCNU & 0.5784 & 0.6121 & 0.8086 \\
EU & AILS-NTUA & 0.5550 & 0.5805 & 0.7108 \\
EU & Team Cantharellus & 0.5339 & 0.5038 & 0.5998 \\
EU & HalluSearch & 0.5251 & 0.4789 & 0.5244 \\
EU & TUM-MiKaNi & 0.5237 & 0.4709 & 0.5369 \\
EU & Deloitte & 0.5218 & 0.5157 & 0.5307 \\
EU & SmurfCat & 0.5195 & 0.4697 & 0.5919 \\
EU & NCL-UoR & 0.5105 & 0.5974 & 0.5382 \\
EU & UZH & 0.5071 & 0.5108 & 0.5180 \\
EU & BlueToad & 0.5061 & 0.4571 & 0.7607 \\
EU & LCTeam & 0.4804 & 0.5499 & 0.8401 \\
EU & nsu-ai & 0.4368 & 0.4210 & 0.6977 \\
EU & keepitsimple & 0.4193 & 0.3525 & 0.7503 \\
EU & REFIND & 0.4074 & 0.2713 & 0.7908 \\
EU & DeepPavlov & 0.3872 & 0.3214 & 0.7855 \\
EU & \emph{Baseline (mark all)} & 0.3671 & 0.0000 & 0.8667 \\
EU & tsotsalab & 0.3524 & 0.0000 & 0.8191 \\
EU & UMUTeam & 0.3272 & 0.3925 & 0.8306 \\
EU & uir-cis & 0.2916 & 0.3989 & 0.8698 \\
EU & Howard University - AI4PC & 0.2461 & 0.1707 & 0.9953 \\
EU & NLP\_CIMAT & 0.1755 & 0.0522 & 0.9316 \\
EU & FENJI & 0.1326 & 0.0000 & 1.0000 \\
EU & \emph{Baseline (neural)} & 0.0208 & 0.1004 & 1.0000 \\
EU & \emph{Baseline (mark none)} & 0.0101 & 0.0000 &  \\
\midrule
FA & AILS-NTUA & 0.7110 & 0.6989 & 0.7241 \\
FA & UCSC & 0.6949 & 0.6955 & 0.7695 \\
FA & MSA & 0.6693 & 0.6795 & 0.5967 \\
FA & CCNU & 0.6600 & 0.6710 & 0.5171 \\
FA & NCL-UoR & 0.6586 & 0.6732 & 0.5360 \\
FA & Team Cantharellus & 0.6551 & 0.6864 & 0.6600 \\
FA & SmurfCat & 0.6375 & 0.6281 & 0.8067 \\
FA & LCTeam & 0.6018 & 0.4559 & 0.7733 \\
FA & Deloitte & 0.5754 & 0.5191 & 0.5473 \\
FA & BlueToad & 0.5711 & 0.5788 & 0.7372 \\
FA & TUM-MiKaNi & 0.5465 & 0.4238 & 0.8633 \\
FA & UZH & 0.5108 & 0.4990 & 0.8789 \\
FA & UMUTeam & 0.4677 & 0.3939 & 0.6963 \\
FA & HalluSearch & 0.4443 & 0.4734 & 0.9583 \\
FA & nsu-ai & 0.3729 & 0.3875 & 0.9510 \\
FA & keepitsimple & 0.3132 & 0.3570 & 0.9975 \\
FA & DeepPavlov & 0.2405 & 0.1859 & 0.9674 \\
FA & \emph{Baseline (mark all)} & 0.2028 & 0.0100 & 0.0000 \\
FA & tsotsalab & 0.2028 & 0.0100 & 0.8532 \\
FA & uir-cis & 0.1661 & 0.3946 & 0.9212 \\
FA & Howard University - AI4PC & 0.1190 & 0.0661 & 0.6139 \\
FA & HalluciSeekers & 0.1126 & 0.0744 & 1.0000 \\
FA & NLP\_CIMAT & 0.0316 & 0.3949 & 0.9998 \\
FA & FENJI & 0.0028 & 0.0100 & 0.8569 \\
FA & \emph{Baseline (neural)} & 0.0001 & 0.1078 & 0.6366 \\
FA & \emph{Baseline (mark none)} & 0.0000 & 0.0100 &  \\
\midrule
FI & UCSC & 0.6483 & 0.6498 & 0.6351 \\
FI & MSA & 0.6422 & 0.5467 & 0.7680 \\
FI & SmurfCat & 0.6310 & 0.5535 & 0.5095 \\
FI & Deloitte & 0.6307 & 0.6356 & 0.6110 \\
FI & TUM-MiKaNi & 0.6267 & 0.5751 & 0.5588 \\
FI & AILS-NTUA & 0.6235 & 0.6204 & 0.8142 \\
FI & UZH & 0.6014 & 0.4736 & 0.7918 \\
FI & nsu-ai & 0.5874 & 0.4922 & 0.5663 \\
FI & DeepPavlov & 0.5845 & 0.4821 & 0.7057 \\
FI & Team Cantharellus & 0.5714 & 0.5646 & 0.5360 \\
FI & BlueToad & 0.5694 & 0.4906 & 0.5195 \\
FI & HalluSearch & 0.5681 & 0.5297 & 0.9810 \\
FI & CCNU & 0.5117 & 0.5631 & 0.5345 \\
FI & NCL-UoR & 0.5096 & 0.4965 & 0.5489 \\
FI & REFIND & 0.5061 & 0.1965 & 0.6705 \\
FI & Swushroomsia & 0.4955 & 0.4298 & 0.6538 \\
FI & \emph{Baseline (mark all)} & 0.4857 & 0.0000 & 0.0000 \\
FI & tsotsalab & 0.4857 & 0.0000 & 0.4983 \\
FI & TU Munich & 0.4857 & 0.0032 & 0.9342 \\
FI & UMUTeam & 0.4563 & 0.5126 & 0.5228 \\
FI & keepitsimple & 0.4554 & 0.3323 & 0.9026 \\
FI & LCTeam & 0.4221 & 0.5300 & 0.7620 \\
FI & Howard University - AI4PC & 0.3996 & 0.3433 & 0.9081 \\
FI & NLP\_CIMAT & 0.3742 & 0.0310 & 1.0000 \\
FI & TrustAI & 0.2955 & 0.1777 & 0.9709 \\
FI & uir-cis & 0.2459 & 0.3366 & 1.0000 \\
FI & FENJI & 0.0941 & 0.0000 & 1.0000 \\
FI & \emph{Baseline (neural)} & 0.0042 & 0.0924 & 1.0000 \\
FI & S1mT5v-FMI & 0.0000 & 0.0014 & 0.0000 \\
FI & \emph{Baseline (mark none)} & 0.0000 & 0.0000 &  \\
\midrule
FR & Deloitte & 0.6469 & 0.6187 & 0.8473 \\
FR & TUM-MiKaNi & 0.6314 & 0.5157 & 0.7031 \\
FR & MSA & 0.6195 & 0.5553 & 0.8684 \\
FR & Swushroomsia & 0.5937 & 0.5429 & 0.6097 \\
FR & UCSC & 0.5868 & 0.5592 & 0.5470 \\
FR & SmurfCat & 0.5838 & 0.5155 & 0.5186 \\
FR & DeepPavlov & 0.5831 & 0.5440 & 0.5269 \\
FR & AILS-NTUA & 0.5812 & 0.6103 & 0.5598 \\
FR & UZH & 0.5765 & 0.4411 & 0.6858 \\
FR & LCTeam & 0.5634 & 0.4883 & 0.9769 \\
FR & ATLANTIS & 0.5190 & 0.4117 & 0.5157 \\
FR & nsu-ai & 0.5181 & 0.4339 & 0.5555 \\
FR & Team Cantharellus & 0.5147 & 0.5317 & 0.8106 \\
FR & tsotsalab & 0.4896 & 0.4575 & 0.5975 \\
FR & CCNU & 0.4823 & 0.5724 & 0.5911 \\
FR & REFIND & 0.4734 & 0.0752 & 0.8088 \\
FR & keepitsimple & 0.4651 & 0.2756 & 0.8789 \\
FR & TU Munich & 0.4547 & 0.0096 & 1.0000 \\
FR & \emph{Baseline (mark all)} & 0.4543 & 0.0000 & 0.6780 \\
FR & BlueToad & 0.4385 & 0.3797 & 0.5235 \\
FR & HalluSearch & 0.4366 & 0.3365 & 0.8049 \\
FR & Howard University - AI4PC & 0.4164 & 0.3990 & 0.6451 \\
FR & NCL-UoR & 0.4058 & 0.4187 & 0.7890 \\
FR & TrustAI & 0.3799 & 0.4992 & 0.9097 \\
FR & NLP\_CIMAT & 0.3533 & 0.0711 & 0.9046 \\
FR & UMUTeam & 0.3200 & 0.4117 & 0.6506 \\
FR & FunghiFunghi & 0.3095 & -0.1521 & 0.9882 \\
FR & uir-cis & 0.2286 & 0.2873 & 1.0000 \\
FR & FENJI & 0.0844 & 0.0000 & 0.9765 \\
FR & HalluciSeekers & 0.0500 & 0.0447 & 1.0000 \\
FR & \emph{Baseline (neural)} & 0.0022 & 0.0208 & 1.0000 \\
FR & \emph{Baseline (mark none)} & 0.0000 & 0.0000 & 0.0000 \\
FR & S1mT5v-FMI & 0.0000 & 0.0000 &  \\
\midrule
HI & CCNU & 0.7466 & 0.7847 & 0.5416 \\
HI & UCSC & 0.7441 & 0.7625 & 0.7904 \\
HI & AILS-NTUA & 0.7259 & 0.7602 & 0.6522 \\
HI & SmurfCat & 0.7164 & 0.5964 & 0.8993 \\
HI & MSA & 0.6842 & 0.7252 & 0.7717 \\
HI & LCTeam & 0.6601 & 0.5122 & 0.5380 \\
HI & Team Cantharellus & 0.6572 & 0.6909 & 0.6528 \\
HI & BlueToad & 0.6447 & 0.6844 & 0.5870 \\
HI & UZH & 0.6377 & 0.6687 & 0.5820 \\
HI & Deloitte & 0.6322 & 0.6391 & 0.5441 \\
HI & NCL-UoR & 0.6286 & 0.6830 & 0.9337 \\
HI & TUM-MiKaNi & 0.5835 & 0.4964 & 0.9574 \\
HI & HalluSearch & 0.5265 & 0.5195 & 0.6682 \\
HI & DeepPavlov & 0.5117 & 0.7320 & 0.9032 \\
HI & nsu-ai & 0.4771 & 0.4438 & 0.7440 \\
HI & Swushroomsia & 0.4534 & 0.4789 & 0.5208 \\
HI & UMUTeam & 0.4510 & 0.4386 & 0.9989 \\
HI & keepitsimple & 0.3598 & 0.3508 & 0.9376 \\
HI & TrustAI & 0.3144 & 0.5050 & 0.9049 \\
HI & TU Munich & 0.2807 & 0.3297 & 0.7051 \\
HI & \emph{Baseline (mark all)} & 0.2711 & 0.0000 & 0.0000 \\
HI & tsotsalab & 0.2711 & 0.0000 & 0.7323 \\
HI & Howard University - AI4PC & 0.2586 & 0.3217 & 1.0000 \\
HI & uir-cis & 0.0613 & 0.5586 & 1.0000 \\
HI & \emph{Baseline (neural)} & 0.0029 & 0.1429 & 0.9999 \\
HI & FENJI & 0.0000 & 0.0000 & 0.0000 \\
HI & \emph{Baseline (mark none)} & 0.0000 & 0.0000 &  \\
\midrule
IT & UCSC & 0.7872 & 0.7873 & 0.8312 \\
IT & AILS-NTUA & 0.7660 & 0.8195 & 0.8213 \\
IT & SmurfCat & 0.7478 & 0.6231 & 0.6926 \\
IT & MSA & 0.7369 & 0.7568 & 0.6386 \\
IT & Swushroomsia & 0.7274 & 0.7292 & 0.7451 \\
IT & NCL-UoR & 0.7123 & 0.7614 & 0.6025 \\
IT & CCNU & 0.7060 & 0.7441 & 0.5030 \\
IT & Deloitte & 0.7059 & 0.6144 & 0.5933 \\
IT & LCTeam & 0.7013 & 0.5487 & 0.6953 \\
IT & Team Cantharellus & 0.6907 & 0.7118 & 0.5958 \\
IT & UZH & 0.6833 & 0.7016 & 0.5643 \\
IT & TUM-MiKaNi & 0.6787 & 0.5388 & 0.9468 \\
IT & BlueToad & 0.6388 & 0.6675 & 0.9977 \\
IT & HalluSearch & 0.5484 & 0.5604 & 0.7456 \\
IT & DeepPavlov & 0.5280 & 0.5529 & 0.9992 \\
IT & UMUTeam & 0.4413 & 0.4601 & 0.5250 \\
IT & nsu-ai & 0.4396 & 0.4402 & 0.9502 \\
IT & keepitsimple & 0.4009 & 0.3860 & 0.5463 \\
IT & uir-cis & 0.3967 & 0.4991 & 0.9130 \\
IT & TrustAI & 0.3441 & 0.2827 & 0.6926 \\
IT & TU Munich & 0.3319 & 0.4210 & 0.5730 \\
IT & REFIND & 0.3255 & 0.2423 & 0.8826 \\
IT & \emph{Baseline (mark all)} & 0.2826 & 0.0000 & 0.0000 \\
IT & tsotsalab & 0.2826 & 0.0000 & 0.5678 \\
IT & FENJI & 0.2765 & 0.0000 & 0.6012 \\
IT & Howard University - AI4PC & 0.2675 & 0.4021 & 0.9983 \\
IT & FunghiFunghi & 0.2111 & -0.2116 & 0.9084 \\
IT & NLP\_CIMAT & 0.1899 & 0.0456 & 1.0000 \\
IT & HalluciSeekers & 0.0350 & 0.0242 & 0.9991 \\
IT & \emph{Baseline (neural)} & 0.0104 & 0.0800 & 1.0000 \\
IT & \emph{Baseline (mark none)} & 0.0000 & 0.0000 &  \\
\midrule
SV & UCSC & 0.6423 & 0.5204 & 0.6115 \\
SV & MSA & 0.6364 & 0.4224 & 0.7683 \\
SV & Deloitte & 0.6220 & 0.5374 & 0.5804 \\
SV & SmurfCat & 0.6174 & 0.5007 & 0.7523 \\
SV & AILS-NTUA & 0.6009 & 0.5622 & 0.6801 \\
SV & TUM-MiKaNi & 0.5886 & 0.3930 & 0.5600 \\
SV & BlueToad & 0.5854 & 0.4267 & 0.8365 \\
SV & HalluSearch & 0.5622 & 0.4290 & 0.5161 \\
SV & UZH & 0.5612 & 0.4125 & 0.6110 \\
SV & NCL-UoR & 0.5547 & 0.4587 & 0.6021 \\
SV & nsu-ai & 0.5478 & 0.3442 & 0.6642 \\
SV & DeepPavlov & 0.5380 & 0.4147 & 0.5194 \\
SV & \emph{Baseline (mark all)} & 0.5373 & 0.0136 & 0.6366 \\
SV & TU Munich & 0.5372 & 0.0054 & 0.8667 \\
SV & tsotsalab & 0.5349 & 0.0136 & 0.7915 \\
SV & CCNU & 0.5045 & 0.5058 & 0.9847 \\
SV & UMUTeam & 0.4393 & 0.3936 & 0.7617 \\
SV & LCTeam & 0.4183 & 0.3700 & 0.5270 \\
SV & FunghiFunghi & 0.4156 & -0.1177 & 0.7785 \\
SV & keepitsimple & 0.3967 & 0.2170 & 0.9123 \\
SV & Swushroomsia & 0.3549 & 0.2265 & 0.9004 \\
SV & uir-cis & 0.3080 & 0.3655 & 0.9391 \\
SV & TrustAI & 0.2484 & 0.2551 & 0.6641 \\
SV & NLP\_CIMAT & 0.2388 & 0.0547 & 1.0000 \\
SV & FENJI & 0.1154 & 0.0136 & 0.5666 \\
SV & Howard University - AI4PC & 0.1110 & 0.0669 & 0.9929 \\
SV & HalluciSeekers & 0.0575 & 0.0856 & 0.9999 \\
SV & \emph{Baseline (neural)} & 0.0308 & 0.0968 & 1.0000 \\
SV & \emph{Baseline (mark none)} & 0.0204 & 0.0136 & 0.0000 \\
SV & S1mT5v-FMI & 0.0204 & 0.0136 &  \\
\midrule
ZH & YNU-HPCC & 0.5540 & 0.3518 & 0.8353 \\
ZH & LCTeam & 0.5232 & 0.5171 & 0.9948 \\
ZH & nsu-ai & 0.4937 & 0.3813 & 0.6401 \\
ZH & DeepPavlov & 0.4900 & 0.2529 & 0.9998 \\
ZH & SmurfCat & 0.4842 & 0.2529 & 0.7478 \\
ZH & UZH & 0.4790 & 0.1783 & 0.6436 \\
ZH & \emph{Baseline (mark all)} & 0.4772 & 0.0000 & 0.0000 \\
ZH & tsotsalab & 0.4772 & 0.0000 & 0.5986 \\
ZH & TUM-MiKaNi & 0.4735 & 0.4095 & 0.5653 \\
ZH & UCSC & 0.4707 & 0.3966 & 0.5092 \\
ZH & keepitsimple & 0.4703 & 0.1601 & 0.6149 \\
ZH & MSA & 0.4631 & 0.4363 & 0.5659 \\
ZH & Deloitte & 0.4600 & 0.2986 & 0.6281 \\
ZH & HalluSearch & 0.4534 & 0.4232 & 0.8504 \\
ZH & TrustAI & 0.4304 & 0.2503 & 0.8820 \\
ZH & Team Cantharellus & 0.4011 & 0.4063 & 0.7328 \\
ZH & UMUTeam & 0.3875 & 0.4916 & 0.5145 \\
ZH & AILS-NTUA & 0.3866 & 0.4564 & 0.5588 \\
ZH & CCNU & 0.3834 & 0.4042 & 0.8326 \\
ZH & NCL-UoR & 0.3606 & 0.3540 & 0.9996 \\
ZH & BlueToad & 0.2783 & 0.2262 & 0.9996 \\
ZH & TU Munich & 0.2160 & 0.0769 & 0.5104 \\
ZH & Howard University - AI4PC & 0.2152 & 0.1119 & 0.6256 \\
ZH & Swushroomsia & 0.2054 & 0.0966 & 0.7185 \\
ZH & uir-cis & 0.1913 & 0.3047 & 1.0000 \\
ZH & S1mT5v-FMI & 0.0619 & -0.0209 & 0.9913 \\
ZH & FENJI & 0.0371 & 0.0000 & 0.9991 \\
ZH & \emph{Baseline (neural)} & 0.0236 & 0.0884 & 1.0000 \\
ZH & \emph{Baseline (mark none)} & 0.0200 & 0.0000 &  \\
\end{supertabular}
\end{center}

%% file: tables/cor-results/results-all.tex
\begin{center}
\tablefirsthead{\toprule 
\textbf{Lang} & \textbf{Team} & \textbf{IoU} & $\rho$ & $\Pr(\mathrm{rank})$ \\ \midrule}
\tablehead{%
\multicolumn{4}{c}%
{{ \smaller (\textit{Continued from previous column})}} \\
\toprule
\textbf{Lang} & \textbf{Team} & \textbf{IoU} & $\rho$ & $\Pr(\mathrm{rank})$ \\ \midrule }
\tabletail{%
\midrule \multicolumn{4}{c}{{\smaller (\textit{Continued on next column})}} \\}
\tablelasttail{%
\bottomrule
}
\bottomcaption{\label{tab:alt rankings} Alternative rankings based on highest $\mathrm{cor}$ score across all team submission. Column $\Pr(\mathrm{rank})$ tracks a bootstrapped probability of a given team outranking the team one rank below. }
\scriptsize\begin{supertabular}{llrrr}
AR & CCNU & 0.6583 & 0.5995 & 0.5659 \\
AR & UCSC & 0.6543 & 0.6059 & 0.5739 \\
AR & MSA & 0.6488 & 0.6700 & 0.6721 \\
AR & Deloitte & 0.6371 & 0.5870 & 0.9823 \\
AR & Team Cantharellus & 0.5886 & 0.5804 & 0.5211 \\
AR & SmurfCat & 0.5869 & 0.5545 & 0.5079 \\
AR & AILS-NTUA & 0.5865 & 0.4967 & 0.6484 \\
AR & DeepPavlov & 0.5754 & 0.5628 & 0.5620 \\
AR & NCL-UoR & 0.5710 & 0.5390 & 0.7234 \\
AR & LCTeam & 0.5537 & 0.5335 & 0.8243 \\
AR & TrustAI & 0.5385 & 0.2843 & 0.6550 \\
AR & HalluSearch & 0.5258 & 0.5362 & 0.6807 \\
AR & TUM-MiKaNi & 0.5114 & 0.4778 & 0.5722 \\
AR & BlueToad & 0.5058 & 0.5470 & 0.5395 \\
AR & UZH & 0.5023 & 0.5029 & 0.7901 \\
AR & tsotsalab & 0.4765 & 0.4673 & 0.8317 \\
AR & uir-cis & 0.4477 & 0.2722 & 0.5060 \\
AR & CUET\_SSTM & 0.4472 & 0.0978 & 0.9110 \\
AR & nsu-ai & 0.4236 & 0.4756 & 0.5476 \\
AR & UMUTEAM & 0.4211 & 0.3436 & 0.8914 \\
AR & TU Munich & 0.3973 & 0.1480 & 0.7806 \\
AR & Howard University - AI4PC & 0.3844 & 0.2138 & 0.9984 \\
AR & Swushroomsia & 0.2874 & 0.3097 & 0.8264 \\
AR & keepitsimple & 0.2499 & 0.3631 & 0.9920 \\
AR & REFIND & 0.1818 & 0.3737 & 0.9943 \\
AR & \emph{Baseline (neural)} & 0.1190 & 0.0418 & 0.7890 \\
AR & NLP\_CIMAT & 0.0969 & 0.1447 & 0.9276 \\
AR & HalluciSeekers & 0.0572 & 0.1180 & 0.8036 \\
AR & Hallucination Detectives & 0.0358 & 0.0755 & 0.9706 \\
AR & \emph{Baseline (mark all)} & 0.0067 & 0.3614 & 0.0000 \\
AR & FENJI & 0.0067 & 0.0467 & 0.0000 \\
AR & \emph{Baseline (mark none)} & 0.0067 & 0.0467 &  \\
\midrule
CA & UCSC & 0.7844 & 0.6711 & 0.9340 \\
CA & CCNU & 0.7479 & 0.6694 & 0.8359 \\
CA & NCL-UoR & 0.7203 & 0.6602 & 0.5959 \\
CA & SmurfCat & 0.7127 & 0.6681 & 0.4948 \\
CA & MSA & 0.7126 & 0.6545 & 0.6662 \\
CA & AILS-NTUA & 0.6986 & 0.6664 & 0.7767 \\
CA & DeepPavlov & 0.6742 & 0.4179 & 0.8452 \\
CA & UZH & 0.6420 & 0.5857 & 0.6978 \\
CA & Deloitte & 0.6219 & 0.5032 & 0.9472 \\
CA & Team Cantharellus & 0.5727 & 0.5231 & 0.5206 \\
CA & HalluSearch & 0.5704 & 0.5215 & 0.6358 \\
CA & TUM-MiKaNi & 0.5551 & 0.5971 & 0.6483 \\
CA & uir-cis & 0.5432 & 0.4644 & 0.5808 \\
CA & nsu-ai & 0.5346 & 0.4682 & 0.6384 \\
CA & tsotsalab & 0.5187 & 0.4607 & 0.7913 \\
CA & LCTeam & 0.4937 & 0.4441 & 1.0000 \\
CA & UMUTEAM & 0.4295 & 0.4301 & 0.9859 \\
CA & Howard University - AI4PC & 0.3749 & 0.2731 & 0.8240 \\
CA & keepitsimple & 0.3377 & 0.3161 & 1.0000 \\
CA & NLP\_CIMAT & 0.0690 & 0.1410 & 0.5481 \\
CA & \emph{Baseline (neural)} & 0.0645 & 0.0524 & 0.5686 \\
CA & \emph{Baseline (mark all)} & 0.0600 & 0.2423 & 0.0000 \\
CA & FENJI & 0.0600 & 0.1796 & 0.0000 \\
CA & \emph{Baseline (mark none)} & 0.0600 & 0.0800 &  \\
\midrule
CS & UCSC & 0.5993 & 0.5072 & 0.9486 \\
CS & AILS-NTUA & 0.5560 & 0.5429 & 0.5223 \\
CS & CCNU & 0.5541 & 0.4852 & 0.5290 \\
CS & MSA & 0.5516 & 0.5073 & 0.6836 \\
CS & SmurfCat & 0.5334 & 0.4510 & 0.5601 \\
CS & NCL-UoR & 0.5285 & 0.4409 & 0.7306 \\
CS & Deloitte & 0.5034 & 0.3740 & 0.5971 \\
CS & HalluSearch & 0.4942 & 0.4911 & 0.8126 \\
CS & TUM-MiKaNi & 0.4580 & 0.3853 & 0.8116 \\
CS & Team Cantharellus & 0.4373 & 0.3823 & 0.5393 \\
CS & LCTeam & 0.4357 & 0.4051 & 0.7623 \\
CS & UZH & 0.4098 & 0.3931 & 0.8792 \\
CS & tsotsalab & 0.3668 & 0.3613 & 0.5444 \\
CS & BlueToad & 0.3628 & 0.3514 & 0.5323 \\
CS & UMUTEAM & 0.3600 & 0.3380 & 0.9570 \\
CS & DeepPavlov & 0.3215 & 0.3405 & 0.7995 \\
CS & Howard University - AI4PC & 0.3066 & 0.2978 & 0.7143 \\
CS & nsu-ai & 0.2948 & 0.3051 & 0.8137 \\
CS & uir-cis & 0.2695 & 0.3060 & 0.7710 \\
CS & keepitsimple & 0.2423 & 0.2895 & 0.8684 \\
CS & REFIND & 0.1861 & 0.2353 & 0.7297 \\
CS & NLP\_CIMAT & 0.1563 & 0.1821 & 0.9164 \\
CS & \emph{Baseline (mark all)} & 0.1000 & 0.2632 & 0.0000 \\
CS & \emph{Baseline (mark none)} & 0.1000 & 0.1300 & 0.0000 \\
CS & FENJI & 0.1000 & 0.1073 & 0.9208 \\
CS & \emph{Baseline (neural)} & 0.0533 & 0.0957 &  \\
\midrule
DE & UCSC & 0.6588 & 0.6221 & 0.8679 \\
DE & AILS-NTUA & 0.6367 & 0.5820 & 0.8566 \\
DE & Swushroomsia & 0.6160 & 0.2911 & 0.5549 \\
DE & DeepPavlov & 0.6126 & 0.5040 & 0.5318 \\
DE & MSA & 0.6107 & 0.6133 & 0.5303 \\
DE & CCNU & 0.6089 & 0.5917 & 0.5777 \\
DE & SmurfCat & 0.6042 & 0.5050 & 0.7648 \\
DE & NCL-UoR & 0.5860 & 0.5473 & 0.8942 \\
DE & Deloitte & 0.5493 & 0.5655 & 0.7009 \\
DE & Team Cantharellus & 0.5361 & 0.5639 & 0.6559 \\
DE & BlueToad & 0.5243 & 0.5439 & 0.6927 \\
DE & TrustAI & 0.5121 & 0.3323 & 0.5664 \\
DE & TUM-MiKaNi & 0.5088 & 0.5569 & 0.5450 \\
DE & HalluSearch & 0.5056 & 0.5187 & 0.5405 \\
DE & LCTeam & 0.5031 & 0.5634 & 0.5028 \\
DE & UZH & 0.5028 & 0.5123 & 0.9320 \\
DE & ATLANTIS & 0.4607 & 0.5204 & 0.5533 \\
DE & nsu-ai & 0.4584 & 0.4841 & 0.8390 \\
DE & UMUTEAM & 0.4403 & 0.4093 & 0.8853 \\
DE & uir-cis & 0.4066 & 0.3400 & 0.9112 \\
DE & tsotsalab & 0.3614 & 0.3969 & 0.5914 \\
DE & REFIND & 0.3530 & 0.3862 & 0.8035 \\
DE & TU Munich & 0.3195 & 0.2704 & 0.9557 \\
DE & Howard University - AI4PC & 0.2764 & 0.2522 & 0.9473 \\
DE & keepitsimple & 0.2199 & 0.3651 & 0.9997 \\
DE & \emph{Baseline (neural)} & 0.1073 & 0.0318 & 0.9999 \\
DE & HalluciSeekers & 0.0440 & 0.0573 & 0.9406 \\
DE & \emph{Baseline (mark all)} & 0.0133 & 0.3451 & 0.0000 \\
DE & FENJI & 0.0133 & 0.1624 & 0.0000 \\
DE & \emph{Baseline (mark none)} & 0.0133 & 0.0267 & 0.8657 \\
DE & S1mT5v-FMI & 0.0109 & 0.0267 &  \\
\midrule
EN & Swushroomsia & 0.6486 & 0.4769 & 0.5207 \\
EN & UCSC & 0.6479 & 0.5686 & 0.6915 \\
EN & AILS-NTUA & 0.6381 & 0.5308 & 0.6903 \\
EN & iai\_MSU & 0.6294 & 0.6509 & 0.8010 \\
EN & DeepPavlov & 0.6116 & 0.4391 & 0.5101 \\
EN & SmurfCat & 0.6116 & 0.5050 & 0.9324 \\
EN & Deloitte & 0.5833 & 0.5114 & 0.7063 \\
EN & CCNU & 0.5713 & 0.5177 & 0.6222 \\
EN & TrustAI & 0.5642 & 0.2980 & 0.5822 \\
EN & LCTeam & 0.5604 & 0.4590 & 0.8283 \\
EN & TUM-MiKaNi & 0.5506 & 0.3385 & 0.5496 \\
EN & NCL-UoR & 0.5477 & 0.5195 & 0.5497 \\
EN & HalluSearch & 0.5444 & 0.5315 & 0.5956 \\
EN & MSA & 0.5380 & 0.5066 & 0.6428 \\
EN & ATLANTIS & 0.5287 & 0.5159 & 0.6456 \\
EN & UZH & 0.5193 & 0.4699 & 0.7892 \\
EN & GIL-IIMAS UNAM & 0.5015 & 0.4607 & 0.5927 \\
EN & UMUTEAM & 0.4966 & 0.3667 & 0.7933 \\
EN & uir-cis & 0.4781 & 0.4025 & 0.6825 \\
EN & Team Cantharellus & 0.4668 & 0.4289 & 0.6325 \\
EN & nsu-ai & 0.4578 & 0.4436 & 0.5961 \\
EN & BlueToad & 0.4509 & 0.4688 & 0.7907 \\
EN & NLP\_CIMAT & 0.4255 & 0.4270 & 0.5462 \\
EN & HausaNLP & 0.4226 & 0.0325 & 0.6726 \\
EN & tsotsalab & 0.4109 & 0.3793 & 0.5395 \\
EN & YNU-HPCC & 0.4075 & 0.4807 & 0.8106 \\
EN & TU Munich & 0.3760 & 0.2089 & 0.6524 \\
EN & VerbaNexAI & 0.3657 & 0.3634 & 0.7146 \\
EN & advacheck & 0.3498 & 0.4440 & 0.9196 \\
EN & MALTO & 0.3117 & 0.2993 & 0.6146 \\
EN & RaggedyFive & 0.3038 & 0.3151 & 0.8209 \\
EN & Howard University - AI4PC & 0.2752 & 0.1325 & 0.9526 \\
EN & COGUMELO & 0.2277 & 0.3107 & 0.7029 \\
EN & keepitsimple & 0.2104 & 0.3660 & 0.5525 \\
EN & REFIND & 0.2058 & 0.2812 & 0.8422 \\
EN & Hallucination Detectives & 0.1682 & 0.2142 & 0.6660 \\
EN & HalluciSeekers & 0.1530 & 0.0542 & 0.9815 \\
EN & \emph{Baseline (neural)} & 0.1190 & 0.0310 & 0.9739 \\
EN & HalluRAG-RUG & 0.0833 & 0.3093 & 0.9999 \\
EN & FunghiFunghi & 0.0116 & 0.2943 & 0.7477 \\
EN & \emph{Baseline (mark all)} & 0.0000 & 0.3489 & 0.0000 \\
EN & FENJI & 0.0000 & 0.1856 & 0.0000 \\
EN & \emph{Baseline (mark none)} & 0.0000 & 0.0325 & 1.0000 \\
EN & DUTJBD & -0.1883 & 0.0571 &  \\
\midrule
ES & UCSC & 0.6193 & 0.4339 & 0.7162 \\
ES & AILS-NTUA & 0.6068 & 0.4396 & 0.8777 \\
ES & Deloitte & 0.5853 & 0.4065 & 0.8558 \\
ES & SmurfCat & 0.5662 & 0.4308 & 0.6621 \\
ES & CCNU & 0.5575 & 0.5111 & 0.6910 \\
ES & MSA & 0.5477 & 0.4022 & 0.5257 \\
ES & NCL-UoR & 0.5464 & 0.5146 & 0.5140 \\
ES & NLP\_CIMAT & 0.5458 & 0.4727 & 0.9241 \\
ES & UZH & 0.5085 & 0.4051 & 0.5888 \\
ES & TUM-MiKaNi & 0.5027 & 0.3739 & 0.5979 \\
ES & TrustAI & 0.4983 & 0.2683 & 0.9633 \\
ES & Team Cantharellus & 0.4489 & 0.3667 & 0.5371 \\
ES & LCTeam & 0.4471 & 0.4188 & 0.5186 \\
ES & HalluSearch & 0.4456 & 0.3883 & 0.7135 \\
ES & BlueToad & 0.4267 & 0.2787 & 0.5961 \\
ES & DeepPavlov & 0.4207 & 0.2098 & 0.6028 \\
ES & UMUTEAM & 0.4152 & 0.2980 & 0.8696 \\
ES & nsu-ai & 0.3966 & 0.2854 & 0.7848 \\
ES & ATLANTIS & 0.3793 & 0.3606 & 0.7197 \\
ES & Howard University - AI4PC & 0.3643 & 0.1341 & 0.9340 \\
ES & GIL-IIMAS UNAM & 0.3243 & 0.2807 & 0.5278 \\
ES & TU Munich & 0.3229 & 0.2578 & 0.6952 \\
ES & uir-cis & 0.3104 & 0.3447 & 0.9604 \\
ES & Swushroomsia & 0.2480 & 0.2466 & 0.6419 \\
ES & keepitsimple & 0.2335 & 0.2131 & 0.9943 \\
ES & REFIND & 0.1699 & 0.2152 & 0.9940 \\
ES & COGUMELO & 0.1013 & 0.1321 & 0.9965 \\
ES & \emph{Baseline (neural)} & 0.0359 & 0.0724 & 0.7277 \\
ES & HalluciSeekers & 0.0266 & 0.0519 & 0.7879 \\
ES & \emph{Baseline (mark all)} & 0.0132 & 0.1853 & 0.0000 \\
ES & tsotsalab & 0.0132 & 0.1853 & 0.0000 \\
ES & FENJI & 0.0132 & 0.1325 & 0.0000 \\
ES & \emph{Baseline (mark none)} & 0.0132 & 0.0855 & 0.0000 \\
ES & S1mT5v-FMI & 0.0132 & 0.0855 & 1.0000 \\
ES & FunghiFunghi & -0.0986 & 0.1616 &  \\
\midrule
EU & UCSC & 0.6265 & 0.5830 & 0.5927 \\
EU & MSA & 0.6202 & 0.6129 & 0.6186 \\
EU & CCNU & 0.6121 & 0.5784 & 0.6618 \\
EU & NCL-UoR & 0.5974 & 0.5105 & 0.6974 \\
EU & AILS-NTUA & 0.5805 & 0.5550 & 0.7788 \\
EU & LCTeam & 0.5560 & 0.4589 & 0.8008 \\
EU & SmurfCat & 0.5234 & 0.5106 & 0.5951 \\
EU & Deloitte & 0.5157 & 0.5218 & 0.5572 \\
EU & UZH & 0.5108 & 0.5071 & 0.5550 \\
EU & Team Cantharellus & 0.5038 & 0.5339 & 0.5503 \\
EU & TUM-MiKaNi & 0.4996 & 0.4289 & 0.6969 \\
EU & HalluSearch & 0.4789 & 0.5251 & 0.6792 \\
EU & BlueToad & 0.4571 & 0.5061 & 0.7887 \\
EU & nsu-ai & 0.4210 & 0.4368 & 0.6682 \\
EU & uir-cis & 0.3989 & 0.2916 & 0.5576 \\
EU & UMUTEAM & 0.3925 & 0.3272 & 0.7759 \\
EU & REFIND & 0.3552 & 0.3869 & 0.5244 \\
EU & keepitsimple & 0.3525 & 0.4193 & 0.7812 \\
EU & DeepPavlov & 0.3214 & 0.3872 & 1.0000 \\
EU & Howard University - AI4PC & 0.1707 & 0.2461 & 0.9669 \\
EU & \emph{Baseline (neural)} & 0.1004 & 0.0208 & 0.8183 \\
EU & NLP\_CIMAT & 0.0712 & 0.1372 & 0.9993 \\
EU & \emph{Baseline (mark all)} & 0.0000 & 0.3671 & 0.0000 \\
EU & tsotsalab & 0.0000 & 0.3524 & 0.0000 \\
EU & FENJI & 0.0000 & 0.1326 & 0.0000 \\
EU & \emph{Baseline (mark none)} & 0.0000 & 0.0101 &  \\
\midrule
FA & MSA & 0.7009 & 0.6392 & 0.5296 \\
FA & AILS-NTUA & 0.6989 & 0.7110 & 0.5455 \\
FA & UCSC & 0.6955 & 0.6949 & 0.5848 \\
FA & CCNU & 0.6886 & 0.6569 & 0.5365 \\
FA & Team Cantharellus & 0.6864 & 0.6551 & 0.6594 \\
FA & NCL-UoR & 0.6732 & 0.6586 & 0.6557 \\
FA & SmurfCat & 0.6584 & 0.6062 & 0.9823 \\
FA & BlueToad & 0.5788 & 0.5711 & 0.8743 \\
FA & Deloitte & 0.5379 & 0.5139 & 0.8779 \\
FA & UZH & 0.4990 & 0.5108 & 0.7325 \\
FA & TUM-MiKaNi & 0.4762 & 0.5315 & 0.5275 \\
FA & HalluSearch & 0.4734 & 0.4443 & 0.6904 \\
FA & LCTeam & 0.4559 & 0.6018 & 0.8420 \\
FA & NLP\_CIMAT & 0.4297 & 0.0248 & 0.7733 \\
FA & uir-cis & 0.3946 & 0.1661 & 0.5078 \\
FA & UMUTEAM & 0.3939 & 0.4677 & 0.5645 \\
FA & nsu-ai & 0.3875 & 0.3729 & 0.7316 \\
FA & keepitsimple & 0.3570 & 0.3132 & 0.9999 \\
FA & DeepPavlov & 0.1859 & 0.2405 & 0.9600 \\
FA & \emph{Baseline (neural)} & 0.1078 & 0.0001 & 0.8757 \\
FA & HalluciSeekers & 0.0744 & 0.1126 & 0.5677 \\
FA & Howard University - AI4PC & 0.0661 & 0.1190 & 0.9199 \\
FA & \emph{Baseline (mark all)} & 0.0100 & 0.2028 & 0.0000 \\
FA & tsotsalab & 0.0100 & 0.2028 & 0.0000 \\
FA & FENJI & 0.0100 & 0.0028 & 0.0000 \\
FA & \emph{Baseline (mark none)} & 0.0100 & 0.0000 &  \\
\midrule
FI & UCSC & 0.6498 & 0.6483 & 0.6407 \\
FI & Deloitte & 0.6424 & 0.6284 & 0.8912 \\
FI & AILS-NTUA & 0.6204 & 0.6235 & 0.9876 \\
FI & TUM-MiKaNi & 0.5751 & 0.6267 & 0.6593 \\
FI & SmurfCat & 0.5650 & 0.5536 & 0.5089 \\
FI & Team Cantharellus & 0.5646 & 0.5714 & 0.5218 \\
FI & CCNU & 0.5631 & 0.5117 & 0.5371 \\
FI & LCTeam & 0.5611 & 0.3933 & 0.6611 \\
FI & NCL-UoR & 0.5524 & 0.4983 & 0.5927 \\
FI & MSA & 0.5467 & 0.6422 & 0.7053 \\
FI & HalluSearch & 0.5297 & 0.5681 & 0.5222 \\
FI & TrustAI & 0.5281 & 0.1072 & 0.8982 \\
FI & UMUTEAM & 0.5126 & 0.4563 & 0.7632 \\
FI & UZH & 0.4934 & 0.5383 & 0.5250 \\
FI & nsu-ai & 0.4922 & 0.5874 & 0.5312 \\
FI & BlueToad & 0.4906 & 0.5694 & 0.6377 \\
FI & DeepPavlov & 0.4821 & 0.5845 & 0.9782 \\
FI & Swushroomsia & 0.4298 & 0.4955 & 0.7400 \\
FI & TU Munich & 0.4121 & 0.4042 & 0.9986 \\
FI & Howard University - AI4PC & 0.3433 & 0.3996 & 0.5857 \\
FI & uir-cis & 0.3366 & 0.2459 & 0.5635 \\
FI & keepitsimple & 0.3323 & 0.4554 & 1.0000 \\
FI & REFIND & 0.1986 & 0.5025 & 1.0000 \\
FI & \emph{Baseline (neural)} & 0.0924 & 0.0042 & 0.9879 \\
FI & NLP\_CIMAT & 0.0418 & 0.3673 & 0.9928 \\
FI & S1mT5v-FMI & 0.0014 & 0.0000 & 0.6301 \\
FI & \emph{Baseline (mark all)} & 0.0000 & 0.4857 & 0.0000 \\
FI & tsotsalab & 0.0000 & 0.4857 & 0.0000 \\
FI & FENJI & 0.0000 & 0.0941 & 0.0000 \\
FI & \emph{Baseline (mark none)} & 0.0000 & 0.0000 &  \\
\midrule
FR & Deloitte & 0.6187 & 0.6469 & 0.6744 \\
FR & AILS-NTUA & 0.6103 & 0.5812 & 0.6102 \\
FR & UCSC & 0.6041 & 0.5812 & 0.7467 \\
FR & Swushroomsia & 0.5908 & 0.4422 & 0.8283 \\
FR & CCNU & 0.5724 & 0.4823 & 0.6038 \\
FR & SmurfCat & 0.5661 & 0.5269 & 0.6639 \\
FR & MSA & 0.5553 & 0.6195 & 0.6668 \\
FR & DeepPavlov & 0.5440 & 0.5831 & 0.6721 \\
FR & Team Cantharellus & 0.5317 & 0.5147 & 0.7363 \\
FR & TUM-MiKaNi & 0.5157 & 0.6314 & 0.8100 \\
FR & TrustAI & 0.4992 & 0.3799 & 0.6345 \\
FR & tsotsalab & 0.4910 & 0.4836 & 0.5531 \\
FR & LCTeam & 0.4883 & 0.5634 & 0.5960 \\
FR & NCL-UoR & 0.4823 & 0.3571 & 0.6846 \\
FR & UZH & 0.4669 & 0.4860 & 0.8853 \\
FR & nsu-ai & 0.4339 & 0.5181 & 0.9411 \\
FR & UMUTEAM & 0.4117 & 0.3200 & 0.4951 \\
FR & ATLANTIS & 0.4117 & 0.5190 & 0.6909 \\
FR & Howard University - AI4PC & 0.3990 & 0.4164 & 0.7127 \\
FR & BlueToad & 0.3797 & 0.4385 & 0.8446 \\
FR & TU Munich & 0.3484 & 0.4152 & 0.6629 \\
FR & HalluSearch & 0.3365 & 0.4366 & 0.9264 \\
FR & uir-cis & 0.2873 & 0.2286 & 0.6152 \\
FR & keepitsimple & 0.2756 & 0.4651 & 0.9996 \\
FR & REFIND & 0.1530 & 0.2120 & 0.9623 \\
FR & NLP\_CIMAT & 0.0898 & 0.3310 & 0.9759 \\
FR & HalluciSeekers & 0.0447 & 0.0500 & 0.9658 \\
FR & \emph{Baseline (neural)} & 0.0208 & 0.0022 & 0.9444 \\
FR & \emph{Baseline (mark all)} & 0.0000 & 0.4543 & 0.0000 \\
FR & FENJI & 0.0000 & 0.0844 & 0.0000 \\
FR & \emph{Baseline (mark none)} & 0.0000 & 0.0000 & 0.0000 \\
FR & S1mT5v-FMI & 0.0000 & 0.0000 & 1.0000 \\
FR & FunghiFunghi & -0.1521 & 0.3095 &  \\
\midrule
HI & CCNU & 0.7847 & 0.7466 & 0.7038 \\
HI & UCSC & 0.7746 & 0.6732 & 0.7657 \\
HI & AILS-NTUA & 0.7602 & 0.7259 & 0.6763 \\
HI & SmurfCat & 0.7502 & 0.7064 & 0.8911 \\
HI & DeepPavlov & 0.7320 & 0.5117 & 0.6111 \\
HI & MSA & 0.7252 & 0.6842 & 0.8703 \\
HI & Team Cantharellus & 0.6945 & 0.6270 & 0.6329 \\
HI & BlueToad & 0.6844 & 0.6447 & 0.5168 \\
HI & NCL-UoR & 0.6830 & 0.6286 & 0.6870 \\
HI & UZH & 0.6687 & 0.6377 & 0.8632 \\
HI & Deloitte & 0.6391 & 0.6322 & 0.9935 \\
HI & uir-cis & 0.5586 & 0.0613 & 0.7124 \\
HI & TUM-MiKaNi & 0.5409 & 0.5737 & 0.7515 \\
HI & HalluSearch & 0.5195 & 0.5265 & 0.5977 \\
HI & LCTeam & 0.5122 & 0.6601 & 0.6684 \\
HI & TrustAI & 0.5050 & 0.3144 & 0.7519 \\
HI & Swushroomsia & 0.4789 & 0.4534 & 0.7191 \\
HI & nsu-ai & 0.4497 & 0.4315 & 0.6228 \\
HI & UMUTEAM & 0.4386 & 0.4510 & 0.9992 \\
HI & keepitsimple & 0.3508 & 0.3598 & 0.7688 \\
HI & TU Munich & 0.3297 & 0.2807 & 0.6292 \\
HI & Howard University - AI4PC & 0.3217 & 0.2586 & 1.0000 \\
HI & \emph{Baseline (neural)} & 0.1429 & 0.0029 & 1.0000 \\
HI & \emph{Baseline (mark all)} & 0.0000 & 0.2711 & 0.0000 \\
HI & tsotsalab & 0.0000 & 0.2711 & 0.0000 \\
HI & FENJI & 0.0000 & 0.0000 & 0.0000 \\
HI & \emph{Baseline (mark none)} & 0.0000 & 0.0000 &  \\
\midrule
IT & AILS-NTUA & 0.8195 & 0.7660 & 0.9316 \\
IT & UCSC & 0.7944 & 0.7509 & 0.9338 \\
IT & NCL-UoR & 0.7637 & 0.6547 & 0.5213 \\
IT & SmurfCat & 0.7628 & 0.7255 & 0.5826 \\
IT & MSA & 0.7587 & 0.7289 & 0.7341 \\
IT & CCNU & 0.7458 & 0.6944 & 0.6055 \\
IT & Swushroomsia & 0.7394 & 0.7149 & 0.8699 \\
IT & Team Cantharellus & 0.7118 & 0.6907 & 0.6501 \\
IT & UZH & 0.7016 & 0.6833 & 0.9027 \\
IT & BlueToad & 0.6675 & 0.6388 & 0.6914 \\
IT & Deloitte & 0.6547 & 0.6253 & 0.9981 \\
IT & TUM-MiKaNi & 0.6233 & 0.6781 & 0.9968 \\
IT & HalluSearch & 0.5604 & 0.5484 & 0.6117 \\
IT & DeepPavlov & 0.5529 & 0.5280 & 0.5982 \\
IT & LCTeam & 0.5487 & 0.7013 & 0.9406 \\
IT & uir-cis & 0.4991 & 0.3967 & 0.7422 \\
IT & TrustAI & 0.4760 & 0.2077 & 0.8149 \\
IT & UMUTEAM & 0.4601 & 0.4413 & 0.8251 \\
IT & nsu-ai & 0.4402 & 0.4396 & 0.8460 \\
IT & TU Munich & 0.4210 & 0.3319 & 0.8165 \\
IT & Howard University - AI4PC & 0.4021 & 0.2675 & 0.6950 \\
IT & keepitsimple & 0.3860 & 0.4009 & 0.9994 \\
IT & REFIND & 0.2423 & 0.3255 & 1.0000 \\
IT & NLP\_CIMAT & 0.0894 & 0.1696 & 0.6335 \\
IT & \emph{Baseline (neural)} & 0.0800 & 0.0104 & 0.9995 \\
IT & HalluciSeekers & 0.0242 & 0.0350 & 0.9263 \\
IT & \emph{Baseline (mark all)} & 0.0000 & 0.2826 & 0.0000 \\
IT & tsotsalab & 0.0000 & 0.2826 & 0.0000 \\
IT & FENJI & 0.0000 & 0.2765 & 0.0000 \\
IT & \emph{Baseline (mark none)} & 0.0000 & 0.0000 & 1.0000 \\
IT & FunghiFunghi & -0.2116 & 0.2111 &  \\
\midrule
SV & AILS-NTUA & 0.5622 & 0.6009 & 0.7148 \\
SV & MSA & 0.5486 & 0.6071 & 0.6811 \\
SV & Deloitte & 0.5374 & 0.6220 & 0.7116 \\
SV & NCL-UoR & 0.5225 & 0.5234 & 0.5271 \\
SV & UCSC & 0.5204 & 0.6423 & 0.6072 \\
SV & CCNU & 0.5129 & 0.4961 & 0.6709 \\
SV & SmurfCat & 0.5007 & 0.6174 & 0.9144 \\
SV & LCTeam & 0.4631 & 0.3016 & 0.8654 \\
SV & UZH & 0.4346 & 0.5263 & 0.5727 \\
SV & HalluSearch & 0.4290 & 0.5622 & 0.5251 \\
SV & BlueToad & 0.4267 & 0.5854 & 0.5685 \\
SV & TrustAI & 0.4219 & 0.1582 & 0.6044 \\
SV & DeepPavlov & 0.4147 & 0.5380 & 0.6592 \\
SV & TUM-MiKaNi & 0.4028 & 0.5614 & 0.6616 \\
SV & UMUTEAM & 0.3936 & 0.4393 & 0.8202 \\
SV & uir-cis & 0.3655 & 0.3080 & 0.7707 \\
SV & nsu-ai & 0.3442 & 0.5478 & 1.0000 \\
SV & TU Munich & 0.2403 & 0.2755 & 0.6705 \\
SV & Swushroomsia & 0.2265 & 0.3549 & 0.6015 \\
SV & keepitsimple & 0.2170 & 0.3967 & 0.9998 \\
SV & \emph{Baseline (neural)} & 0.0968 & 0.0308 & 0.6999 \\
SV & HalluciSeekers & 0.0856 & 0.0575 & 0.5466 \\
SV & NLP\_CIMAT & 0.0823 & 0.1772 & 0.7176 \\
SV & Howard University - AI4PC & 0.0669 & 0.1110 & 0.9976 \\
SV & \emph{Baseline (mark all)} & 0.0136 & 0.5373 & 0.0000 \\
SV & tsotsalab & 0.0136 & 0.5349 & 0.0000 \\
SV & FENJI & 0.0136 & 0.1154 & 0.0000 \\
SV & \emph{Baseline (mark none)} & 0.0136 & 0.0204 & 0.0000 \\
SV & S1mT5v-FMI & 0.0136 & 0.0204 & 1.0000 \\
SV & FunghiFunghi & -0.1177 & 0.4156 &  \\
\midrule
ZH & LCTeam & 0.5171 & 0.5232 & 0.9837 \\
ZH & UMUTEAM & 0.4916 & 0.3875 & 0.7342 \\
ZH & AILS-NTUA & 0.4791 & 0.3083 & 0.6070 \\
ZH & TrustAI & 0.4735 & 0.3423 & 0.7057 \\
ZH & TUM-MiKaNi & 0.4676 & 0.4490 & 0.9411 \\
ZH & MSA & 0.4363 & 0.4631 & 0.5568 \\
ZH & CCNU & 0.4335 & 0.3718 & 0.6708 \\
ZH & HalluSearch & 0.4232 & 0.4534 & 0.5759 \\
ZH & UCSC & 0.4187 & 0.4633 & 0.7076 \\
ZH & Team Cantharellus & 0.4063 & 0.4011 & 0.8344 \\
ZH & NCL-UoR & 0.3830 & 0.3493 & 0.5335 \\
ZH & nsu-ai & 0.3813 & 0.4937 & 0.9025 \\
ZH & UZH & 0.3520 & 0.3993 & 0.5027 \\
ZH & YNU-HPCC & 0.3518 & 0.5540 & 0.5600 \\
ZH & SmurfCat & 0.3457 & 0.4017 & 0.7567 \\
ZH & uir-cis & 0.3278 & 0.1786 & 0.5413 \\
ZH & DeepPavlov & 0.3251 & 0.4849 & 0.5801 \\
ZH & Deloitte & 0.3203 & 0.4479 & 0.9978 \\
ZH & TU Munich & 0.2771 & 0.1750 & 0.9958 \\
ZH & BlueToad & 0.2262 & 0.2783 & 0.9924 \\
ZH & keepitsimple & 0.1601 & 0.4703 & 0.9817 \\
ZH & Howard University - AI4PC & 0.1119 & 0.2152 & 0.7297 \\
ZH & Swushroomsia & 0.0966 & 0.2054 & 0.6454 \\
ZH & \emph{Baseline (neural)} & 0.0884 & 0.0236 & 1.0000 \\
ZH & \emph{Baseline (mark all)} & 0.0000 & 0.4772 & 0.0000 \\
ZH & tsotsalab & 0.0000 & 0.4772 & 0.0000 \\
ZH & FENJI & 0.0000 & 0.0371 & 0.0000 \\
ZH & \emph{Baseline (mark none)} & 0.0000 & 0.0200 & 0.9995 \\
ZH & S1mT5v-FMI & -0.0209 & 0.0619 &  \\
\end{supertabular}
\end{center}

%% file: main.bbl
\begin{thebibliography}{78}
\providecommand{\natexlab}[1]{#1}

\bibitem[{{01. AI} et~al.(2024){01. AI}, Young, Chen, Li, Huang, Zhang, Zhang, Li, Zhu, Chen, Chang, Yu, Liu, Liu, Yue, Yang, Yang, Yu, Xie, Huang, Hu, Ren, Niu, Nie, Xu, Liu, Wang, Cai, Gu, Liu, and Dai}]{ai2024yiopenfoundationmodels}
{01. AI}, Alex Young, Bei Chen, Chao Li, Chengen Huang, Ge~Zhang, Guanwei Zhang, Heng Li, Jiangcheng Zhu, Jianqun Chen, Jing Chang, Kaidong Yu, Peng Liu, Qiang Liu, Shawn Yue, Senbin Yang, Shiming Yang, Tao Yu, Wen Xie, and 12 others. 2024.
\newblock \href {https://arxiv.org/abs/2403.04652} {Yi: Open foundation models by 01.ai}.
\newblock \emph{Preprint}, arXiv:2403.04652.

\bibitem[{Abdallah and El-Beltagy(2025)}]{abdallah-el-beltagy-2025-hallusearch}
Mohamed~A. Abdallah and Samhaa~R. El-Beltagy. 2025.
\newblock {HalluSearch at SemEval-2025} task 3: A search-enhanced {RAG} pipeline for hallucination detection.
\newblock In \emph{Proceedings of the 19th International Workshop on Semantic Evaluation (SemEval-2025)}.

\bibitem[{Abdi et~al.(2025)Abdi, Hassani, Kinds, Strijbis, and Terpstra}]{abdi-etal-2025-hallurag}
Silvana Abdi, Mahrokh Hassani, Rosalien Kinds, Timo Strijbis, and Roman Terpstra. 2025.
\newblock {HalluRAG-RUG at SemEval-2025} task 3: Using retrieval-augmented generation for hallucination detection in model outputs.
\newblock In \emph{Proceedings of the 19th International Workshop on Semantic Evaluation (SemEval-2025)}.

\bibitem[{Alberts et~al.(2025)Alberts, Bruinier, de~Palm, Paetzelt, and Varecha}]{alberts-etal-2025-fenji}
Flor Alberts, Ivo~B.A. Bruinier, Nathalie de~Palm, Justin Paetzelt, and Erik Varecha. 2025.
\newblock {FENJI at SemEval-2025} task 3: Retrieval-augmented generation and hallucination span detection.
\newblock In \emph{Proceedings of the 19th International Workshop on Semantic Evaluation (SemEval-2025)}.

\bibitem[{Almazrouei et~al.(2023)Almazrouei, Alobeidli, Alshamsi, Cappelli, Cojocaru, Debbah, Goffinet, Heslow, Launay, Malartic, Noune, Pannier, and Penedo}]{falcon40b}
Ebtesam Almazrouei, Hamza Alobeidli, Abdulaziz Alshamsi, Alessandro Cappelli, Ruxandra Cojocaru, Merouane Debbah, Etienne Goffinet, Daniel Heslow, Julien Launay, Quentin Malartic, Badreddine Noune, Baptiste Pannier, and Guilherme Penedo. 2023.
\newblock {Falcon-40B}: an open large language model with state-of-the-art performance.

\bibitem[{Anschütz et~al.(2025)Anschütz, Gikalo, Herbster, and Groh}]{anschutz-etal-2025-tum}
Miriam Anschütz, Ekaterina Gikalo, Niklas Herbster, and Georg Groh. 2025.
\newblock {TUM-MiKaNi at SemEval-2025} task 3: Towards multilingual and knowledge-aware non-factual hallucination identification.
\newblock In \emph{Proceedings of the 19th International Workshop on Semantic Evaluation (SemEval-2025)}.

\bibitem[{Aroyo and Welty(2015)}]{Aroyo2015}
Lora Aroyo and Chris Welty. 2015.
\newblock \href {https://doi.org/10.1609/aimag.v36i1.2564} {Truth is a lie: Crowd truth and the seven myths of human annotation}.
\newblock \emph{AI Magazine}, 36(1):15–24.

\bibitem[{Aryabumi et~al.(2024)Aryabumi, Dang, Talupuru, Dash, Cairuz, Lin, Venkitesh, Smith, Campos, Tan, Marchisio, Bartolo, Ruder, Locatelli, Kreutzer, Frosst, Gomez, Blunsom, Fadaee, Üstün, and Hooker}]{aryabumi2024aya23openweight}
Viraat Aryabumi, John Dang, Dwarak Talupuru, Saurabh Dash, David Cairuz, Hangyu Lin, Bharat Venkitesh, Madeline Smith, Jon~Ander Campos, Yi~Chern Tan, Kelly Marchisio, Max Bartolo, Sebastian Ruder, Acyr Locatelli, Julia Kreutzer, Nick Frosst, Aidan Gomez, Phil Blunsom, Marzieh Fadaee, and 2 others. 2024.
\newblock \href {https://arxiv.org/abs/2405.15032} {Aya 23: Open weight releases to further multilingual progress}.
\newblock \emph{Preprint}, arXiv:2405.15032.

\bibitem[{Aryal and Akomoize(2025)}]{aryal-akomoize-2025-howard}
Saurav~K. Aryal and Mildness Akomoize. 2025.
\newblock {Howard University - AI4PC at SemEval-2025} task 3: Logit-based supervised token classification for multilingual hallucination span identification using {XGBOD}.
\newblock In \emph{Proceedings of the 19th International Workshop on Semantic Evaluation (SemEval-2025)}.

\bibitem[{Augenstein et~al.(2024)Augenstein, Baldwin, Cha, Chakraborty, Ciampaglia, Corney, DiResta, Ferrara, Hale, Halevy, Hovy, Ji, Menczer, Miguez, Nakov, Scheufele, Sharma, and Zagni}]{Augenstein2024}
Isabelle Augenstein, Timothy Baldwin, Meeyoung Cha, Tanmoy Chakraborty, Giovanni~Luca Ciampaglia, David Corney, Renee DiResta, Emilio Ferrara, Scott Hale, Alon Halevy, Eduard Hovy, Heng Ji, Filippo Menczer, Ruben Miguez, Preslav Nakov, Dietram Scheufele, Shivam Sharma, and Giovanni Zagni. 2024.
\newblock \href {https://doi.org/10.1038/s42256-024-00881-z} {Factuality challenges in the era of large language models and opportunities for fact-checking}.
\newblock \emph{Nature Machine Intelligence}, 6(8):852--863.

\bibitem[{Baan et~al.(2022)Baan, Aziz, Plank, and Fernandez}]{baan-etal-2022-stop}
Joris Baan, Wilker Aziz, Barbara Plank, and Raquel Fernandez. 2022.
\newblock \href {https://doi.org/10.18653/v1/2022.emnlp-main.124} {Stop measuring calibration when humans disagree}.
\newblock In \emph{Proceedings of the 2022 Conference on Empirical Methods in Natural Language Processing}, pages 1892--1915, Abu Dhabi, United Arab Emirates. Association for Computational Linguistics.

\bibitem[{Bai et~al.(2023)Bai, Bai, Chu, Cui, Dang, Deng, Fan, Ge, Han, Huang, Hui, Ji, Li, Lin, Lin, Liu, Liu, Lu, Lu, Ma, Men, Ren, Ren, Tan, Tan, Tu, Wang, Wang, Wang, Wu, Xu, Xu, Yang, Yang, Yang, Yang, Yao, Yu, Yuan, Yuan, Zhang, Zhang, Zhang, Zhang, Zhou, Zhou, Zhou, and Zhu}]{qwen}
Jinze Bai, Shuai Bai, Yunfei Chu, Zeyu Cui, Kai Dang, Xiaodong Deng, Yang Fan, Wenbin Ge, Yu~Han, Fei Huang, Binyuan Hui, Luo Ji, Mei Li, Junyang Lin, Runji Lin, Dayiheng Liu, Gao Liu, Chengqiang Lu, Keming Lu, and 29 others. 2023.
\newblock Qwen technical report.

\bibitem[{Bala et~al.(2025)Bala, Abubakar, Abubakar, Bichi, Ahmad, Sani, Abdulmumin, Muhammad, and Ahmad}]{bala-etal-2025-hausanlp}
Maryam Bala, Amina~Imam Abubakar, Abdulhamid Abubakar, Abdulkadir~Shehu Bichi, Hafsa~Kabir Ahmad, Sani~Abdullahi Sani, Idris Abdulmumin, Shamsuddeen~Hassan Muhammad, and Ibrahim~Said Ahmad. 2025.
\newblock {HausaNLP at SemEval-2025} task 3: Towards a fine-grained model-aware hallucination detection.
\newblock In \emph{Proceedings of the 19th International Workshop on Semantic Evaluation (SemEval-2025)}.

\bibitem[{Ballout et~al.(2025)Ballout, Jansma, Koops, and Zhou}]{ballout-etal-2025-funghifunghi}
Tariq Ballout, Pieter Jansma, Nander Koops, and Yong~Hui Zhou. 2025.
\newblock {FunghiFunghi at SemEval-2025} task 3: {Mu-SHROOM}, the multilingual shared-task on hallucinations and related observable overgeneration mistakes.
\newblock In \emph{Proceedings of the 19th International Workshop on Semantic Evaluation (SemEval-2025)}.

\bibitem[{Cai et~al.(2024)Cai, Cao, Chen, Chen, Chen, Chen, Chen, Chen, Chen, Chu, Dong, Duan, Fan, Fei, Gao, Ge, Gu, Gu, Gui, Guo, Guo, He, Hu, Huang, Jiang, Jiao, Jin, Lei, Li, Li, Li, Li, Li, Li, Liu, Liu, Hong, Liu, Liu, Liu, Lv, Lv, Lv, Ma, Ma, Ma, Ning, Ouyang, Qiu, Qu, Shang, Shao, Song, Song, Sui, Sun, Sun, Tang, Wang, Wang, Wang, Wang, Wang, Wang, Wang, Wei, Weng, Wu, Xiong, Xu, Xu, Yan, Yan, Yang, Ye, Ying, Yu, Yu, Zang, Zhang, Zhang, Zhang, Zhang, Zhang, Zhang, Zhang, Zhang, Zhang, Zhang, Zhang, Zhao, Zhao, Zhao, Zhou, Zhou, Zhuo, Zou, Qiu, Qiao, and Lin}]{cai2024internlm2}
Zheng Cai, Maosong Cao, Haojiong Chen, Kai Chen, Keyu Chen, Xin Chen, Xun Chen, Zehui Chen, Zhi Chen, Pei Chu, Xiaoyi Dong, Haodong Duan, Qi~Fan, Zhaoye Fei, Yang Gao, Jiaye Ge, Chenya Gu, Yuzhe Gu, Tao Gui, and 81 others. 2024.
\newblock \href {https://arxiv.org/abs/2403.17297} {Internlm2 technical report}.
\newblock \emph{Preprint}, arXiv:2403.17297.

\bibitem[{Chandler et~al.(2025)Chandler, Abburi, Bhattacharya, Bowen, and Pudota}]{chandler-etal-2025-deloitte}
Alex Chandler, Harika Abburi, Sanmitra Bhattacharya, Edward Bowen, and Nirmala Pudota. 2025.
\newblock {Deloitte (Drocks) at SemEval-2025} task 3: Fine-grained multi-lingual hallucination detection using internal {LLM} weights.
\newblock In \emph{Proceedings of the 19th International Workshop on Semantic Evaluation (SemEval-2025)}.

\bibitem[{Chen et~al.(2025)Chen, Wang, and Zhang}]{chen-etal-2025-ynu}
Shen Chen, Jin Wang, and Xuejie Zhang. 2025.
\newblock {YNU-HPCC at SemEval-2025} task3: Leveraging zero-shot learning for hallucination detection.
\newblock In \emph{Proceedings of the 19th International Workshop on Semantic Evaluation (SemEval-2025)}.

\bibitem[{Chen et~al.(2024)Chen, Song, Gui, Wang, Zhang, Jiang, Huang, Lyu, Zhang, and Chen}]{ijcai2024p0687}
Xiang Chen, Duanzheng Song, Honghao Gui, Chenxi Wang, Ningyu Zhang, Yong Jiang, Fei Huang, Chengfei Lyu, Dan Zhang, and Huajun Chen. 2024.
\newblock \href {https://doi.org/10.24963/ijcai.2024/687} {Factchd: Benchmarking fact-conflicting hallucination detection}.
\newblock In \emph{Proceedings of the Thirty-Third International Joint Conference on Artificial Intelligence, {IJCAI-24}}, pages 6216--6224. International Joint Conferences on Artificial Intelligence Organization.
\newblock Main Track.

\bibitem[{Cheng et~al.(2023)Cheng, Sun, Zhang, Wang, Liu, Zhang, He, Huang, Yin, Chen, and Qiu}]{cheng2023evaluatinghallucinationschineselarge}
Qinyuan Cheng, Tianxiang Sun, Wenwei Zhang, Siyin Wang, Xiangyang Liu, Mozhi Zhang, Junliang He, Mianqiu Huang, Zhangyue Yin, Kai Chen, and Xipeng Qiu. 2023.
\newblock \href {https://arxiv.org/abs/2310.03368} {Evaluating hallucinations in chinese large language models}.
\newblock \emph{Preprint}, arXiv:2310.03368.

\bibitem[{Creo et~al.(2025)Creo, Cerezo-Costas, Lagos, and Doval}]{creo-etal-2025-cogumelo}
Aldan Creo, Héctor Cerezo-Costas, Maximiliano~Hormazábal Lagos, and Pedro~Alonso Doval. 2025.
\newblock {COGUMELO at SemEval-2025} task 3: A synthetic approach to detecting hallucinations in language models based on named entity recognition.
\newblock In \emph{Proceedings of the 19th International Workshop on Semantic Evaluation (SemEval-2025)}.

\bibitem[{Elchafei and Abu-Elkheir(2025)}]{elchafei-abu-elkheir-2025-hallucination}
Passant Elchafei and Mervat Abu-Elkheir. 2025.
\newblock {Hallucination Detectives at SemEval-2025} task 3: Span-level hallucination detection for {LLM}-generated answers.
\newblock In \emph{Proceedings of the 19th International Workshop on Semantic Evaluation (SemEval-2025)}.

\bibitem[{Farquhar et~al.(2024)Farquhar, Kossen, Kuhn, and Gal}]{Farquhar2024}
Sebastian Farquhar, Jannik Kossen, Lorenz Kuhn, and Yarin Gal. 2024.
\newblock \href {https://doi.org/10.1038/s41586-024-07421-0} {Detecting hallucinations in large language models using semantic entropy}.
\newblock \emph{Nature}, 630(8017):625--630.

\bibitem[{Faysse et~al.(2024)Faysse, Fernandes, Guerreiro, Loison, Alves, Corro, Boizard, Alves, Rei, Martins, Casademunt, Yvon, Martins, Viaud, Hudelot, and Colombo}]{faysse2024croissantllm}
Manuel Faysse, Patrick Fernandes, Nuno~M. Guerreiro, António Loison, Duarte~M. Alves, Caio Corro, Nicolas Boizard, João Alves, Ricardo Rei, Pedro~H. Martins, Antoni~Bigata Casademunt, François Yvon, André F.~T. Martins, Gautier Viaud, Céline Hudelot, and Pierre Colombo. 2024.
\newblock \href {https://arxiv.org/abs/2402.00786} {{CroissantLLM}: A truly bilingual french-english language model}.
\newblock \emph{Preprint}, arXiv:2402.00786.

\bibitem[{Grattafiori et~al.(2024)Grattafiori, Dubey, Jauhri, Pandey, Kadian, Al-Dahle, Letman, Mathur, Schelten, Vaughan, Yang, Fan, Goyal, Hartshorn, Yang, Mitra, Sravankumar, Korenev, Hinsvark, Rao, Zhang, Rodriguez, Gregerson, Spataru, Roziere, Biron, Tang, Chern, Caucheteux, Nayak, Bi, Marra, McConnell, Keller, Touret, Wu, Wong, Ferrer, Nikolaidis, Allonsius, Song, Pintz, Livshits, Wyatt, Esiobu, Choudhary, Mahajan, Garcia-Olano, Perino, Hupkes, Lakomkin, AlBadawy, Lobanova, Dinan, Smith, Radenovic, Guzmán, Zhang, Synnaeve, Lee, Anderson, Thattai, Nail, Mialon, Pang, Cucurell, Nguyen, Korevaar, Xu, Touvron, Zarov, Ibarra, Kloumann, Misra, Evtimov, Zhang, Copet, Lee, Geffert, Vranes, Park, Mahadeokar, Shah, van~der Linde, Billock, Hong, Lee, Fu, Chi, Huang, Liu, Wang, Yu, Bitton, Spisak, Park, Rocca, Johnstun, Saxe, Jia, Alwala, Prasad, Upasani, Plawiak, Li, Heafield, Stone, El-Arini, Iyer, Malik, Chiu, Bhalla, Lakhotia, Rantala-Yeary, van~der Maaten, Chen, Tan, Jenkins, Martin, Madaan, Malo, Blecher,
  Landzaat, de~Oliveira, Muzzi, Pasupuleti, Singh, Paluri, Kardas, Tsimpoukelli, Oldham, Rita, Pavlova, Kambadur, Lewis, Si, Singh, Hassan, Goyal, Torabi, Bashlykov, Bogoychev, Chatterji, Zhang, Duchenne, Çelebi, Alrassy, Zhang, Li, Vasic, Weng, Bhargava, Dubal, Krishnan, Koura, Xu, He, Dong, Srinivasan, Ganapathy, Calderer, Cabral, Stojnic, Raileanu, Maheswari, Girdhar, Patel, Sauvestre, Polidoro, Sumbaly, Taylor, Silva, Hou, Wang, Hosseini, Chennabasappa, Singh, Bell, Kim, Edunov, Nie, Narang, Raparthy, Shen, Wan, Bhosale, Zhang, Vandenhende, Batra, Whitman, Sootla, Collot, Gururangan, Borodinsky, Herman, Fowler, Sheasha, Georgiou, Scialom, Speckbacher, Mihaylov, Xiao, Karn, Goswami, Gupta, Ramanathan, Kerkez, Gonguet, Do, Vogeti, Albiero, Petrovic, Chu, Xiong, Fu, Meers, Martinet, Wang, Wang, Tan, Xia, Xie, Jia, Wang, Goldschlag, Gaur, Babaei, Wen, Song, Zhang, Li, Mao, Coudert, Yan, Chen, Papakipos, Singh, Srivastava, Jain, Kelsey, Shajnfeld, Gangidi, Victoria, Goldstand, Menon, Sharma, Boesenberg,
  Baevski, Feinstein, Kallet, Sangani, Teo, Yunus, Lupu, Alvarado, Caples, Gu, Ho, Poulton, Ryan, Ramchandani, Dong, Franco, Goyal, Saraf, Chowdhury, Gabriel, Bharambe, Eisenman, Yazdan, James, Maurer, Leonhardi, Huang, Loyd, Paola, Paranjape, Liu, Wu, Ni, Hancock, Wasti, Spence, Stojkovic, Gamido, Montalvo, Parker, Burton, Mejia, Liu, Wang, Kim, Zhou, Hu, Chu, Cai, Tindal, Feichtenhofer, Gao, Civin, Beaty, Kreymer, Li, Adkins, Xu, Testuggine, David, Parikh, Liskovich, Foss, Wang, Le, Holland, Dowling, Jamil, Montgomery, Presani, Hahn, Wood, Le, Brinkman, Arcaute, Dunbar, Smothers, Sun, Kreuk, Tian, Kokkinos, Ozgenel, Caggioni, Kanayet, Seide, Florez, Schwarz, Badeer, Swee, Halpern, Herman, Sizov, Guangyi, Zhang, Lakshminarayanan, Inan, Shojanazeri, Zou, Wang, Zha, Habeeb, Rudolph, Suk, Aspegren, Goldman, Zhan, Damlaj, Molybog, Tufanov, Leontiadis, Veliche, Gat, Weissman, Geboski, Kohli, Lam, Asher, Gaya, Marcus, Tang, Chan, Zhen, Reizenstein, Teboul, Zhong, Jin, Yang, Cummings, Carvill, Shepard, McPhie,
  Torres, Ginsburg, Wang, Wu, U, Saxena, Khandelwal, Zand, Matosich, Veeraraghavan, Michelena, Li, Jagadeesh, Huang, Chawla, Huang, Chen, Garg, A, Silva, Bell, Zhang, Guo, Yu, Moshkovich, Wehrstedt, Khabsa, Avalani, Bhatt, Mankus, Hasson, Lennie, Reso, Groshev, Naumov, Lathi, Keneally, Liu, Seltzer, Valko, Restrepo, Patel, Vyatskov, Samvelyan, Clark, Macey, Wang, Hermoso, Metanat, Rastegari, Bansal, Santhanam, Parks, White, Bawa, Singhal, Egebo, Usunier, Mehta, Laptev, Dong, Cheng, Chernoguz, Hart, Salpekar, Kalinli, Kent, Parekh, Saab, Balaji, Rittner, Bontrager, Roux, Dollar, Zvyagina, Ratanchandani, Yuvraj, Liang, Alao, Rodriguez, Ayub, Murthy, Nayani, Mitra, Parthasarathy, Li, Hogan, Battey, Wang, Howes, Rinott, Mehta, Siby, Bondu, Datta, Chugh, Hunt, Dhillon, Sidorov, Pan, Mahajan, Verma, Yamamoto, Ramaswamy, Lindsay, Lindsay, Feng, Lin, Zha, Patil, Shankar, Zhang, Zhang, Wang, Agarwal, Sajuyigbe, Chintala, Max, Chen, Kehoe, Satterfield, Govindaprasad, Gupta, Deng, Cho, Virk, Subramanian, Choudhury,
  Goldman, Remez, Glaser, Best, Koehler, Robinson, Li, Zhang, Matthews, Chou, Shaked, Vontimitta, Ajayi, Montanez, Mohan, Kumar, Mangla, Ionescu, Poenaru, Mihailescu, Ivanov, Li, Wang, Jiang, Bouaziz, Constable, Tang, Wu, Wang, Wu, Gao, Kleinman, Chen, Hu, Jia, Qi, Li, Zhang, Zhang, Adi, Nam, Yu, Wang, Zhao, Hao, Qian, Li, He, Rait, DeVito, Rosnbrick, Wen, Yang, Zhao, and Ma}]{grattafiori2024llama3herdmodels}
Aaron Grattafiori, Abhimanyu Dubey, Abhinav Jauhri, Abhinav Pandey, Abhishek Kadian, Ahmad Al-Dahle, Aiesha Letman, Akhil Mathur, Alan Schelten, Alex Vaughan, Amy Yang, Angela Fan, Anirudh Goyal, Anthony Hartshorn, Aobo Yang, Archi Mitra, Archie Sravankumar, Artem Korenev, Arthur Hinsvark, and 542 others. 2024.
\newblock \href {https://arxiv.org/abs/2407.21783} {The llama 3 herd of models}.
\newblock \emph{Preprint}, arXiv:2407.21783.

\bibitem[{Gu et~al.(2024)Gu, Ji, Zhang, Lyu, Lin, and Chen}]{gu2024anahv}
Yuzhe Gu, Ziwei Ji, Wenwei Zhang, Chengqi Lyu, Dahua Lin, and Kai Chen. 2024.
\newblock \href {https://openreview.net/forum?id=NrwASKGm7A} {{ANAH}-v2: Scaling analytical hallucination annotation of large language models}.
\newblock In \emph{The Thirty-eighth Annual Conference on Neural Information Processing Systems}.

\bibitem[{Guerreiro et~al.(2023)Guerreiro, Voita, and Martins}]{guerreiro-etal-2023-looking}
Nuno~M. Guerreiro, Elena Voita, and Andr{\'e} Martins. 2023.
\newblock \href {https://doi.org/10.18653/v1/2023.eacl-main.75} {Looking for a needle in a haystack: A comprehensive study of hallucinations in neural machine translation}.
\newblock In \emph{Proceedings of the 17th Conference of the European Chapter of the Association for Computational Linguistics}, pages 1059--1075, Dubrovnik, Croatia. Association for Computational Linguistics.

\bibitem[{Heerema et~al.(2025)Heerema, Krooneman, van Loon, Top, and Voors}]{heerema-etal-2025-raggedyfive}
Wessel Heerema, Collin Krooneman, Simon van Loon, Jelmer Top, and Maurice Voors. 2025.
\newblock {RaggedyFive at SemEval-2025} task 3: Hallucination span detection using unverifiable answer detection.
\newblock In \emph{Proceedings of the 19th International Workshop on Semantic Evaluation (SemEval-2025)}.

\bibitem[{Hicks et~al.(2024)Hicks, Humphries, and Slater}]{Hicks2024}
Michael~Townsen Hicks, James Humphries, and Joe Slater. 2024.
\newblock \href {https://doi.org/10.1007/s10676-024-09775-5} {Chatgpt is bullshit}.
\newblock \emph{Ethics and Information Technology}, 26(2).

\bibitem[{Hikal et~al.(2025)Hikal, Nasreldin, and Hamdi}]{hikal-etal-2025-msa}
Baraa Hikal, Ahmed Nasreldin, and Ali Hamdi. 2025.
\newblock {MSA at SemEval-2025} task 3: High quality weak labeling and {LLM} ensemble verification for multilingual hallucination detection.
\newblock In \emph{Proceedings of the 19th International Workshop on Semantic Evaluation (SemEval-2025)}.

\bibitem[{Hong et~al.(2025)Hong, Markchom, Xu, Wu, and Liang}]{hong-etal-2025-ncl}
Jiaying Hong, Thanet Markchom, Jianfei Xu, Tong Wu, and Huizhi Liang. 2025.
\newblock {NCL-UoR at SemEval-2025} task 3: Detecting multilingual hallucination and related observable overgeneration text spans with modified {R}ef{C}hecker and modified {S}efl{C}heck{GPT}.
\newblock In \emph{Proceedings of the 19th International Workshop on Semantic Evaluation (SemEval-2025)}.

\bibitem[{Huang et~al.(2025{\natexlab{a}})Huang, Zhao, Zhao, Chen, Zhao, Lin, Chen, and Li}]{huang-etal-2025-uir}
Jia Huang, Shuli Zhao, Yaru Zhao, Tao Chen, Weijia Zhao, Hangui Lin, Yiyang Chen, and Binyang Li. 2025{\natexlab{a}}.
\newblock {uir-cis at SemEval-2025} task 3: Detection of hallucinations in generated text.
\newblock In \emph{Proceedings of the 19th International Workshop on Semantic Evaluation (SemEval-2025)}.

\bibitem[{Huang et~al.(2024)Huang, Yu, Ma, Zhong, Feng, Wang, Chen, Peng, Feng, Qin, and Liu}]{huang-etal-2024-survey}
Lei Huang, Weijiang Yu, Weitao Ma, Weihong Zhong, Zhangyin Feng, Haotian Wang, Qianglong Chen, Weihua Peng, Xiaocheng Feng, Bing Qin, and Ting Liu. 2024.
\newblock \href {https://doi.org/10.1145/3703155} {A survey on hallucination in large language models: Principles, taxonomy, challenges, and open questions}.
\newblock \emph{ACM Trans. Inf. Syst.}, 43(2).

\bibitem[{Huang et~al.(2025{\natexlab{b}})Huang, He, Huang, Anandan, Chakraborty, and Lane}]{huang-etal-2025-ucsc}
Sicong Huang, Jincheng He, Shiyuan Huang, Karthik~Raja Anandan, Arkajyoti Chakraborty, and Ian Lane. 2025{\natexlab{b}}.
\newblock {UCSC at SemEval-2025} task 3: Context, models and prompt optimization for automated hallucination detection in {LLM} output.
\newblock In \emph{Proceedings of the 19th International Workshop on Semantic Evaluation (SemEval-2025)}.

\bibitem[{Ji et~al.(2023)Ji, Lee, Frieske, Yu, Su, Xu, Ishii, Bang, Madotto, and Fung}]{ji-etal-2023-survey}
Ziwei Ji, Nayeon Lee, Rita Frieske, Tiezheng Yu, Dan Su, Yan Xu, Etsuko Ishii, Ye~Jin Bang, Andrea Madotto, and Pascale Fung. 2023.
\newblock \href {https://doi.org/10.1145/3571730} {Survey of hallucination in natural language generation}.
\newblock \emph{ACM Comput. Surv.}, 55(12).

\bibitem[{Karkani et~al.(2025)Karkani, Lymperaiou, Filandrianos, Spanos, Voulodimos, and Stamou}]{karkani-etal-2025-ails}
Dimitra Karkani, Maria Lymperaiou, George Filandrianos, Nikolaos Spanos, Athanasios Voulodimos, and Giorgos Stamou. 2025.
\newblock {AILS-NTUA at SemEval-2025} task 3: Leveraging large language models and translation strategies for multilingual hallucination detection.
\newblock In \emph{Proceedings of the 19th International Workshop on Semantic Evaluation (SemEval-2025)}.

\bibitem[{Kobus et~al.(2025)Kobus, Lancelot, Martin, and Amer}]{kobus-etal-2025-atlantis}
Catherine Kobus, Francois Lancelot, Marion-Cecile Martin, and Nawal~Ould Amer. 2025.
\newblock {ATLANTIS at SemEval-2025} task 3: Detecting hallucinated text spans in question answering.
\newblock In \emph{Proceedings of the 19th International Workshop on Semantic Evaluation (SemEval-2025)}.

\bibitem[{Lee and Yu(2025)}]{lee-yu-2025-refind}
DongGeon Lee and Hwanjo Yu. 2025.
\newblock {REFIND at SemEval-2025} task 3: Retrieval-augmented factuality hallucination detection in large language models.
\newblock In \emph{Proceedings of the 19th International Workshop on Semantic Evaluation (SemEval-2025)}.

\bibitem[{Lee et~al.(2018)Lee, Firat, Agarwal, Fannjiang, and Sussillo}]{Lee2018HallucinationsIN}
Katherine Lee, Orhan Firat, Ashish Agarwal, Clara Fannjiang, and David Sussillo. 2018.
\newblock \href {https://api.semanticscholar.org/CorpusID:53593076} {Hallucinations in neural machine translation}.

\bibitem[{Li et~al.(2023)Li, Cheng, Zhao, Nie, and Wen}]{li-etal-2023-halueval}
Junyi Li, Xiaoxue Cheng, Xin Zhao, Jian-Yun Nie, and Ji-Rong Wen. 2023.
\newblock \href {https://doi.org/10.18653/v1/2023.emnlp-main.397} {{H}alu{E}val: A large-scale hallucination evaluation benchmark for large language models}.
\newblock In \emph{Proceedings of the 2023 Conference on Empirical Methods in Natural Language Processing}, pages 6449--6464, Singapore. Association for Computational Linguistics.

\bibitem[{Liu et~al.(2024)Liu, Liu, Shi, Huang, Wang, Yang, Zhang, Li, and Ma}]{liu2024hallucinationscode}
Fang Liu, Yang Liu, Lin Shi, Houkun Huang, Ruifeng Wang, Zhen Yang, Li~Zhang, Zhongqi Li, and Yuchi Ma. 2024.
\newblock \href {https://arxiv.org/abs/2404.00971} {Exploring and evaluating hallucinations in llm-powered code generation}.
\newblock \emph{Preprint}, arXiv:2404.00971.

\bibitem[{Liu et~al.(2022)Liu, Zhang, Brockett, Mao, Sui, Chen, and Dolan}]{liu-etal-2022-token}
Tianyu Liu, Yizhe Zhang, Chris Brockett, Yi~Mao, Zhifang Sui, Weizhu Chen, and Bill Dolan. 2022.
\newblock \href {https://doi.org/10.18653/v1/2022.acl-long.464} {A token-level reference-free hallucination detection benchmark for free-form text generation}.
\newblock In \emph{Proceedings of the 60th Annual Meeting of the Association for Computational Linguistics (Volume 1: Long Papers)}, pages 6723--6737, Dublin, Ireland. Association for Computational Linguistics.

\bibitem[{Liu and Chen(2025)}]{liu-chen-2025-ccnu}
Xu~Liu and Guanyi Chen. 2025.
\newblock {CCNU at SemEval-2025} task 3: Leveraging internal and external knowledge of large language models for multilingual hallucination annotation.
\newblock In \emph{Proceedings of the 19th International Workshop on Semantic Evaluation (SemEval-2025)}.

\bibitem[{Lopez-Ponce et~al.(2025)Lopez-Ponce, Salas-Jimenez, Juárez-Pérez, Hernández-Bustamente, Bel-Enguix, and Gomez-Adorno}]{lopez-ponce-etal-2025-gil}
Francisco~Fernando Lopez-Ponce, Karla Salas-Jimenez, Adrián Juárez-Pérez, Diego Hernández-Bustamente, Gemma Bel-Enguix, and Helena Gomez-Adorno. 2025.
\newblock {GIL-IIMAS UNAM at SemEval-2025} task 3: {MeSSI}, a multilmodule system to detect hallucinated segments in trivia-like inquiries.
\newblock In \emph{Proceedings of the 19th International Workshop on Semantic Evaluation (SemEval-2025)}.

\bibitem[{Luukkonen et~al.(2024)Luukkonen, Burdge, Zosa, Talman, Komulainen, Hatanpää, Sarlin, and Pyysalo}]{luukkonen2024poro}
Risto Luukkonen, Jonathan Burdge, Elaine Zosa, Aarne Talman, Ville Komulainen, Väinö Hatanpää, Peter Sarlin, and Sampo Pyysalo. 2024.
\newblock \href {https://arxiv.org/abs/2404.01856} {Poro 34b and the blessing of multilinguality}.
\newblock \emph{Preprint}, arXiv:2404.01856.

\bibitem[{Maldonado~Rodríguez et~al.(2025)Maldonado~Rodríguez, Kovalev, Shcharbakova, Hailemariam, and Basar}]{maldonado-rodríguez-etal-2025-lcteam}
Jose Maldonado~Rodríguez, Roman Kovalev, Hanna Shcharbakova, Araya~Kiros Hailemariam, and Ezgi Basar. 2025.
\newblock {LCTeam at SemEval-2025} task 3: Multilingual detection of hallucinations and overgeneration mistakes using {XLM}-{R}oberta.
\newblock In \emph{Proceedings of the 19th International Workshop on Semantic Evaluation (SemEval-2025)}.

\bibitem[{Malhotra et~al.(2024)Malhotra, Brahme, Mishra, and Sharma}]{malhotra2024projectindus}
Nikhil Malhotra, Nilesh Brahme, Satish Mishra, and Vinay Sharma. 2024.
\newblock \href {https://www.techmahindra.com/en-in/innovation/the-indus-project/} {Project indus: A foundational model for indian languages}.

\bibitem[{Maynez et~al.(2020)Maynez, Narayan, Bohnet, and McDonald}]{maynez-etal-2020-faithfulness}
Joshua Maynez, Shashi Narayan, Bernd Bohnet, and Ryan McDonald. 2020.
\newblock \href {https://doi.org/10.18653/v1/2020.acl-main.173} {On faithfulness and factuality in abstractive summarization}.
\newblock In \emph{Proceedings of the 58th Annual Meeting of the Association for Computational Linguistics}, pages 1906--1919, Online. Association for Computational Linguistics.

\bibitem[{Mickus et~al.(2024)Mickus, Zosa, Vazquez, Vahtola, Tiedemann, Segonne, Raganato, and Apidianaki}]{mickus-etal-2024-semeval}
Timothee Mickus, Elaine Zosa, Raul Vazquez, Teemu Vahtola, J{\"o}rg Tiedemann, Vincent Segonne, Alessandro Raganato, and Marianna Apidianaki. 2024.
\newblock \href {https://doi.org/10.18653/v1/2024.semeval-1.273} {{S}em{E}val-2024 task 6: {SHROOM}, a shared-task on hallucinations and related observable overgeneration mistakes}.
\newblock In \emph{Proceedings of the 18th International Workshop on Semantic Evaluation (SemEval-2024)}, pages 1979--1993, Mexico City, Mexico. Association for Computational Linguistics.

\bibitem[{Mishra et~al.(2024)Mishra, Asai, Balachandran, Wang, Neubig, Tsvetkov, and Hajishirzi}]{mishra2024finegrained}
Abhika Mishra, Akari Asai, Vidhisha Balachandran, Yizhong Wang, Graham Neubig, Yulia Tsvetkov, and Hannaneh Hajishirzi. 2024.
\newblock \href {https://openreview.net/forum?id=dJMTn3QOWO} {Fine-grained hallucination detection and editing for language models}.
\newblock In \emph{First Conference on Language Modeling}.

\bibitem[{Mitrović et~al.(2025)Mitrović, Cornelius, Kletz, Dolamic, and Rinaldi}]{mitrovic-etal-2025-swushroomsia}
Sandra Mitrović, Joseph Cornelius, David Kletz, Ljiljana Dolamic, and Fabio Rinaldi. 2025.
\newblock {Swushroomsia at SemEval-2025} task 3: Probing {LLM}s' collective intelligence for multilingual hallucination detection.
\newblock In \emph{Proceedings of the 19th International Workshop on Semantic Evaluation (SemEval-2025)}.

\bibitem[{Mo et~al.(2025)Mo, Vorontsov, and Zang}]{mo-etal-2025-team}
Xinyuan Mo, Nikolay Vorontsov, and Tiankai Zang. 2025.
\newblock {Team Cantharellus at SemEval-2025} task 3: Hallucination span detection with fine tuning on weakly supervised synthetic data.
\newblock In \emph{Proceedings of the 19th International Workshop on Semantic Evaluation (SemEval-2025)}.

\bibitem[{Morillo et~al.(2025)Morillo, Puertas, and Santos}]{morillo-etal-2025-verbanexai}
Anderson Morillo, Edwin Puertas, and Juan Carlos~Martinez Santos. 2025.
\newblock {VerbaNexAI at SemEval-2025} task 3: Fact retrieval with {G}oogle snippets for {LLM} context filtering to identify hallucinations.
\newblock In \emph{Proceedings of the 19th International Workshop on Semantic Evaluation (SemEval-2025)}.

\bibitem[{Nguyen et~al.(2023)Nguyen, Zhang, Li, Aljunied, Xu, Chan, Hu, Shen, Chia, Li, Wang, Tan, Cheng, Chen, Deng, Yang, Liu, Zhang, and Bing}]{damonlpsg2023seallm}
Xuan-Phi Nguyen, Wenxuan Zhang, Xin Li, Mahani Aljunied, Weiwen Xu, Hou~Pong Chan, Zhiqiang Hu, Chenhui Shen, Yew~Ken Chia, Xingxuan Li, Jianyu Wang, Qingyu Tan, Liying Cheng, Guanzheng Chen, Yue Deng, Sen Yang, Chaoqun Liu, Hang Zhang, and Lidong Bing. 2023.
\newblock \href {https://arxiv.org/abs/arXiv:2312.00738} {Seallms - large language models for southeast asia}.

\bibitem[{Niu et~al.(2024)Niu, Wu, Zhu, Xu, Shum, Zhong, Song, and Zhang}]{niu-etal-2024-ragtruth}
Cheng Niu, Yuanhao Wu, Juno Zhu, Siliang Xu, KaShun Shum, Randy Zhong, Juntong Song, and Tong Zhang. 2024.
\newblock \href {https://doi.org/10.18653/v1/2024.acl-long.585} {{RAGT}ruth: A hallucination corpus for developing trustworthy retrieval-augmented language models}.
\newblock In \emph{Proceedings of the 62nd Annual Meeting of the Association for Computational Linguistics (Volume 1: Long Papers)}, pages 10862--10878, Bangkok, Thailand. Association for Computational Linguistics.

\bibitem[{Ostendorff and Rehm(2023)}]{ostendorff2023efficientlanguagemodeltraining}
Malte Ostendorff and Georg Rehm. 2023.
\newblock \href {https://arxiv.org/abs/2301.09626} {Efficient language model training through cross-lingual and progressive transfer learning}.
\newblock \emph{Preprint}, arXiv:2301.09626.

\bibitem[{Pan et~al.(2025)Pan, Bernal-Beltrán, García-Díaz, and Valencia-García}]{pan-etal-2025-umuteam}
Ronghao Pan, Tomás Bernal-Beltrán, José~Antonio García-Díaz, and Rafael Valencia-García. 2025.
\newblock {UMUTeam at SemEval-2025} task 3: Detecting hallucinations in multilingual texts using encoder-only models guided by large language models.
\newblock In \emph{Proceedings of the 19th International Workshop on Semantic Evaluation (SemEval-2025)}.

\bibitem[{Plank(2022)}]{plank-2022-problem}
Barbara Plank. 2022.
\newblock \href {https://doi.org/10.18653/v1/2022.emnlp-main.731} {The {\textquotedblleft}problem{\textquotedblright} of human label variation: On ground truth in data, modeling and evaluation}.
\newblock In \emph{Proceedings of the 2022 Conference on Empirical Methods in Natural Language Processing}, pages 10671--10682, Abu Dhabi, United Arab Emirates. Association for Computational Linguistics.

\bibitem[{Pronk et~al.(2025)Pronk, Kamyshanova, Adam, and van der Maesen~de Sombreff}]{pronk-etal-2025-bluetoad}
Michiel~T. Pronk, Ekaterina~A. Kamyshanova, Thijmen~W. Adam, and Maxim~A.X. van der Maesen~de Sombreff. 2025.
\newblock {BlueToad at SemEval-2025} task 3: Using question-answering-based language models to extract hallucinations from machine-generated text.
\newblock In \emph{Proceedings of the 19th International Workshop on Semantic Evaluation (SemEval-2025)}.

\bibitem[{Pukemo et~al.(2025)Pukemo, Levykin, Melikhov, Skiba, Ischenko, and Vorontsov}]{pukemo-etal-2025-iai_msu}
Mikhail Pukemo, Aleksandr Levykin, Dmitrii Melikhov, Gleb Skiba, Roman Ischenko, and Konstantin Vorontsov. 2025.
\newblock {iai\_MSU at SemEval-2025} task-3: {Mu-SHROOM}, the multilingual shared-task on hallucinations and related observable overgeneration mistakes in english.
\newblock In \emph{Proceedings of the 19th International Workshop on Semantic Evaluation (SemEval-2025)}.

\bibitem[{Raunak et~al.(2021)Raunak, Menezes, and Junczys-Dowmunt}]{raunak-etal-2021-curious}
Vikas Raunak, Arul Menezes, and Marcin Junczys-Dowmunt. 2021.
\newblock \href {https://doi.org/10.18653/v1/2021.naacl-main.92} {The curious case of hallucinations in neural machine translation}.
\newblock In \emph{Proceedings of the 2021 Conference of the North American Chapter of the Association for Computational Linguistics: Human Language Technologies}, pages 1172--1183, Online. Association for Computational Linguistics.

\bibitem[{Rawte et~al.(2023)Rawte, Chakraborty, Pathak, Sarkar, Tonmoy, Chadha, Sheth, and Das}]{rawte-etal-2023-troubling}
Vipula Rawte, Swagata Chakraborty, Agnibh Pathak, Anubhav Sarkar, S.M Towhidul~Islam Tonmoy, Aman Chadha, Amit Sheth, and Amitava Das. 2023.
\newblock \href {https://doi.org/10.18653/v1/2023.emnlp-main.155} {The troubling emergence of hallucination in large language models - an extensive definition, quantification, and prescriptive remediations}.
\newblock In \emph{Proceedings of the 2023 Conference on Empirical Methods in Natural Language Processing}, pages 2541--2573, Singapore. Association for Computational Linguistics.

\bibitem[{Rohrbach et~al.(2018)Rohrbach, Hendricks, Burns, Darrell, and Saenko}]{rohrbach-etal-2018-object}
Anna Rohrbach, Lisa~Anne Hendricks, Kaylee Burns, Trevor Darrell, and Kate Saenko. 2018.
\newblock \href {https://doi.org/10.18653/v1/D18-1437} {Object hallucination in image captioning}.
\newblock In \emph{Proceedings of the 2018 Conference on Empirical Methods in Natural Language Processing}, pages 4035--4045, Brussels, Belgium. Association for Computational Linguistics.

\bibitem[{Rostami et~al.(2024)Rostami, Salemi, and Dousti}]{rostami2024persianmindcrosslingualpersianenglishlarge}
Pedram Rostami, Ali Salemi, and Mohammad~Javad Dousti. 2024.
\newblock \href {https://arxiv.org/abs/2401.06466} {{PersianMind}: A cross-lingual persian-english large language model}.
\newblock \emph{Preprint}, arXiv:2401.06466.

\bibitem[{Rykov et~al.(2025)Rykov, Olisov, Savkin, Vazhentsev, Titova, Panchenko, Konovalov, and Belikova}]{rykov-etal-2025-smurfcat}
Elisei Rykov, Valerii Olisov, Maksim Savkin, Artem Vazhentsev, Kseniia Titova, Alexander Panchenko, Vasily Konovalov, and Julia Belikova. 2025.
\newblock {SmurfCat at SemEval-2025} task 3: Bridging external knowledge and model uncertainty for enhanced hallucination detection.
\newblock In \emph{Proceedings of the 19th International Workshop on Semantic Evaluation (SemEval-2025)}.

\bibitem[{Savelli et~al.(2025)Savelli, Koudounas, and Giobergia}]{savelli-etal-2025-malto}
Claudio Savelli, Alkis Koudounas, and Flavio Giobergia. 2025.
\newblock {MALTO at SemEval-2025} task 3: Detecting hallucinations in {LLM}s via uncertainty quantification and larger model validation.
\newblock In \emph{Proceedings of the 19th International Workshop on Semantic Evaluation (SemEval-2025)}.

\bibitem[{Stack-Sánchez et~al.(2025)Stack-Sánchez, Alvarez-Carmona, and Monroy}]{stack-sanchez-etal-2025-nlp_cimat}
Jaime Stack-Sánchez, Miguel~Angel Alvarez-Carmona, and Adrian Pastor~Lopez Monroy. 2025.
\newblock {NLP\_CIMAT at SemEval-2025} task 3: Just ask {GPT} or look inside. a prompt and neural networks approach to hallucination detection.
\newblock In \emph{Proceedings of the 19th International Workshop on Semantic Evaluation (SemEval-2025)}.

\bibitem[{{Team GLM} et~al.(2024){Team GLM}, Zeng, Xu, Wang, Zhang, Yin, Rojas, Feng, Zhao, Lai, Yu, Wang, Sun, Zhang, Cheng, Gui, Tang, Zhang, Li, Zhao, Wu, Zhong, Liu, Huang, Zhang, Zheng, Lu, Duan, Zhang, Cao, Yang, Tam, Zhao, Liu, Xia, Zhang, Gu, Lv, Liu, Liu, Yang, Song, Zhang, An, Xu, Niu, Yang, Li, Bai, Dong, Qi, Wang, Yang, Du, Hou, and Wang}]{glm2024chatglm}
{Team GLM}, Aohan Zeng, Bin Xu, Bowen Wang, Chenhui Zhang, Da~Yin, Diego Rojas, Guanyu Feng, Hanlin Zhao, Hanyu Lai, Hao Yu, Hongning Wang, Jiadai Sun, Jiajie Zhang, Jiale Cheng, Jiayi Gui, Jie Tang, Jing Zhang, Juanzi Li, and 37 others. 2024.
\newblock \href {https://arxiv.org/abs/2406.12793} {Chatglm: A family of large language models from glm-130b to glm-4 all tools}.
\newblock \emph{Preprint}, arXiv:2406.12793.

\bibitem[{Tufa et~al.(2025)Tufa, Hassan, Collell, Tu, Tu, Ni, and Tan}]{tufa-etal-2025-firc}
Wondimagegnhue~Tsegaye Tufa, Fadi Hassan, Guillem Collell, Dandan Tu, Yi~Tu, Sang Ni, and Kuan~Eeik Tan. 2025.
\newblock {FiRC-NLP at SemEval-2025} task 3: Exploring prompting approaches for detecting hallucinations in llms.
\newblock In \emph{Proceedings of the 19th International Workshop on Semantic Evaluation (SemEval-2025)}.

\bibitem[{Vemula and Krishnamurthy(2025)}]{vemula-krishnamurthy-2025-keepitsimple}
Saketh~Reddy Vemula and Parameswari Krishnamurthy. 2025.
\newblock {keepitsimple at SemEval-2025} task 3: {LLM}-uncertainty based approach for multilingual hallucination span detection.
\newblock In \emph{Proceedings of the 19th International Workshop on Semantic Evaluation (SemEval-2025)}.

\bibitem[{Vinyals and Le(2015)}]{vinyals2015neural}
Oriol Vinyals and Quoc Le. 2015.
\newblock \href {https://arxiv.org/abs/1506.05869} {A neural conversational model}.
\newblock \emph{Preprint}, arXiv:1506.05869.

\bibitem[{Voznyuk et~al.(2025)Voznyuk, Gritsai, and Grabovoy}]{voznyuk-etal-2025-advacheck}
Anastasia Voznyuk, German Gritsai, and Andrey Grabovoy. 2025.
\newblock {Advacheck at SemEval-2025} task 3: Combining {NER} and {RAG} to spot hallucinations in {LLM} answers.
\newblock In \emph{Proceedings of the 19th International Workshop on Semantic Evaluation (SemEval-2025)}.

\bibitem[{Wang et~al.(2023)Wang, Cheng, Zhan, Li, Song, and Liu}]{wang2023openchat}
Guan Wang, Sijie Cheng, Xianyuan Zhan, Xiangang Li, Sen Song, and Yang Liu. 2023.
\newblock Openchat: Advancing open-source language models with mixed-quality data.

\bibitem[{Wastl et~al.(2025)Wastl, Vamvas, and Sennrich}]{wastl-etal-2025-uzh}
Michelle Wastl, Jannis Vamvas, and Rico Sennrich. 2025.
\newblock {UZH at SemEval-2025} task 3: Token-level self-consistency for hallucination detection.
\newblock In \emph{Proceedings of the 19th International Workshop on Semantic Evaluation (SemEval-2025)}.

\bibitem[{Yang et~al.(2024)Yang, Yang, Hui, Zheng, Yu, Zhou, Li, Li, Liu, Huang, Dong, Wei, Lin, Tang, Wang, Yang, Tu, Zhang, Ma, Yang, Xu, Zhou, Bai, He, Lin, Dang, Lu, Chen, Yang, Li, Xue, Ni, Zhang, Wang, Peng, Men, Gao, Lin, Wang, Bai, Tan, Zhu, Li, Liu, Ge, Deng, Zhou, Ren, Zhang, Wei, Ren, Liu, Fan, Yao, Zhang, Wan, Chu, Liu, Cui, Zhang, Guo, and Fan}]{yang2024qwen2technicalreport}
An~Yang, Baosong Yang, Binyuan Hui, Bo~Zheng, Bowen Yu, Chang Zhou, Chengpeng Li, Chengyuan Li, Dayiheng Liu, Fei Huang, Guanting Dong, Haoran Wei, Huan Lin, Jialong Tang, Jialin Wang, Jian Yang, Jianhong Tu, Jianwei Zhang, Jianxin Ma, and 43 others. 2024.
\newblock \href {https://arxiv.org/abs/2407.10671} {Qwen2 technical report}.
\newblock \emph{Preprint}, arXiv:2407.10671.

\bibitem[{Yin et~al.(2025)Yin, Wang, Yang, and Lin}]{yin-etal-2025-dutjbd}
Shengdi Yin, Zekun Wang, Liang Yang, and Hongfei Lin. 2025.
\newblock {DUTJBD at SemEval-2025} task 3: A range of approaches for predicting hallucination generation in models.
\newblock In \emph{Proceedings of the 19th International Workshop on Semantic Evaluation (SemEval-2025)}.

\bibitem[{Zhang et~al.(2023{\natexlab{a}})Zhang, Press, Merrill, Liu, and Smith}]{zhang2023snowball}
Muru Zhang, Ofir Press, William Merrill, Alisa Liu, and Noah~A. Smith. 2023{\natexlab{a}}.
\newblock \href {https://arxiv.org/abs/2305.13534} {How language model hallucinations can snowball}.
\newblock \emph{Preprint}, arXiv:2305.13534.

\bibitem[{Zhang et~al.(2023{\natexlab{b}})Zhang, Li, Cui, Cai, Liu, Fu, Huang, Zhao, Zhang, Chen, Wang, Luu, Bi, Shi, and Shi}]{Zhang2023SirensSI}
Yue Zhang, Yafu Li, Leyang Cui, Deng Cai, Lemao Liu, Tingchen Fu, Xinting Huang, Enbo Zhao, Yu~Zhang, Yulong Chen, Longyue Wang, Anh~Tuan Luu, Wei Bi, Freda Shi, and Shuming Shi. 2023{\natexlab{b}}.
\newblock \href {https://api.semanticscholar.org/CorpusID:261530162} {Siren's song in the ai ocean: A survey on hallucination in large language models}.
\newblock \emph{ArXiv}, abs/2309.01219.

\bibitem[{Zhou et~al.(2021)Zhou, Neubig, Gu, Diab, Guzm{\'a}n, Zettlemoyer, and Ghazvininejad}]{zhou-etal-2021-detecting}
Chunting Zhou, Graham Neubig, Jiatao Gu, Mona Diab, Francisco Guzm{\'a}n, Luke Zettlemoyer, and Marjan Ghazvininejad. 2021.
\newblock \href {https://doi.org/10.18653/v1/2021.findings-acl.120} {Detecting hallucinated content in conditional neural sequence generation}.
\newblock In \emph{Findings of the Association for Computational Linguistics: ACL-IJCNLP 2021}, pages 1393--1404, Online. Association for Computational Linguistics.

\end{thebibliography}
